\documentclass{article} % For LaTeX2e
\usepackage{iclr2027_conference,times}

\usepackage{hyperref}
\usepackage{url}
\usepackage{booktabs}
\usepackage{amsmath}
\usepackage{amssymb}
\usepackage{mathtools}
\usepackage{amsthm}
\usepackage{dsfont}
\usepackage{xfrac}
\usepackage{multirow}
\usepackage{enumitem}
\usepackage{subcaption}
\usepackage{algorithm}
\usepackage{algpseudocode}
\usepackage{placeins}
\usepackage{fix-cm}
\usepackage{siunitx}

\DeclareMathSizes{6}{6}{4}{3}

\usepackage{tikz}
\usetikzlibrary{tikzmark,calc}
\usetikzlibrary{automata, positioning, arrows.meta}
\usetikzlibrary{fit, backgrounds, shapes}
\usepackage{xcolor}
\newcommand{\cond}{\,|\,}
\newcommand{\bigcond}{\,\big|\,}

\newcommand{\placeholder}{\, \cdot \,}
\DeclareMathOperator*{\expectation}{\mathbb{E}}
\DeclareMathOperator*{\prob}{P}
\DeclareMathOperator*{\argmax}{arg\,max}
\DeclareMathOperator*{\trace}{tr}
\DeclareMathOperator*{\supp}{supp}

\DeclareMathOperator*{\clip}{clip}
\DeclareMathOperator*{\softmax}{softmax}

\newcommand{\U}{\mathbin{\mathsf{U}}}

\newcommand{\X}{\mathop{\mathsf{X}}} 
\newcommand{\G}{\mathop{\mathsf{G}}}
\newcommand{\F}{\mathop{\mathsf{F}}}

\newcommand{\true}{\top}
\newcommand{\false}{\bot}

\usepackage[capitalize,noabbrev]{cleveref}

\theoremstyle{plain}
\newtheorem{theorem}{Theorem}[section]

\theoremstyle{definition}
\newtheorem{definition}[theorem]{Definition}
\newtheorem{assumption}[theorem]{Assumption}
\newtheorem{example}[theorem]{Example}

\theoremstyle{remark}

\title{PlatoLTL: Scaling LTL-Guided Multi-Task RL}
\author{Jacques Cloete \& Ioannis Havoutis \\
Oxford Robotics Institute\\
University of Oxford\\
Oxford, United Kingdom \\
\texttt{\{jacques,ioannis\}@robots.ox.ac.uk} \\
\AND
Mathias Jackermeier \& Alessandro Abate \\
Department of Computer Science \\
University of Oxford \\
Oxford, United Kingdom \\
\texttt{\{mathias.jackermeier,alessandro.abate\}@cs.ox.ac.uk} \\
}
\author{%
Jacques Cloete$^1$ \qquad Mathias Jackermeier$^2$ \qquad Ioannis Havoutis$^1$ \qquad Alessandro Abate$^2$\\
$^1$Oxford Robotics Institute, University of Oxford\\
$^2$Department of Computer Science, University of Oxford\\
\texttt{jacques@robots.ox.ac.uk}
}

\iclrfinalcopy % Uncomment for camera-ready version, but NOT for submission.
\begin{document}

\maketitle

\begin{abstract}
    % A central challenge in multi-task reinforcement learning (RL) is to train generalist policies capable of performing tasks not seen during training.
    % To facilitate such generalization, linear temporal logic (LTL) has emerged as a powerful formalism for specifying structured, temporally extended tasks to RL agents.
    Linear temporal logic (LTL) has emerged as a powerful formalism for specifying structured, temporally extended tasks in multi-task reinforcement learning (RL).
    However, while existing approaches in LTL-guided multi-task RL demonstrate success in simple environments, they suffer from challenges in \textit{representation learning} and \textit{exploration efficiency} when applied to high-dimensional environments with \textit{parameterized} specifications.
    We present PlatoLTL, which elevates state-of-the-art methods to address both challenges.
    We model atomic propositions as instances of \textit{atomic predicates} and inject task parameters directly into the goal embedding to enable efficient generalization.
    We also leverage \textit{simple priors} on closeness to predicate satisfaction to enable and accelerate learning of complex tasks without biasing the optimal policy.
    We validate our approach on challenging environments including robotic manipulation and multi-drone navigation.
\end{abstract}

% TLDR: PlatoLTL elevates LTL-guided multi-task RL to achieve strong performance in high-dimensional environments with parameterized specifications by learning parameter-aware predicate embeddings and using coarse domain knowledge to guide exploration.

%%%%%%%%%%%%%%%%%%%%%%%%%%%%%%%%
% INTRODUCTION
%%%%%%%%%%%%%%%%%%%%%%%%%%%%%%%%

\section{Introduction}
\label{sec:introduction}

A key challenge in robotics and artificial intelligence (AI) is to create agents capable of performing complex, long-horizon tasks. A rapidly growing \textit{neuro-symbolic} paradigm uses formal languages to define task objectives and guide the learning of policies in frameworks such as reinforcement learning (RL). Accordingly, linear temporal logic (LTL; \citealp{pnueli_temporal_1977}) has recently been adopted as a powerful formalism for specifying complex, temporally extended, and non-Markovian tasks in RL \citep{hasanbeig_certified_2023} --- a paradigm particularly valuable in settings that demand precise, verifiable behavior yet afford no large pre-training datasets, as is typical of safety-critical applications \citep{baier_principles_2008}. Recent approaches, which we refer to as \textit{backbones}, learn \textit{multi-task} policies that zero-shot generalize across tasks, i.e., LTL specifications \citep{jackermeier_deepltl_2025}.

However, when scaling to high-dimensional environments such as robotics, backbones face two key challenges. The first is \textit{representation learning}. LTL tasks are built from \textit{atomic propositions}, i.e., Boolean facts about the environment. For these environments, propositions may be \textit{parameterized}: a robotic arm may need to pick up an object $obj$, move it to position $\mathbf{x}$, or rotate its gripper to orientation $\boldsymbol{\theta}$. Backbones must either discretize the parameter space, which scales poorly with resolution and dimension, or append parameters to observations, forcing the policy to learn complex associations between goal and state embeddings, which degrades in multi-task settings. We instead model propositions as instances of \textit{atomic predicates} and inject parameters into the goal embedding; this \textit{inductive bias} yields low-dimensional per-predicate embeddings that generalize across parameters.

The second challenge is \textit{exploration efficiency}; a backbone's reward signal is typically \textit{sparse} and does not provide guidance on \textit{where} the next subgoal is, instead only signaling satisfaction or violation of the full task. For high-dimensional environments such as robotic manipulation, learning subgoals such as pick-and-place becomes infeasible under unguided exploration. We introduce \textit{robustness metrics} representing \textit{simple priors} about closeness to predicate satisfaction, and compose these metrics into a shaped reward that guides exploration without biasing the optimal policy.

% Our contributions are:
% (i) We present the \textit{Plato embedding module}, which greatly improves learning for \textit{parameterized} LTL specifications by injecting predicate parameters into the goal embedding.
% (ii) We present \textit{Plato potential-based reward shaping} (P\textsuperscript{2}BRS), which enables straightforward injection of \textit{coarse} domain knowledge to guide learning of complex tasks while keeping the optimal policy unbiased.
% (iii) We introduce high-dimensional robotics environments for LTL-guided RL including manipulation as well as quadruped and multi-drone navigation.
% (iv) We show that PlatoLTL elevates state-of-the-art backbones, raising near-zero success rates to $75$--$97\%$ in our hardest environment.

We introduce PlatoLTL (\textbf{P}arameterized \textbf{l}ogic \textbf{a}rchitecture for \textbf{t}emporal \textbf{o}bjectives in \textbf{LTL}), a backbone-agnostic framework that augments existing LTL-guided multi-task RL methods with two complementary components addressing these challenges.
Our contributions are:
(i) the \textit{Plato embedding module}, which greatly improves learning for \textit{parameterized} LTL specifications by injecting predicate parameters into the goal embedding;
(ii) \textit{Plato potential-based reward shaping} (P\textsuperscript{2}BRS), which enables straightforward injection of \textit{coarse} domain knowledge to guide learning of complex tasks while keeping the optimal policy unbiased;
(iii) a suite of high-dimensional robotics environments for LTL-guided RL including manipulation as well as quadruped and multi-drone navigation;
(iv) an empirical evaluation showing that PlatoLTL elevates state-of-the-art backbones, raising near-zero success rates to $75$--$97\%$ in our hardest environment.

\begin{figure*}[t]
    \centering
    \includegraphics[width=0.995\linewidth]{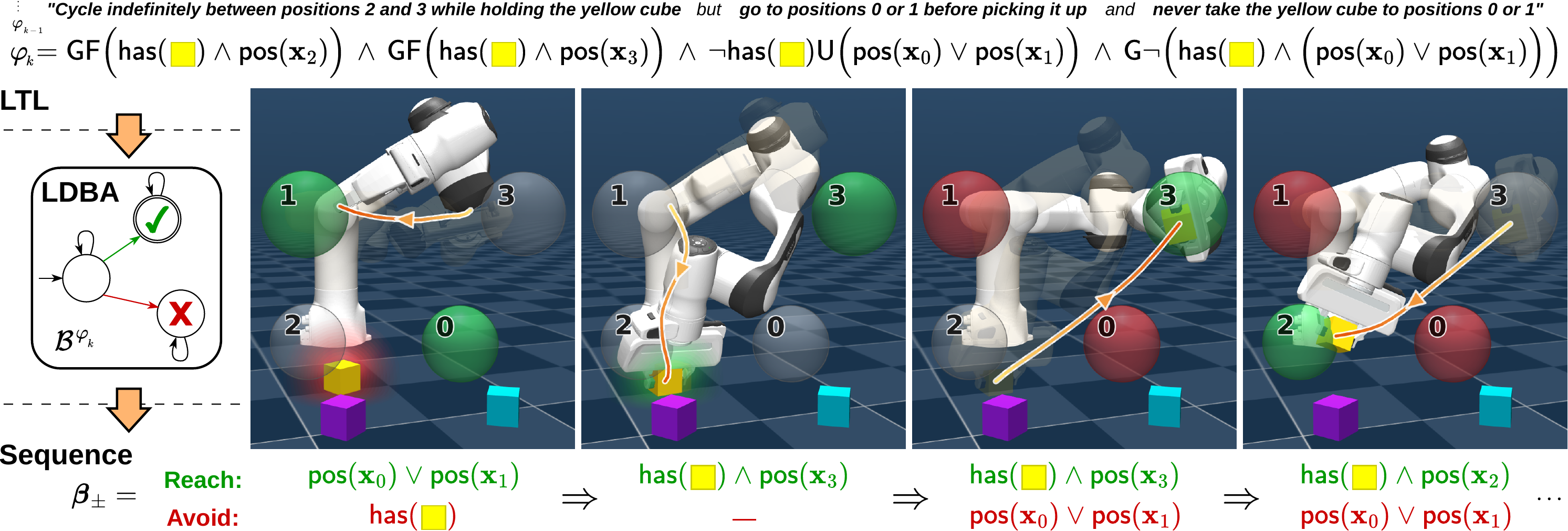}
    \caption{During training, we convert LTL specifications into LDBAs and use depth-first search to extract \textit{reach-avoid sequences} corresponding to LTL satisfaction. P\textsuperscript{2}BRS combines these sequences with per-predicate \textit{robustness metrics} to construct dense potential functions to guide policy learning.}
    \label{fig:illustration}
    \vskip -0.1in
\end{figure*}

%%%%%%%%%%%%%%%%%%%%%%%%%%%%%%%%
% RELATED WORK
%%%%%%%%%%%%%%%%%%%%%%%%%%%%%%%%

\section{Related Work}
\label{sec:related_work}

\textbf{LTL-Guided RL.} Methods for learning LTL-guided policies generally follow one of two approaches. The first converts the LTL formula into an automaton and conditions the policy directly on the automaton's states or transitions \citep{icarte_using_2018, hahn_omega-regular_2019, bertrand_deep_2020, hasanbeig_certified_2023, bozkurt_control_2020, voloshin_eventual_2023}; since these correspond only to a particular specification, such methods produce \textit{single-task} policies. In contrast, the methods we refer to as \textit{backbones} condition the policy on a learned task \textit{representation}, i.e., a goal, in a space shared across specifications \citep{vaezipoor_ltl2action_2021, qiu_instructing_2023, jackermeier_deepltl_2025, guo_one_2025, jackermeier_zero-shot_2026}, yielding \textit{multi-task} policies when trained over a distribution of specifications. Backbones typically compose goal embeddings from a fixed, discrete vocabulary of propositions, which scales poorly to the \textit{parameterized} specifications natural in robotics: binning continuous parameters into propositions is polynomial in resolution and exponential in dimension, while appending parameters to the observation \citep{jothimurugan_compositional_2021, shukla_logical_2024}, even with feature-wise modulation \citep{hatanaka_reinforcement_2025}, forces the policy to learn complex associations between goal and state embeddings that degrade in multi-task settings. We instead model propositions as instances of \textit{atomic predicates} and inject parameters into the goal embedding.

\textbf{Exploration in LTL-Guided RL.} Since LTL rewards are typically \textit{sparse}, exploration efficiency is a key challenge. Training curricula \citep{jackermeier_deepltl_2025} and hindsight relabeling \citep{qiu_instructing_2023} can speed up learning, but provide no explicit guidance towards states satisfying complex subgoals, which may be visited impractically rarely. Potential functions over automaton states \citep{camacho_ltl_2019, icarte_reward_2022, bagatella_directed_2025} preserve the optimal policy, but are constant in the environment state, and thus provide no gradient towards satisfying the current subgoal. Ground-Compose-Reinforce \citep{li_ground-compose-reinforce_2025} composes proposition-level value functions into dense potentials, but learns them offline from a dataset of labeled trajectories, trains single-task policies, and ignores \textit{implicit} orderings within conjunctions. Planner-derived dense rewards \citep{cai_overcoming_2023} are geometric and single-task. Alternative specification languages with \textit{quantitative semantics} \citep{maler_monitoring_2004, li_reinforcement_2017, jothimurugan_composable_2019} densify the satisfaction signal, but require accurate per-predicate robustness metrics, bias optimization towards expected robustness rather than satisfaction probability, and are semantically incompatible with LTL-based backbones. In contrast, our robustness metrics are \textit{simple priors} encoding coarse domain knowledge that require no pre-training data, we provide dense guidance in multi-task settings without biasing the optimal policy, and our tasks remain fundamentally LTL. See Appendix~\ref{app:extended_related_work} for extended related work.

%%%%%%%%%%%%%%%%%%%%%%%%%%%%%%%%
% BACKGROUND
%%%%%%%%%%%%%%%%%%%%%%%%%%%%%%%%

\section{Background}
\label{sec:background}

\textbf{Reinforcement Learning.} We consider a model-free reinforcement learning (RL) setup where an agent interacts with an unknown environment modeled as a \textit{Markov Decision Process} (MDP; \citealp{sutton_reinforcement_2014}). An MDP is defined as a tuple $\mathcal{M} = \langle\mathcal{S}, \mathcal{A}, p, \mu, r, \gamma\rangle$, with state space $\mathcal{S}$ and action space $\mathcal{A}$, where $p(s' \cond s, a): \mathcal{S} \times \mathcal{A} \to \Delta(\mathcal{S})$ and $\mu(s) \in \Delta(\mathcal{S})$ are the (unknown) state-transition and initial state distributions, respectively, $r(s, a, s'): \mathcal{S} \times \mathcal{A} \times \mathcal{S} \to \mathbb{R}$ is the reward function, and $\gamma \in [0, 1)$ is the discount factor. We aim to find a stochastic policy $\pi(a \cond s): \mathcal{S} \to \Delta(\mathcal{A})$ that maximizes the \textit{expected discounted return} $J(\pi) = \expectation_{\tau \sim \pi} \big[ \sum_{t = 0}^{\infty} \gamma^t r_t \big]$, where $\tau = (s_0, a_0, r_0, s_1, \dots)$ is a trajectory generated under $\pi$ and $p$ starting from $s_0 \sim \mu$, such that $a_t \sim \pi(\placeholder \cond s_t)$, $s_{t+1} \sim p(\placeholder \cond s_t, a_t)$, and $r_t = r(s_t, a_t, s_{t+1})$. The \textit{value function} of $\pi$, $V^{\pi}(s) = \expectation_{\tau \sim \pi} \big[ \sum_{t = 0}^{\infty} \gamma^t r_t \mid s_0 = s \big]$, is the expected discounted return under $\pi$ starting from $s$.

\textbf{Linear Temporal Logic.} \textit{Linear temporal logic} (LTL; \citealp{pnueli_temporal_1977}) is a human-readable formalism for specifying properties over infinite trajectories, expressing complex temporally extended behaviors including safety, liveness, recurrence, and persistence \citep{manna_hierarchy_1990}. For example, Figure~\ref{fig:illustration} specifies a robotic manipulation task in which a robot must cycle a cube indefinitely between two positions while never carrying it to two other positions. LTL formulae are composed from a finite set of \textit{atomic propositions}, $\mathit{AP}$, which represent Boolean statements about the environment. The syntax of LTL is recursively defined as
$\varphi ::= \true \mid \mathsf{p} \mid \lnot \varphi \mid \varphi \land \psi \mid \X \varphi \mid \varphi \U \psi$,
where $\true$ denotes true, $\mathsf{p} \in \mathit{AP}$ is a proposition, $\lnot$ (negation) and $\land$ (conjunction) are standard Boolean operators, $\varphi$ and $\psi$ are LTL formulae, and $\X$ (next) and $\U$ (until) are temporal operators. We define temporal operators $\F$ (eventually) and $\G$ (always) as $\F \varphi \equiv \true \U \varphi$ and $\G \varphi \equiv \lnot \F \lnot \varphi$, respectively.

Let $\Sigma \subseteq 2^{\mathit{AP}}$ be the \textit{alphabet}, in other words, the set of all permitted combinations of proposition truth values. Let $\sigma \in \Sigma$ be an \textit{assignment}, i.e., the set of propositions that are true at a given time. An \textit{$\omega$-word} $w = (\sigma_0, \sigma_1, \sigma_2, \dots)$ is an infinite sequence of assignments; we denote the set of all $\omega$-words over $\Sigma$ as $\Sigma^{\omega}$, such that $w \in \Sigma^{\omega}$. The satisfaction semantics of LTL are defined via the recursive satisfaction relation $w \models \varphi$; see Appendix~\ref{app:ltl_satisfaction_semantics} for details. To ground LTL satisfaction in an MDP, we assume access to a \textit{labeling function} $L(s): \mathcal{S} \to \Sigma$, which returns the set of propositions that are true for a given state $s$. A trajectory $\tau$ is mapped to an $\omega$-word via its trace $\trace(\tau) = \big(L(s_0), L(s_1), \dots\big)$, and $\tau \models \varphi$ is shorthand for $\trace(\tau) \models \varphi$. We thus define the \textit{satisfaction probability} of an LTL formula $\varphi$ under policy $\pi$ in MDP $\mathcal M$ as $\prob(\pi \models \varphi) = \expectation_{\tau \sim \pi} \big[ \mathds{1} [ \tau \models \varphi ] \big]$.

\textbf{B\"{u}chi Automata.} A more convenient way of reasoning about LTL satisfaction is to use \textit{B\"{u}chi automata} \citep{nagel_symposium_1966}, which are finite-state machines that can be employed to monitor the progress towards satisfying a given LTL specification. In this work we consider \textit{limit-deterministic B\"{u}chi automata} (LDBAs; \citealp{chaudhuri_limit-deterministic_2016}), which limit non-determinism to special transitions, called \textit{$\varepsilon$-transitions}, and as such are particularly suitable for use alongside MDPs. An LDBA is defined as a tuple $\mathcal{B} = \langle\mathcal{Q}, q_0, \Sigma, \delta, \mathcal{F}, \mathcal{E}\rangle$, where $\mathcal{Q}$ is the finite set of states, $q_0 \in \mathcal{Q}$ is the initial state, $\Sigma \subseteq 2^{\mathit{AP}}$ is the finite alphabet, $\delta(q,\sigma) : \mathcal{Q} \times (\Sigma \cup \mathcal{E}) \to \mathcal{Q}$ is the transition function, $\mathcal{F}$ is the set of accepting states, and $\mathcal{E}$ is the set of $\varepsilon$-transitions. An $\omega$-word $w$ is \textit{accepted} by $\mathcal{B}$ if it induces a \textit{run} $\mathbf{q} = (q_0, q_1, q_2, \dots)$ that visits accepting states $q \in \mathcal{F}$ infinitely often. While \textit{almost} deterministic, LDBAs retain the expressiveness of non-deterministic B\"{u}chi automata, and they can thus be used to represent any LTL specification: given an LTL formula $\varphi$, one can automatically construct an LDBA $\mathcal{B}^{\varphi} = \langle\mathcal{Q}^\varphi, q^\varphi_0, \Sigma^\varphi, \delta^\varphi, \mathcal{F}^\varphi, \mathcal{E}^\varphi\rangle$ such that $\mathcal{B}^\varphi$ accepts exactly the $\omega$-words $w$ for which $w \models \varphi$ \citep{chaudhuri_limit-deterministic_2016}. See Appendix~\ref{app:ldba} for further details on LDBAs alongside an example.

\textbf{Product MDPs.} A \textit{product MDP} $\mathcal{M}^\varphi$ extends the state space of $\mathcal{M}$ to monitor the state of $\mathcal{B}^\varphi$, yielding a product state $(s, q^\varphi) \in \mathcal{S} \times \mathcal{Q}^\varphi$; see Appendix~\ref{app:product_mdps} for a formal definition. Since $q^\varphi$ encodes the memory necessary to satisfy $\varphi$, this allows policies conditioned on $(s, q^\varphi)$ to be \textit{memoryless} \citep{hahn_omega-regular_2019, bozkurt_control_2020}. In practice, we never explicitly construct the product MDP, but instead update $q^\varphi$ using the assignment $\sigma^\varphi = L^\varphi(s)$ observed at each time step.

%%%%%%%%%%%%%%%%%%%%%%%%%%%%%%%%
% PROBLEM SETTING
%%%%%%%%%%%%%%%%%%%%%%%%%%%%%%%%

\section{Problem Setting}
\label{sec:problem_setting}

We use RL to obtain a policy $\pi(a \cond s, q^\varphi)$ that maximizes the probability of satisfying an LTL formula $\varphi \sim \xi$, where $\xi$ is an arbitrary distribution over LTL formulae with support $\supp(\xi)$. The optimal policy $\pi^*$ over the product MDP $\mathcal{M}^\varphi$ is thus defined as
\begin{equation}
    \label{eqn:rl_problem}
    \pi^* = \argmax_\pi \expectation_{\substack{\varphi \sim \xi, \\ \tau \sim \pi \cond \varphi}}\big[ \mathds{1} [ \tau \models \varphi ] \big].
\end{equation}
There exist a variety of approaches to solving Equation \ref{eqn:rl_problem}; we refer to such approaches as \textit{backbones}, which design and optimize reward functions $r^\varphi(s, q^{\varphi}, a, s^\prime, q^{\prime\varphi})$ aligned with LTL satisfaction.

However, in a departure from existing literature, we model all propositions as instances drawn from a shared set of known \textit{atomic predicates}, $\mathit{Pred}$, which represent \textit{parameterized} Boolean statements about the environment, such that the truth value depends on the input parameter values. In other words, each proposition in a specification is equivalently a \textit{predicate instance} with a particular set of fixed parameters.\footnote{While not technically needed, Appendix~\ref{app:plato_theory_of_forms} draws a connection to Plato's Theory of Forms \citep{plato_republic_1992}.} The number of instances of each predicate, and their parameter values, vary across formulae $\varphi \sim \xi$, giving each formula its own set of predicate instances $\mathit{AP}^\varphi$.

Given parameters $x_\mathsf{f} \in \mathcal{X}_\mathsf{f}$, where $\mathcal{X}_\mathsf{f} \subseteq \mathbb{R}^{n_\mathsf{f}}$ is the $n_\mathsf{f}$-dimensional parameter space for $\mathsf{f}$, the mapping $\mathsf{f} : \mathcal{X}_\mathsf{f} \to \mathit{AP}^\varphi$ instantiates predicate $\mathsf{f} \in \mathit{Pred}$ as a proposition $\mathsf{p} \in \mathit{AP}^\varphi$; thus $\mathsf{p} = \mathsf{f}(x_\mathsf{f})$ is a predicate instance of $\mathsf{f}$. Each predicate $\mathsf{f}$ has a labeling function $L_{\mathsf{f}}(s, x_\mathsf{f}): \mathcal{S} \times \mathcal{X}_\mathsf{f} \to \{0, 1\}$; given fixed parameters $x_\mathsf{f}$, the labeling function for each instance becomes $L_{\mathsf{f}(x_\mathsf{f})}(s): \mathcal{S} \to \{0, 1\}$, and these combine into the labeling function $L^\varphi(s): \mathcal{S} \to \Sigma^\varphi$ across all predicate instances in $\varphi$, where $\Sigma^\varphi \subseteq 2^{\mathit{AP}^\varphi}$. Our modeling assumption is thus formalized as follows:
\begin{assumption}
    \label{asm:pred_inst}
    Given an LTL specification $\varphi \sim \xi$, all atomic propositions $\mathsf{p} \in \mathit{AP}^\varphi$ are such that $\mathsf{p} = \mathsf{f}(x_\mathsf{f})$ for some atomic predicate $\mathsf{f} \in \mathit{Pred}$ and fixed parameters $x_\mathsf{f} \in \mathcal{X}_\mathsf{f}$. While $\mathit{Pred}$ and $\mathcal{X}_\mathsf{f}$ are shared across specifications, each proposition's $\mathsf{f}$ and $x_\mathsf{f}$ can vary between specifications.
\end{assumption}
Note that $\mathcal{X}_\mathsf{f}$ can be a singleton, in which case $\mathsf{f}$ always creates the same proposition $\mathsf{p}$: thus, Assumption \ref{asm:pred_inst} is compatible with atomic propositions (e.g., from standard LTL) that are not parameterized. Parameterization induces two axes of (zero-shot) generalization: over formula \emph{structures} and over predicate \emph{parameters}. By \emph{multi-task}, we refer to generalization over structures; generalizing only over parameters of a fixed formula structure remains \emph{single-task}. Our setting requires both.

\textbf{Problem Statement.} While Equation \ref{eqn:rl_problem} can \textit{in principle} be solved under these new assumptions using existing backbones\footnote{Although our parameterized specifications formally belong to a fragment of first-order LTL (FOLTL; \citealp{gigante_introduction_2025}), due to the parameters being fixed, the logic reduces to LTL; see Appendix~\ref{app:fo-ltl} for details.}, the parameterized specifications induced by Assumption \ref{asm:pred_inst}, and the complex high-dimensional environments to which they often apply, present significant challenges in \textit{representation learning} and \textit{exploration efficiency} that can prevent learning altogether in practice. Accordingly, rather than introducing an entirely new backbone, we present novel methods that augment existing backbones, enabling and accelerating learning in these settings; Section~\ref{sec:plato} addresses representation learning, and Section~\ref{sec:p2brs} addresses exploration efficiency.

\textbf{Choice of Backbone.} While PlatoLTL is not tied to a particular backbone, we use StructLTL \citep{jackermeier_zero-shot_2026} as the example backbone throughout this paper; assumptions regarding backbones are made clear where needed. StructLTL first trains a policy $\pi_\beta(a \cond s, \boldsymbol{\beta}_\pm)$ conditioned on an arbitrary \textit{reach-avoid sequence} $\boldsymbol{\beta}_\pm$ of Boolean formulae of propositions. Intuitively, $\boldsymbol{\beta}_\pm$ represents an accepting run in an LDBA, informing the policy of the subgoals required for satisfaction. To obtain a learned embedding of $\boldsymbol{\beta}_\pm$, StructLTL uses a hierarchical architecture to encode the logical structure of each formula, and applies an attention mechanism to reason about future subgoals.

At evaluation time, given a new LTL specification $\varphi$, StructLTL constructs the LDBA $\mathcal{B}^\varphi$, and uses a depth-first search (DFS) to extract a reach-avoid sequence representing an accepting run from $q^\varphi$, i.e., $\boldsymbol{\beta}^\varphi_\pm = \text{DFS}_{\mathcal{B}^\varphi}(q^\varphi)$; the policy $\pi$ is thus induced as $\pi(a \cond s, q^\varphi) = \pi_\beta\big(a \cond s, \text{DFS}_{\mathcal{B}^\varphi}(q^\varphi)\big)$. Each Boolean formula in the sequence is associated with a transition in $\mathcal{B}^\varphi$, and succinctly represents the set of assignments that trigger that transition. Since the policy was trained on arbitrary sequences, it generalizes across LDBAs, and therefore across LTL specifications. See Appendix~\ref{app:struct_ltl} for details.

% we assume that each LTL formula $\varphi \sim \xi$ has its own (finite) set of propositions $\mathit{AP}^\varphi$; accordingly, the alphabet $\Sigma^\varphi \subseteq 2^{\mathit{AP}^\varphi}$, assignments $\sigma^\varphi \in \Sigma^\varphi$, and labeling function $L^\varphi(s): \mathcal{S} \to \Sigma^\varphi$ are also conditioned on $\varphi$.

%%%%%%%%%%%%%%%%%%%%%%%%%%%%%%%%
% PLATO EMBEDDING MODULE
%%%%%%%%%%%%%%%%%%%%%%%%%%%%%%%%

% \section{Method}
\section{Plato Embedding Module}
\label{sec:plato}

A key challenge in solving Equation \ref{eqn:rl_problem} is \textit{representation learning}: learning an encoder for $q^\varphi$ that generalizes across specifications. State-of-the-art backbones map $q^\varphi$ to a useful symbolic representation which they encode into a learned goal embedding \citep{vaezipoor_ltl2action_2021, jackermeier_zero-shot_2026}.
The symbolic representations are typically composed from individual propositions, and the goal embeddings are thus likewise composed from embeddings of the propositions, given a learned encoder $\phi$. In summary, we assume the following about our backbones:
\begin{assumption}
    \label{asm:embedding}
    The backbone learns an encoder $\phi(\mathsf{p}) : \mathbb{Z}_{\geq 0} \to \mathbb{R}^{n_\phi}$ that maps the \textit{discrete} token ID of an atomic proposition $\mathsf{p}$ into an $n_\phi$-dimensional embedding space.
\end{assumption}

To accept a \textit{parameterized} proposition $\mathsf{p} = \mathsf{f}(x_\mathsf{f})$ induced by Assumption \ref{asm:pred_inst}, such backbones must either discretize the parameter space $\mathcal{X}_\mathsf{f}$, which scales poorly with resolution and parameter dimension, or augment the MDP state $s$ with the parameters $x_\mathsf{f}$, which requires learning a complex association between propositions in the goal embedding and parameters in the state embedding.

We propose replacing the learned mapping with the \textit{Plato embedding module}, which learns a small MLP $\phi_\mathsf{f}: \mathbb{R}^{n_\mathsf{f}} \to \mathbb{R}^{n_\phi}$ for each predicate such that $\phi\big(\mathsf{f}(x_\mathsf{f})\big) = \phi_\mathsf{f}(x_\mathsf{f})$. This simple but effective design provides a strong \textit{inductive bias} by injecting the parameters directly into the goal structure, helping the policy learn low-dimensional geometries for each predicate in embedding space and exploit underlying structures relating the propositions, and thus facilitating generalization across parameters. The model size of each MLP $\phi_\mathsf{f}$ is very small and remains orders of magnitude below the overall policy network; thus, the total parameter cost grows linearly and mildly in $|\mathit{Pred}|$ (which is itself small). Predicates with singleton parameter spaces reduce to the simple learned mapping.

%%%%%%%%%%%%%%%%%%%%%%%%%%%%%%%%
% POTENTIAL-BASED REWARD SHAPING
%%%%%%%%%%%%%%%%%%%%%%%%%%%%%%%%

\section{Plato Potential-Based Reward Shaping}
\label{sec:p2brs}

Another key challenge in solving Equation \ref{eqn:rl_problem} is \textit{exploration efficiency}. For existing backbones, the reward function $r^\varphi$ is typically \textit{sparse} and does not provide guidance on \textit{how} to achieve the next subgoal, instead only signaling satisfaction or violation of the full task \citep{vaezipoor_ltl2action_2021, jackermeier_deepltl_2025}. For complex, high-dimensional tasks such as robotic manipulation, learning how to execute subgoals such as pick-and-place becomes infeasible under unguided exploration.

We propose guiding exploration using \textit{Plato potential-based reward shaping} (P\textsuperscript{2}BRS), which guides the learning of subgoals while remaining \textit{unbiased} with regard to solving Equation \ref{eqn:rl_problem} using a given backbone. We leverage potential-based reward shaping (PBRS; \citealp{ng_policy_1999}), augmenting the backbone reward with the discounted difference of a potential function $\Phi(s, q^{\varphi})\colon \mathcal{S} \times \mathcal{Q}^\varphi \to \mathbb{R}$, i.e., $r^\varphi_\mathrm{aug} = r^\varphi + \gamma \Phi(s^\prime, q^{\prime\varphi}) - \Phi(s, q^{\varphi})$. For any bounded $\Phi$, this is well known to leave the optimal policy unchanged \citep{ng_policy_1999}; see Appendix~\ref{app:pbrs_proof} for the argument in our setting.

The remainder of this section synthesizes a suitable $\Phi$ for guiding subgoal satisfaction. Section~\ref{sec:p2brs:robustness} introduces \textit{robustness metrics} that encode how close a proposition is to satisfaction as a real-valued signal; these metrics act as \textit{simple priors}, requiring only coarse domain knowledge while densifying the sparse reward. Section~\ref{sec:p2brs:robustness:bounded_robustness} bounds and adaptively calibrates these metrics, and Section~\ref{sec:p2brs:robustness:boolean_formulae} composes them into robustness metrics for Boolean formulae. Section~\ref{sec:p2brs:potential_function} uses these to define the potential function $\Phi_\beta(s, \boldsymbol{\beta}_\pm)$ over reach-avoid sequences $\boldsymbol{\beta}_\pm$ of such formulae, i.e., sequences of subgoals; we use $\Phi_\beta$ to guide the agent along accepting runs extracted from the LDBA $\mathcal{B}^\varphi$.

\subsection{From Boolean Satisfaction to Dense Guidance}
\label{sec:p2brs:robustness}

Inspired by specification languages with quantitative semantics \citep{maler_monitoring_2004, li_reinforcement_2017, jothimurugan_composable_2019}, we introduce a \textit{robustness metric} $g_\mathsf{f}(s, x_\mathsf{f}): \mathcal{S} \times \mathcal{X}_\mathsf{f} \to \mathbb{R} \cup \{ \pm \infty \}$ for each predicate $\mathsf{f}$, which indicates how close $\mathsf{f}(x_\mathsf{f})$ is to being satisfied as a dense signal that is positive on satisfaction and negative on violation, i.e., $L_\mathsf{f}(s, x_\mathsf{f}) = \mathcal{H}\big(g_\mathsf{f}(s, x_\mathsf{f})\big)$. This can be ensured by applying the following construction to some metric $\hat{g}_\mathsf{f}(s, x_\mathsf{f}): \mathcal{S} \times \mathcal{X}_\mathsf{f} \to \mathbb{R} \cup \{ \pm \infty \}$:
\begin{equation*}
    g_\mathsf{f}(s, x_\mathsf{f}) =
    \begin{cases}
        \max\big\{\delta, \hat{g}_\mathsf{f}(s, x_\mathsf{f})\big\} & \text{if } L_\mathsf{f}(s, x_\mathsf{f}) = 1, \\
        \min\big\{-\delta, \hat{g}_\mathsf{f}(s, x_\mathsf{f})\big\} & \text{else},  
    \end{cases}
    \quad \text{for arbitrarily small } \delta > 0.
\end{equation*}

Since we bound the robustness metrics in Section \ref{sec:p2brs:robustness:bounded_robustness}, \textit{any} choice of $\hat{g}_\mathsf{f}$ leaves the optimal policy unbiased. However, not all metrics are useful for \textit{learning}, and a good metric should provide guidance on closeness to satisfaction. This guidance can be \textit{coarse}; for example, for learning to pick up an object with a manipulator arm (with proposition $\mathsf{has}(obj)$), negated distance between its grasp center and the estimated object center can suffice. By contrast, quantitative semantics would require a metric whose sign exactly characterizes a successful grasp; since our construction takes the sign from $L_\mathsf{f}$, our robustness metrics need not be highly accurate, but instead act as \textit{simple priors} that inject some domain knowledge to help guide the agent towards satisfying a particular proposition. If there is no domain knowledge available for a predicate $\mathsf{f}$, we can trivially construct a valid robustness metric $g_\mathsf{f}(s, x_\mathsf{f}) = \pm \infty$ for $L_\mathsf{f}(s, x_\mathsf{f}) = 1 / 0$ to unblock the use of shaping for other predicates.

\subsubsection{Bounded Metrics and Adaptive Calibration}
\label{sec:p2brs:robustness:bounded_robustness}

Unbiased shaping requires that $\Phi(s, q^\varphi)$ remains \textit{bounded}. Furthermore, arbitrary robustness metrics may have vastly different scales between different predicates, rendering comparisons unsuitable. Accordingly, we use the following transformation to obtain the bounded robustness metric $\rho_\mathsf{f}$:
\begin{equation*}
    \rho_\mathsf{f}(s, x_\mathsf{f}) = \frac{g_\mathsf{f}(s, x_\mathsf{f})}{\kappa_\mathsf{f} \tilde{g}_\mathsf{f} + |g_\mathsf{f}(s, x_\mathsf{f})|}, \quad \kappa_\mathsf{f}, \tilde{g}_\mathsf{f} > 0, \text{ with } \rho_\mathsf{f} = \pm 1 \text{ when } g_\mathsf{f} = \pm \infty.
\end{equation*}
This transformation preserves $L_\mathsf{f}(s, x_\mathsf{f}) = \mathcal{H}\big(\rho_\mathsf{f}(s, x_\mathsf{f})\big)$, while bounding the output of $\rho_\mathsf{f}$ to $[-1, 1]$ and tuning the sensitivity of $\rho_\mathsf{f}$ such that $|g_\mathsf{f}(s, x_\mathsf{f})| = \kappa_\mathsf{f} \tilde{g}_\mathsf{f} \iff |\rho_\mathsf{f}(s, x_\mathsf{f})| = 0.5$.

The \textit{initial robustness scale} $\tilde{g}_\mathsf{f}$ and \textit{robustness scale factor} $\kappa_\mathsf{f}$ control the breadth of guidance from the decision boundary; a larger $\kappa_\mathsf{f}\tilde{g}_\mathsf{f}$ yields broader guidance with a gradient that remains non-flat further from the boundary, while a smaller $\kappa_\mathsf{f}\tilde{g}_\mathsf{f}$ yields sharper guidance close to the boundary but a flatter gradient when at a distance. Thus, the \textit{robustness scale} $\kappa_\mathsf{f}\tilde{g}_\mathsf{f}$ acts as a normalizer for $g_\mathsf{f}$.

The initial robustness scale acts as an initial estimate for the normalizer; we obtain $\tilde{g}_\mathsf{f}$ as the quantile $q_g$ of $\mathcal{D}_\mathsf{f}^{n_g}$, i.e., $\tilde{g}_\mathsf{f} = Q_{q_g}\big(\mathcal{D}_\mathsf{f}^{n_g}\big)$, where $\mathcal{D}_\mathsf{f}^{n_g} = \big \{| g_\mathsf{f}[i]| \bigcond \delta < |g_\mathsf{f}[i]| < \infty \big\}^{n_g}_{i=1}$ consists of the first $n_g$ data points encountered during training for the magnitude of an active $g_\mathsf{f}$. Before obtaining $\tilde{g}_\mathsf{f}$, a reasonable placeholder value is $\tilde{g}_\mathsf{f} = 1$, and $\kappa_\mathsf{f} = 1$ as well during this phase.

As training progresses, the distribution of agent trajectories shifts, and thus the distribution of robustness metric values shifts. We adjust $\kappa_\mathsf{f}$ accordingly; let $\mathcal{D}_\mathsf{f}^{n_\kappa} = \big \{\sfrac{|g_\mathsf{f}[i]|}{\tilde{g}_\mathsf{f}} \bigcond \delta < |g_\mathsf{f}[i]| < \infty \big\}^{n_\kappa}_{i=1}$ be a sliding window of the $n_\kappa$ most recent data points encountered during training for the magnitude of an active $\sfrac{g_\mathsf{f}}{\tilde{g}_\mathsf{f}}$ (after having obtained $\tilde{g}_\mathsf{f}$). The per-iteration update scheme for $\kappa_\mathsf{f}$ is given by:
\begin{gather*}
    \log\big(\kappa_\mathsf{f}^{(k)}\big) = \clip\Big( \log\big(\bar{\kappa}_\mathsf{f}^{(k)}\big), \log(\kappa_\mathrm{min}), \log(\kappa_\mathrm{max}) \Big), \\
    \text{where} \quad \log\big(\bar{\kappa}_\mathsf{f}^{(k)}\big) = (1-\lambda_\kappa) \log\big(\bar{\kappa}_\mathsf{f}^{(k-1)}\big) + \lambda_\kappa \log\Big(Q_{q_\kappa}\big(\mathcal{D}_\mathsf{f}^{n_\kappa}\big)\Big) \quad \text{and} \quad \log\big(\bar{\kappa}_\mathsf{f}^{(0)}\big) = 0.
\end{gather*}
The scale factor therefore tracks the agent's typical distance from the decision boundary: trajectories further away yield a larger $\kappa_\mathsf{f}$ and broader, more global shaping that indicates roughly where the boundary lies, while trajectories closer in yield a smaller $\kappa_\mathsf{f}$ and finer, more local shaping that resolves its exact location. The online calibration scheme thus acts as an \textit{automatic curriculum}. Note that $\kappa_\mathsf{f}$ and $\tilde{g}_\mathsf{f}$ are frozen per episode to ensure telescoping of the potential function.

\subsubsection{Dense Guidance for Boolean Formulae}
\label{sec:p2brs:robustness:boolean_formulae}

Given per-proposition robustness metrics, we seek a robustness metric $\rho_\beta$ for a Boolean formula $\beta$ composed from those propositions. We assume that $\beta$ is in disjunctive normal form (DNF), which holds for all Boolean formulae $\beta^+_i$ and $\beta^-_i$ in a sequence $\boldsymbol{\beta}^\varphi_\pm$ extracted as in Section~\ref{sec:p2brs:potential_function}:
\begin{assumption}
    \label{asm:dnf}
    $\beta = \bigvee_{k=1}^{n_\beta}\bigwedge_{j=1}^{m_{\beta k}} l_{kj}$, where all $l_{kj} \in \{\mathsf{f}(x_\mathsf{f}), \lnot \mathsf{f}(x_\mathsf{f})\}$ and $\mathsf{f}\in \mathit{Pred}, x_\mathsf{f} \in \mathcal{X}_\mathsf{f}$.
\end{assumption}

A natural choice negates $\rho_\mathsf{f}$ for negated literals, as in quantitative semantics~\citep{maler_monitoring_2004}, and takes the $\min$ over the literals of each clause and the $\max$ over clauses (as also in \citealp{li_ground-compose-reinforce_2025}). However, the $\min$ ignores \textit{implicit} orderings induced by the environment dynamics, where satisfying one literal first requires satisfying another and then maintaining it. For example, a manipulator tasked with $\mathsf{has}(obj) \land \mathsf{pos}(x,y,z)$, i.e., holding $obj$ with its grasp center at $(x,y,z)$, must first grasp the object and then carry it there, yet the $\min$ guides the gripper towards a point between the object and the target (Appendix~\ref{app:ablation_studies:p2brs:persistence_gating}), leaving the grasp to be discovered by chance.

We resolve this with \textit{persistence gating}. We annotate each predicate $\mathsf{f}$ as either \textit{persistent} ($\mathsf{f} \in \mathit{Pred}^+ \subseteq \mathit{Pred}$) or not, where a persistent predicate is one whose satisfaction typically carries over while the agent works towards the other literals of a clause, and should thus be prioritized. In the example above, a held object travels with the gripper, so $\mathsf{has}$ is persistent, whereas the gripper position does not survive fetching the object, so $\mathsf{pos}$ is not; the same applies to, e.g., possessing an item or riding in a vehicle. Like the robustness metrics, these binary labels represent coarse domain knowledge and affect only the guidance, not the optimal policy.

Persistence gating treats each clause $c_k$ as a two-stage subgoal: first satisfy its persistent literals, then the rest while maintaining them. Let $h_k(s)$ and $t_k(s)$ be the $\min$ of the literal robustness metrics over the persistent and non-persistent literals of $c_k$, respectively. We split the unsatisfied range $[-1, 0)$ at $-\sigma$, for $\sigma \in (0, 1)$, so that the first stage occupies $[-1, -\sigma)$ and tracks only $h_k$, the second stage occupies $[-\sigma, 0)$ and tracks only $t_k$, and a satisfied clause keeps the plain $\min$ over literals:
\begin{equation*}
    \rho_{c_k}(s) = h_k(s) \wedge_\sigma t_k(s), \quad
    a \wedge_\sigma b =
    \begin{cases}
        -\sigma + (1-\sigma)\, a & \text{if } a < 0, \\
        \hphantom{-}\sigma\, b & \text{if } a \geq 0 > b, \\
        \hphantom{-}\min(a, b) & \text{otherwise},
    \end{cases} \quad
    \rho_\beta(s) = \max_{k} \rho_{c_k}(s).
\end{equation*}
Every second-stage state thus scores above every first-stage state, so $\rho_{c_k}$ guides the agent to satisfy persistent literals first (e.g., grasping the object before moving it). Clauses whose literals are all persistent or all non-persistent use the plain $\min$ over literals; see Appendix~\ref{app:persistence_gating} for the full definition.

\subsection{Sequence-Conditioned Potential Function}
\label{sec:p2brs:potential_function}

We note that under PBRS (and thus P\textsuperscript{2}BRS), the potential function acts as an a priori initialization of the value estimates \citep{wiewiora_potential-based_2003}. Thus, given an $n$-step reach-avoid sequence of Boolean formulae $\boldsymbol\beta_\pm =  \big((\beta^+_i, \beta^-_i)\big)^{n-1}_{i=0}$, we use $\Phi_\beta(s, \boldsymbol\beta_\pm) = r_\mathrm{rch} \rho_{\beta_0^+}(s) - r_\mathrm{avd} \rho_{\beta_0^-}(s)$, where $r_\mathrm{rch}, r_\mathrm{avd} \in \mathbb{R}^+$; this design induces the prior that there is increasing value when closer to satisfying the current reach subgoal, and that there is decreasing value when closer to satisfying the current avoid subgoal.

During training, given an LTL specification $\varphi$, we apply DFS to extract a valid sequence representing an accepting run from $q^\varphi$, i.e., $\boldsymbol{\beta}^\varphi_\pm = \text{DFS}_{\mathcal{B}^\varphi}(q^\varphi)$, and compute $\Phi(s, q^\varphi) = \Phi_\beta\big(s, \text{DFS}_{\mathcal{B}^\varphi}(q^\varphi)\big)$; see Appendix~\ref{app:reach_avoid_sequence_extraction} for details on extracting $\boldsymbol{\beta}^\varphi_\pm$. If the training-time policy is also sequence-conditioned, i.e., $\pi_\beta(a \cond s, \boldsymbol{\beta}_\pm)$, as is the case for StructLTL, then we forgo the DFS and use $\Phi_\beta(s, \boldsymbol\beta_\pm)$ directly.

\textbf{Terminal Potential Function.} Unbiased return assumes an infinite horizon; however, this does not hold in the case of early terminations where $T < \infty$, e.g., upon satisfaction or violation of a sequence or any environment-specific termination conditions, which yields a bias of $\gamma^{T} \Phi_\beta(s_{T}, \boldsymbol\beta_{\pm, T})$. To eliminate bias, we override $\Phi_\beta(s_{T}, \boldsymbol\beta_{\pm, T}) = 0$ for all terminations, regardless of episode outcome.

\textbf{Robustness Metric for $\boldsymbol{\varepsilon}$-Subgoals.} An $\varepsilon$-transition appears at most once per sequence and only in the reach half. Such transitions represent the non-deterministic jumps required for \textit{persistence} ($\F \G$) tasks, where the agent must satisfy some property for all time after the jump. An $\varepsilon$-transition has no well-defined robustness metric of its own, but it is always immediately followed by a non-empty reach subgoal, i.e., the property to be persistently satisfied, which must already hold for the jump to be taken. Since exploration for the jump is trivial, we set $\rho_{\beta_0^+}(s) = \rho_{\beta_1^+}(s)$ whenever $\beta_0^+ = \varepsilon$.

%%%%%%%%%%%%%%%%%%%%%%%%%%%%%%%%
% EXPERIMENTS
%%%%%%%%%%%%%%%%%%%%%%%%%%%%%%%%

\section{Experiments}
\label{sec:experiments}

We conduct experiments to answer the following research questions: (\textbf{Q1}) How does the Plato embedding module compare to existing multi-task approaches for learning \textit{parameterized} LTL specifications? (\textbf{Q2}) Does P\textsuperscript{2}BRS enable and accelerate learning in \textit{high-dimensional} environments?

\textbf{Baselines.} We compare the Plato embedding module (``Plato'') to three baseline approaches for embedding parameterized LTL specifications. ``Discretize'' bins parameters into a discrete set of embeddings spanning the parameter space, representing the unaugmented backbone \citep{jackermeier_zero-shot_2026}. Inspired by \citet{jothimurugan_compositional_2021}, ``Concatenate'' appends the parameters to the observation space.
% (assuming a maximum number of unique propositions per specification).
Finally, SIAMS \citep{hatanaka_reinforcement_2025} uses Feature-wise Linear Modulation (FiLM; \citealp{perez_film_2018}) to transform the backbone's state embedding using the parameters.

We do not include the observation reduction scheme of \citet{guo_one_2025}, which remaps observation components to generic ``reach'' and ``avoid'' observations. This remapping assumes that the observation space can be partitioned into proposition-specific observations, produced by identical observation functions under output transformation, and that a ``fusion operator'' can meaningfully fuse them. This assumption fails for our environments, even with parameters appended to the observations; though proposition-specific, the parameters cannot be fused into conjunctions or disjunctions.

\textbf{Environments.} Environments from the literature are typically simple abstractions with a small set of propositions shared by all specifications \citep{jackermeier_zero-shot_2026}. We instead introduce high-dimensional, predicate-based robotics environments supporting high specification complexity.
This enables evaluating generalization at scales of complexity beyond existing benchmarks.

The continuous \textit{Go1RGBZoneEnv} environment expands upon \citet{vaezipoor_ltl2action_2021}'s \textit{ZoneEnv}, and requires a Unitree Go1 quadruped robot to navigate between colored zones. To observe zones, the quadruped has a LiDAR sensor that measures the distance and RGB value of the closest zone in each bin. The policy controls the velocity command for a simple pre-trained locomotion controller.

The continuous \textit{X2DronesEnv} environment loosely expands upon \citet{voloshin_eventual_2023}'s \textit{FlatWorld}, and requires two Skydio X2 quadcopter drones to navigate between an arbitrary number of positions in space. The environment admits \textit{homogeneous conjunctions} ($\mathsf{pos}_i \land \mathsf{pos}_j$) since the drones can be at two positions simultaneously. The policy commands each drone's model-based velocity controller.

The continuous \textit{FrankaCubesEnv} environment loosely expands upon \citet{jackermeier_zero-shot_2026}'s \textit{WarehouseEnv}, and requires a Franka Panda manipulator arm to pick and place a collection of colored cubes. The environment admits \textit{heterogeneous conjunctions} (e.g., $\mathsf{has}_i \land \mathsf{pos}_j \land \mathsf{rot}_k$) since the arm can move to a position and orientation while holding a cube. The policy controls the grasp-center twist command, tracked by a simple model-based position controller, and the gripper state.

The discrete \textit{FalloutWorld} environment expands upon \citet{vaezipoor_ltl2action_2021}'s \textit{LetterWorld}, and requires a 2D grid-based agent to navigate between an arbitrary number of coordinates within a dense radiation field. See Appendix~\ref{app:experimental_details:environments} for details on environments, including predicates and metrics.

\textbf{LTL Specifications.} We evaluate the trained policy on a variety of LTL specifications representing advanced (and environment-specific) behaviors including reachability, safety, recurrence, persistence, and combinations thereof, and include both \textit{finite}-horizon and \textit{infinite}-horizon specifications. See Appendix~\ref{app:experimental_details:eval_ltl_specifications} for details on the template LTL specifications used for evaluation.

\textbf{Evaluation Protocol.} All methods are trained using the StructLTL backbone; see Appendix~\ref{app:experimental_details:training_details} for training details. Tables report success rate and average episode length on success (or N/A if success rate $< 5\%$) for \textit{finite} tasks, and average number of visits to accepting states for \textit{infinite} tasks, over $512$ episodes. Figures report success rate and average episode length over all \textit{finite} tasks, over $16$ episodes per specification. We use an evaluation horizon of $2000$ steps, average results over $10$ seeds, and provide 95\% stratified bootstrap confidence intervals for all results \citep{agarwal_deep_2021}.

%%%%%%%%%%%%%%%%%%%%%%%%%%%%%%%%
% RESULTS & DISCUSSION
%%%%%%%%%%%%%%%%%%%%%%%%%%%%%%%%

\section{Results \& Discussion}
\label{sec:results_discussion}

\begin{figure*}[t]
    \centering
    \includegraphics[width=0.995\linewidth]{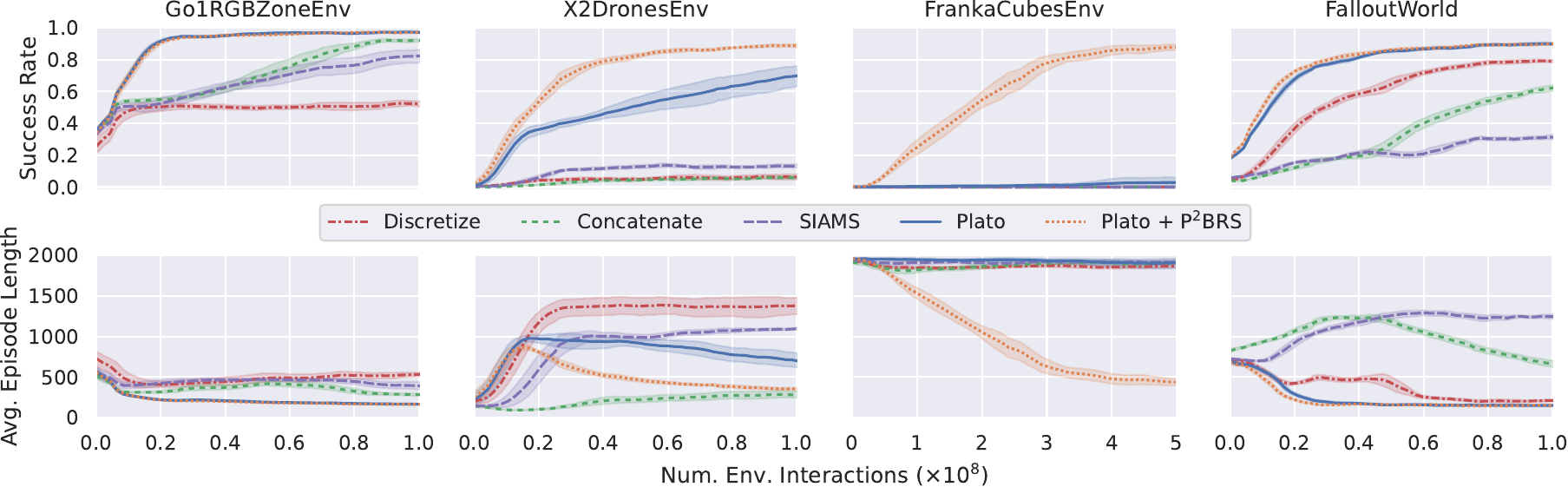}
    \caption{Evaluation curves during training for \textit{finite}-horizon specifications.}
    \label{fig:eval_curves}
    \vskip -0.1in
\end{figure*}

Figure~\ref{fig:eval_curves} illustrates evaluation curves during training for \textit{finite} tasks. Table~\ref{tab:results_finite} presents evaluation results for \textit{finite} tasks. See Appendix~\ref{app:infinite} for \textit{infinite} tasks, including $\psi_{16}$ illustrated by Figure~\ref{fig:illustration}, and see Appendix~\ref{app:trajectory_visualizations} for visualizations of example trajectories for Plato with P\textsuperscript{2}BRS.

\begin{table*}[t]
    \caption{Evaluation results for \textit{finite}-horizon LTL specifications.}
    \label{tab:results_finite}
    \begin{center}
    \renewcommand{\arraystretch}{0.85}
    \begin{tiny}

    % Reduce column spacing
    % Default is 6pt. 2pt is usually tight enough to fit 12 cols.
    \setlength{\tabcolsep}{2.5pt}

    \makeatletter
    \newcommand{\tightscripts}{%
      \fontdimen14\textfont2=1.4pt   % superscript shift up
      \fontdimen16\textfont2=1.0pt   % subscript shift down
      \fontdimen17\textfont2=1.0pt   % subscript shift down (with superscript)
      \fontdimen8\textfont3=0.12pt   % min sup/sub gap = 4x this
    }
    \makeatother

    \newcommand{\res}[3]{\ensuremath{\tightscripts #1^{+#2}_{-#3}}}

    \setlength{\aboverulesep}{0.3ex}
    \setlength{\belowrulesep}{0.3ex}

    \begin{tabular}{ll ccccc ccccc}
        \toprule

        & & \multicolumn{5}{c}{Success Rate (\%) ($\uparrow$)} & \multicolumn{5}{c}{Average Episode Length on Success ($\downarrow$)} \\
        \cmidrule(lr){3-7} \cmidrule(lr){8-12}

        & $\varphi$ & Discretize & Concatenate & SIAMS & Plato & Plato + P\textsuperscript{2}BRS & Discretize & Concatenate & SIAMS & Plato & Plato + P\textsuperscript{2}BRS \\
        \midrule

        % ================= Go1RGBZoneEnv =================
        \multirow{8}{*}[-4pt]{\rotatebox[origin=c]{90}{Go1RGBZoneEnv}}
        % varphi_1 = F (_at_0_ & (!_at_1_ U _at_2_)) & F _at_3_
        & $\varphi_{1}$
        & \res{72.4}{2.2}{2.0} & \res{97.9}{0.5}{0.7} & \res{96.4}{1.0}{1.4} & \res{99.0}{0.2}{0.3} & \res{\mathbf{99.1}}{0.3}{0.3}
        & \res{441.4}{28.3}{23.4} & \res{178.9}{11.5}{9.6} & \res{242.5}{32.2}{28.7} & \res{117.9}{4.0}{3.7} & \res{\mathbf{116.3}}{4.1}{2.2} \\

        % varphi_2 = (F _at_0_) & (!_at_0_ U (_at_1_ & F _at_2_))
        & $\varphi_{2}$
        & \res{53.6}{2.4}{3.4} & \res{94.8}{0.6}{0.8} & \res{90.2}{2.0}{3.8} & \res{\mathbf{98.4}}{0.2}{0.2} & \res{\mathbf{98.4}}{0.3}{0.2}
        & \res{313.4}{14.5}{13.1} & \res{175.1}{9.5}{8.0} & \res{197.3}{19.0}{15.6} & \res{\mathbf{112.5}}{2.7}{2.4} & \res{114.2}{2.6}{2.4} \\

        % varphi_3 = F (_at_0_ | _at_1_) & F _at_2_ & F _at_3_
        & $\varphi_{3}$
        & \res{97.9}{0.5}{0.5} & \res{99.2}{0.3}{0.7} & \res{98.8}{0.3}{0.3} & \res{99.0}{0.2}{0.2} & \res{\mathbf{99.3}}{0.2}{0.2}
        & \res{161.7}{6.3}{9.3} & \res{128.6}{5.1}{3.9} & \res{143.9}{11.5}{7.7} & \res{\mathbf{97.2}}{3.2}{2.5} & \res{101.4}{2.3}{2.3} \\

        % varphi_4 = !(_at_0_ | _at_1_) U (_at_2_ & F _at_3_)
        & $\varphi_{4}$
        & \res{35.9}{1.5}{2.7} & \res{86.1}{2.0}{1.4} & \res{78.3}{3.8}{4.6} & \res{95.6}{0.5}{0.4} & \res{\mathbf{96.5}}{0.5}{0.6}
        & \res{203.2}{16.2}{14.1} & \res{135.3}{11.5}{8.1} & \res{150.9}{21.0}{14.5} & \res{81.2}{3.0}{2.3} & \res{\mathbf{80.4}}{3.6}{1.9} \\

        % varphi_5 = !_at_0_ U ((_at_1_ | _at_2_) & (!_at_0_ U _at_3_))
        & $\varphi_{5}$
        & \res{36.1}{2.1}{2.6} & \res{92.7}{0.9}{1.2} & \res{86.8}{1.3}{1.9} & \res{97.3}{0.3}{0.3} & \res{\mathbf{97.6}}{0.4}{0.4}
        & \res{288.9}{27.0}{26.6} & \res{126.5}{16.6}{10.9} & \res{150.8}{23.2}{18.7} & \res{75.2}{4.0}{3.6} & \res{\mathbf{74.7}}{3.3}{1.8} \\

        % varphi_6 = ((_at_0_ | _at_1_) => (!_at_2_ U _at_3_)) U _at_2_
        & $\varphi_{6}$
        & \res{50.1}{2.8}{4.2} & \res{97.2}{0.8}{1.0} & \res{93.3}{1.9}{2.5} & \res{\mathbf{99.7}}{0.2}{0.2} & \res{99.6}{0.1}{0.1}
        & \res{231.2}{17.6}{18.8} & \res{95.1}{9.7}{10.0} & \res{123.2}{18.4}{16.5} & \res{46.3}{2.6}{2.1} & \res{\mathbf{44.1}}{2.4}{1.5} \\

        % varphi_7 = F (_at_0_ & X F (_at_1_ & X F (_at_2_ & X F (_at_3_ & X F (_at_2_ & X F(_at_0_ & X F (_at_3_ & X F (_at_1_ & X F (_at_3_ & X F(_at_1_ & X F (_at_2_ & X F _at_0_)))))))))))
        & $\varphi_{7}$
        & \res{84.1}{1.5}{1.4} & \res{91.9}{1.2}{1.1} & \res{68.9}{10.9}{14.9} & \res{95.0}{0.5}{0.9} & \res{\mathbf{95.7}}{0.5}{0.5}
        & \res{1053.8}{59.0}{48.2} & \res{770.0}{24.8}{30.0} & \res{1132.8}{117.1}{121.3} & \res{451.8}{17.9}{12.6} & \res{\mathbf{442.0}}{12.6}{8.6} \\

        % varphi_8 = (! _at_0_ U (_at_1_ & (! _at_2_ U (_at_0_ & (! _at_3_ U (_at_2_ & (!_at_1_ U (_at_3_ & (! _at_2_ U (_at_1_ & (! _at_0_ U (_at_2_ & (! _at_3_ U (_at_0_ & (!_at_1_ U _at_3_)))))))))))))))
        & $\varphi_{8}$
        & \res{3.1}{1.1}{1.3} & \res{83.0}{2.8}{1.9} & \res{64.1}{8.1}{13.3} & \res{94.9}{0.4}{0.4} & \res{\mathbf{95.5}}{0.7}{1.0}
        & N/A & \res{649.6}{64.1}{42.3} & \res{912.7}{103.5}{97.0} & \res{352.6}{14.3}{13.4} & \res{\mathbf{343.4}}{16.7}{7.7} \\

        \midrule

        % ================= X2DronesEnv =================
        \multirow{8}{*}[-4pt]{\rotatebox[origin=c]{90}{X2DronesEnv}}
        % varphi_9 = !_pos_0_ U (_pos_1_ | (_pos_2_ & _pos_3_))
        & $\varphi_{9}$
        & \res{26.8}{4.1}{2.6} & \res{22.9}{2.1}{3.0} & \res{34.3}{3.5}{2.5} & \res{91.3}{2.7}{6.9} & \res{\mathbf{96.2}}{0.4}{0.4}
        & \res{554.8}{35.1}{31.9} & \res{221.8}{17.5}{21.7} & \res{356.0}{14.0}{16.7} & \res{133.7}{38.9}{24.4} & \res{\mathbf{99.6}}{5.5}{4.5} \\

        % varphi_10 = F _pos_0_ & (!(_pos_1_ | _pos_2_) U _pos_3_)
        & $\varphi_{10}$
        & \res{8.2}{3.2}{2.3} & \res{8.8}{1.0}{1.0} & \res{24.4}{2.4}{1.9} & \res{85.2}{3.0}{5.8} & \res{\mathbf{89.5}}{1.2}{0.7}
        & \res{868.1}{76.1}{78.9} & \res{281.2}{46.5}{24.4} & \res{542.8}{11.7}{15.4} & \res{191.1}{43.8}{22.1} & \res{\mathbf{157.5}}{4.9}{4.3} \\

        % varphi_11 = ((_pos_0_ | _pos_1_ | _pos_2_) => F (_pos_3_ & _pos_4_)) U (_pos_2_ & _pos_3_)
        & $\varphi_{11}$
        & \res{0.1}{0.2}{0.1} & \res{0.7}{1.0}{0.4} & \res{1.7}{0.7}{0.4} & \res{60.8}{7.2}{8.8} & \res{\mathbf{93.0}}{0.4}{0.5}
        & N/A & N/A & N/A & \res{452.6}{34.7}{30.5} & \res{\mathbf{299.2}}{18.0}{16.6} \\

        % varphi_12 = F (_pos_0_ & F ((_pos_1_ & _pos_2_) & F ((_pos_2_ & _pos_3_) & (!(_pos_0_ | _pos_1_) U (_pos_3_ & _pos_4_)))))
        & $\varphi_{12}$
        & \res{0.0}{0.0}{0.0} & \res{0.0}{0.0}{0.0} & \res{0.0}{0.0}{0.0} & \res{37.5}{9.6}{10.8} & \res{\mathbf{85.9}}{1.0}{1.2}
        & N/A & N/A & N/A & \res{828.1}{68.2}{59.3} & \res{\mathbf{571.4}}{18.5}{20.0} \\

        % varphi_13 = !(_pos_0_ | _pos_1_ | (_pos_2_ & _pos_3_)) U ((_pos_3_ & _pos_4_) & F (_pos_4_ & _pos_5_))
        & $\varphi_{13}$
        & \res{0.0}{0.0}{0.0} & \res{0.0}{0.1}{0.0} & \res{0.2}{0.2}{0.1} & \res{45.4}{9.3}{11.7} & \res{\mathbf{84.4}}{1.0}{1.0}
        & N/A & N/A & N/A & \res{606.4}{57.4}{44.4} & \res{\mathbf{364.7}}{10.1}{14.7} \\

        % varphi_14 = F (_pos_0_ & F (_pos_1_ & F (_pos_2_ & F (_pos_3_ & F (_pos_4_ & F (_pos_0_ & F (_pos_2_ & F _pos_4_)))))))
        & $\varphi_{14}$
        & \res{0.0}{0.0}{0.0} & \res{0.0}{0.0}{0.0} & \res{0.0}{0.1}{0.0} & \res{94.3}{1.4}{1.7} & \res{\mathbf{95.3}}{1.1}{1.0}
        & N/A & N/A & N/A & \res{543.2}{12.0}{13.8} & \res{\mathbf{492.9}}{11.7}{13.0} \\

        % varphi_15 = (!_pos_0_ U (_pos_1_ & F _pos_2_)) | (!_pos_0_ U (_pos_3_ & F _pos_4_))
        & $\varphi_{15}$
        & \res{10.9}{4.0}{3.1} & \res{16.4}{1.2}{1.9} & \res{37.4}{1.7}{1.2} & \res{\mathbf{90.6}}{1.4}{1.6} & \res{90.0}{1.5}{1.4}
        & \res{919.0}{50.9}{50.1} & \res{257.0}{26.7}{19.2} & \res{592.7}{23.1}{27.1} & \res{202.3}{19.8}{14.8} & \res{\mathbf{171.4}}{7.4}{6.0} \\

        % varphi_16 = (!(_pos_0_ | _pos_1_) U _pos_2_) & F (_pos_0_ & _pos_1_)
        & $\varphi_{16}$
        & \res{0.1}{0.1}{0.1} & \res{0.7}{0.2}{0.2} & \res{2.1}{0.7}{0.6} & \res{68.5}{6.1}{8.2} & \res{\mathbf{90.0}}{1.0}{1.2}
        & N/A & N/A & N/A & \res{385.3}{54.7}{42.0} & \res{\mathbf{247.3}}{8.5}{9.1} \\

        \midrule

        % ================= FrankaCubesEnv =================
        \multirow{8}{*}[-4pt]{\rotatebox[origin=c]{90}{FrankaCubesEnv}}
        % varphi_17 = F (_pos_0_ & F (_rot_0_ & F (_pos_1_ & F (_pos_2_ & F (_rot_1_ & F (_pos_3_ & F (_pos_0_ & F (_rot_0_ & F _pos_2_))))))))
        & $\varphi_{17}$
        & \res{0.0}{0.1}{0.0} & \res{0.0}{0.0}{0.0} & \res{0.0}{0.0}{0.0} & \res{14.4}{15.2}{5.6} & \res{\mathbf{97.1}}{0.5}{0.4}
        & N/A & N/A & N/A & \res{939.9}{57.9}{48.5} & \res{\mathbf{255.7}}{15.5}{9.8} \\

        % varphi_18 = !_pos_0_ U (_pos_1_ & _has_0_)
        & $\varphi_{18}$
        & \res{0.0}{0.1}{0.0} & \res{0.0}{0.1}{0.0} & \res{0.0}{0.0}{0.0} & \res{3.0}{5.9}{3.0} & \res{\mathbf{83.2}}{1.3}{1.1}
        & N/A & N/A & N/A & N/A & \res{\mathbf{95.7}}{7.2}{7.4} \\

        % varphi_19 = F (_has_0_ & F ((_pos_0_ & _has_0_) & F (_pos_1_ & _rot_0_ & _has_0_)))
        & $\varphi_{19}$
        & \res{0.0}{0.0}{0.0} & \res{0.0}{0.0}{0.0} & \res{0.0}{0.0}{0.0} & \res{0.0}{0.0}{0.0} & \res{\mathbf{85.7}}{1.7}{1.7}
        & N/A & N/A & N/A & N/A & \res{\mathbf{255.9}}{24.6}{17.6} \\

        % varphi_20 = F ((_has_0_ & _rot_0_) & F ((_has_0_ & _rot_1_) & F (_has_0_ & (_pos_0_ | _pos_1_))))
        & $\varphi_{20}$
        & \res{0.0}{0.0}{0.0} & \res{0.0}{0.0}{0.0} & \res{0.0}{0.0}{0.0} & \res{0.0}{0.0}{0.0} & \res{\mathbf{85.0}}{2.3}{2.0}
        & N/A & N/A & N/A & N/A & \res{\mathbf{299.1}}{23.3}{19.9} \\

        % varphi_21 = !_pos_0_ U (_has_0_ & (!_pos_1_ U (_pos_2_ & _has_0_)))
        & $\varphi_{21}$
        & \res{0.0}{0.0}{0.0} & \res{0.0}{0.0}{0.0} & \res{0.0}{0.0}{0.0} & \res{3.2}{7.7}{3.2} & \res{\mathbf{86.5}}{1.5}{1.0}
        & N/A & N/A & N/A & N/A & \res{\mathbf{97.1}}{7.5}{7.0} \\

        % varphi_22 = F ((_has_0_ & _pos_0_) & (_pos_0_ U (!_has_0_ & _pos_0_))) & F ((_has_1_ & _pos_1_) & (_pos_1_ U (!_has_1_ & _pos_1_)))
        & $\varphi_{22}$
        & \res{0.0}{0.0}{0.0} & \res{0.0}{0.0}{0.0} & \res{0.0}{0.0}{0.0} & \res{0.1}{0.3}{0.1} & \res{\mathbf{75.7}}{4.6}{5.8}
        & N/A & N/A & N/A & N/A & \res{\mathbf{390.8}}{49.7}{38.8} \\

        % varphi_23 = F ((_has_0_ & _pos_0_) & (_pos_0_ U ((!_has_0_ & _pos_0_) & F ((_has_0_ & _pos_1_) & (_pos_1_ U (!_has_0_ & _pos_1_))))))
        & $\varphi_{23}$
        & \res{0.0}{0.0}{0.0} & \res{0.0}{0.0}{0.0} & \res{0.0}{0.0}{0.0} & \res{0.1}{0.4}{0.1} & \res{\mathbf{79.5}}{3.3}{5.3}
        & N/A & N/A & N/A & N/A & \res{\mathbf{326.3}}{29.9}{37.9} \\

        % varphi_24 = F ((_has_0_ & (_pos_0_ | _pos_1_)) & ((_pos_0_ | _pos_1_) U (!_has_0_ & (_pos_0_ | _pos_1_))))
        & $\varphi_{24}$
        & \res{0.0}{0.1}{0.0} & \res{0.0}{0.1}{0.0} & \res{0.0}{0.0}{0.0} & \res{2.6}{7.3}{2.4} & \res{\mathbf{81.9}}{3.6}{6.9}
        & N/A & N/A & N/A & N/A & \res{\mathbf{268.1}}{47.2}{27.7} \\

        \midrule

        % ================= FalloutWorld =================
        \multirow{8}{*}[-4pt]{\rotatebox[origin=c]{90}{FalloutWorld}}
        % varphi_25 = !(_rad_0_ | _loc_0_) U ((_loc_1_ & !_rad_0_) & F _loc_2_)
        & $\varphi_{25}$
        & \res{74.1}{1.4}{1.1} & \res{58.8}{2.4}{2.6} & \res{19.7}{3.0}{1.7} & \res{90.2}{0.7}{1.8} & \res{\mathbf{91.0}}{0.5}{0.6}
        & \res{80.7}{5.3}{4.7} & \res{288.6}{32.0}{23.6} & \res{594.5}{27.6}{41.9} & \res{\mathbf{62.4}}{3.8}{4.3} & \res{66.2}{5.9}{2.4} \\

        % varphi_26 = !_rad_0_ U ((_loc_0_ | _loc_1_) & (!_rad_0_ U (_loc_2_ & !_rad_0_)))
        & $\varphi_{26}$
        & \res{71.2}{1.9}{1.5} & \res{51.7}{2.2}{2.3} & \res{24.4}{1.6}{1.1} & \res{\mathbf{85.5}}{0.7}{1.1} & \res{85.4}{0.5}{0.6}
        & \res{131.9}{9.9}{10.0} & \res{307.8}{31.7}{21.5} & \res{608.9}{25.2}{27.7} & \res{\mathbf{81.2}}{5.9}{5.4} & \res{82.0}{8.0}{6.4} \\

        % varphi_27 = (_rad_0_ => (! _loc_0_ U (_loc_1_ | _loc_2_))) U (_loc_0_ & !_rad_0_)
        & $\varphi_{27}$
        & \res{83.3}{1.3}{1.2} & \res{46.2}{3.3}{3.0} & \res{24.1}{2.6}{2.3} & \res{\mathbf{91.5}}{1.2}{1.6} & \res{90.8}{1.9}{2.6}
        & \res{103.3}{7.3}{8.6} & \res{392.5}{29.4}{20.8} & \res{564.7}{25.4}{26.7} & \res{\mathbf{75.8}}{8.5}{7.7} & \res{79.7}{11.8}{10.6} \\

        % varphi_28 = (!_rad_0_ U (_loc_0_ & !_rad_0_)) & F (_rad_0_ & F _loc_1_)
        & $\varphi_{28}$
        & \res{82.3}{1.0}{0.8} & \res{63.2}{2.2}{2.3} & \res{32.9}{2.5}{2.1} & \res{91.8}{0.4}{0.6} & \res{\mathbf{92.3}}{0.5}{0.6}
        & \res{92.7}{7.2}{4.9} & \res{249.6}{24.2}{15.6} & \res{492.7}{21.3}{18.9} & \res{\mathbf{57.0}}{3.2}{2.4} & \res{59.8}{3.8}{2.7} \\

        % varphi_29 = F _loc_0_ & F _loc_1_ & (!_rad_0_ U (_loc_2_ & !_rad_0_))
        & $\varphi_{29}$
        & \res{76.8}{1.0}{1.0} & \res{48.1}{2.2}{1.9} & \res{24.0}{2.1}{1.7} & \res{\mathbf{89.7}}{1.0}{0.7} & \res{89.0}{0.6}{0.6}
        & \res{121.1}{6.1}{8.1} & \res{432.8}{21.0}{26.2} & \res{783.5}{14.4}{21.4} & \res{\mathbf{79.6}}{2.8}{2.2} & \res{81.8}{8.6}{4.6} \\

        % varphi_30 = (!(_loc_0_ | _loc_1_) U _rad_0_) & F (_loc_0_ & F _loc_1_)
        & $\varphi_{30}$
        & \res{\mathbf{99.6}}{0.1}{0.3} & \res{98.5}{0.4}{0.6} & \res{70.1}{7.3}{7.4} & \res{\mathbf{99.6}}{0.1}{0.1} & \res{99.5}{0.2}{0.3}
        & \res{40.0}{0.7}{0.6} & \res{94.7}{12.3}{10.5} & \res{371.8}{25.6}{24.3} & \res{\mathbf{35.9}}{0.4}{0.4} & \res{36.7}{0.4}{0.5} \\

        % varphi_31 = F (_loc_0_ & F (_loc_1_ & F (_loc_2_ & F (_rad_0_ & F (_loc_3_ & F (_loc_4_ & F _loc_5_))))))
        & $\varphi_{31}$
        & \res{99.7}{0.1}{0.2} & \res{73.5}{6.2}{9.5} & \res{1.0}{0.8}{0.5} & \res{\mathbf{99.9}}{0.1}{0.1} & \res{99.8}{0.1}{0.1}
        & \res{116.0}{2.4}{2.5} & \res{611.4}{54.1}{43.7} & N/A & \res{\mathbf{106.0}}{1.2}{1.0} & \res{108.0}{1.1}{1.2} \\

        % varphi_32 = ((!_rad_0_ U (_loc_0_ & !_rad_0_)) & F _loc_1_) | ((!_rad_0_ U (_loc_2_ & !_rad_0_)) & F _loc_3_)
        & $\varphi_{32}$
        & \res{78.8}{1.2}{1.4} & \res{63.9}{1.3}{1.8} & \res{43.5}{1.7}{1.4} & \res{\mathbf{92.5}}{0.9}{1.0} & \res{91.6}{0.4}{0.4}
        & \res{92.6}{10.2}{8.5} & \res{255.6}{20.2}{14.3} & \res{493.7}{24.6}{24.2} & \res{\mathbf{63.5}}{3.8}{5.0} & \res{64.0}{3.1}{3.2} \\

        \bottomrule
    \end{tabular}
    \end{tiny}
    \end{center}
    \vskip -0.1in
\end{table*}

\textbf{Q1.} Figure~\ref{fig:eval_curves} and Table~\ref{tab:results_finite} show that Plato quickly achieves high success rates and low episode lengths in \textit{Go1RGBZoneEnv} ($94.9$--$99.7\%$) and \textit{FalloutWorld} ($85.5$--$99.9\%$), while the baselines generally perform substantially worse. Plato also makes steady learning progress in \textit{X2DronesEnv}, while the baselines remain at low success rates. This is because Plato provides a strong \textit{inductive bias} by embedding predicate parameters in the goal structure, whereas Concatenate and SIAMS must learn a complex association between propositions in the sequence embedding and parameters in the state embedding, and Discretize must learn each embedding from scratch for its hundreds to thousands of bins. Discretize is most competitive in \textit{FalloutWorld}, where its bins coincide with grid cells.
%Note also that these baselines assume a maximum number of unique propositions per specification, restricting the set of admissable specifications at evaluation time; meanwhile, Plato makes no such assumptions.

\textbf{Q2.} We see that P\textsuperscript{2}BRS markedly accelerates learning in \textit{X2DronesEnv}, with success rates converging to $84.4$--$96.2\%$ while unguided Plato is still far from convergence. The impact of P\textsuperscript{2}BRS is even starker for \textit{FrankaCubesEnv}, taking Plato's success rates from near zero to $75.7$--$97.1\%$. This is to be expected, since the learning signal is extremely sparse in these high-dimensional robotics environments, especially when learning conjunctions. In \textit{X2DronesEnv}, satisfying $\mathsf{pos}_i \land \mathsf{pos}_j$ requires each drone to enter specific regions in a large 3D space, which is an exceedingly rare occurrence. In \textit{FrankaCubesEnv}, satisfying $\mathsf{has}_i \land \mathsf{pos}_j \land \mathsf{rot}_k$ requires the arm to pick up the correct cube and bring it to a specific position and orientation, which is virtually impossible under unguided exploration. Thus, a dense guiding signal is essential for effective exploration in these environments. Meanwhile, we see little to no benefit in \textit{Go1RGBZoneEnv} and \textit{FalloutWorld}; this suggests that exploration is not the learning bottleneck for these simpler environments with the StructLTL backbone.

\textbf{Ablations.} Appendix~\ref{app:ablation_studies:pca_prop_embeddings} uses principal component analysis to demonstrate that Plato learns low-dimensional predicate embedding geometries. Appendix~\ref{app:ablation_studies:ltl2action} applies PlatoLTL to LTL2Action, a second backbone. Appendix~\ref{app:ablation_studies:p2brs} ablates P\textsuperscript{2}BRS, validating the need for adaptive robustness scale calibration and persistence gating, and demonstrating that automaton-based PBRS is insufficient.

\textbf{Limitations.} The benefits of PlatoLTL scale with the task structure its inputs expose. The Plato embedding module relies on propositions being represented as instances of predicates known a priori; while such a parameterization always exists, its benefit requires the labeling function to vary structurally with the parameters, and in the absence of such structure we would not expect gains over the unaugmented backbone. Similarly, the guidance of P\textsuperscript{2}BRS is only as informative as the supplied robustness metrics; a valid metric always exists and shaping never biases the optimal policy, but informative metrics may be challenging to hand-specify for semantically complex predicates. Such predicates compress into a single atom the structure that a more granular specification would expose with readily specified metrics. Under weak or trivial metrics, or overly coarse specifications, P\textsuperscript{2}BRS reverts to unguided exploration for the affected predicates. See Appendix~\ref{app:limitations} for further discussion.

%%%%%%%%%%%%%%%%%%%%%%%%%%%%%%%%
% CONCLUSION
%%%%%%%%%%%%%%%%%%%%%%%%%%%%%%%%

% \section{Conclusion}
% \label{sec:conclusion}

\textbf{Conclusion.} The Plato embedding module enhances representation learning for parameterized specifications, while P\textsuperscript{2}BRS accelerates exploration in high-dimensional robotics environments, lifting success rates from near zero to $75.7$--$97.1\%$ in \textit{FrankaCubesEnv}. Future work could investigate using uncertain and learned labeling functions in a step towards enabling real-world robotic deployment, and pre-training robustness metrics from demonstrations; see Appendix~\ref{app:future_work} for further discussion.

\subsection*{AI use statement}

In this work, we used generative AI tools to assist with implementing methods. We have not used generative AI tools to help develop theoretical models or conceptual frameworks, formulate mathematical claims, provide critical ingredients for or assist in writing proofs, propose or refine hypotheses, design or provide feedback on research methodology or experiments, or interpret results. Generating synthetic datasets, translation, cleaning or reformatting datasets, and qualitative or thematic data analysis are not applicable to this work. Additionally, we used generative AI tools to create and edit software code (including plotting, evaluation, and tooling scripts), and to check for typos and suggest rewording to improve readability. We have reviewed all AI-assisted work. AI-assisted code was checked by the authors and tested with unit tests, and suggested rewordings were reviewed and refined by the authors before adoption. We take responsibility for the final content of this work, including text, claims, or artifacts produced with the aid of generative AI.

\subsection*{Reproducibility statement}

We have made several efforts to ensure the reproducibility of this work.
% Source code for PlatoLTL, including implementations of all four environments, the pre-training pipeline for the Go1 locomotion controller, all baselines, the training and evaluation pipelines, and all ablation studies, is available via an anonymous repository linked here: \url{https://anonymous.4open.science/r/plato-ltl-iclr-2027-anonymous-FC6E}.
The problem setting and its assumptions are stated in Section~\ref{sec:problem_setting}, and the Plato embedding module and P\textsuperscript{2}BRS are specified in full in Sections~\ref{sec:plato} and~\ref{sec:p2brs}, respectively.  Appendix~\ref{app:pbrs_proof} proves that P\textsuperscript{2}BRS leaves the optimal policy unchanged, and Appendix~\ref{app:fo-ltl} justifies the equivalence of our parameterized specifications to LTL. Appendices~\ref{app:struct_ltl} and~\ref{app:reach_avoid_sequence_extraction} detail the StructLTL backbone and the reach-avoid sequence extraction procedure used by both the backbone and P\textsuperscript{2}BRS, and Appendix~\ref{app:ablation_studies:ltl2action} describes how PlatoLTL is applied to LTL2Action. Appendix~\ref{app:experimental_details:implementation} lists the software used in our implementation and the hardware used for running experiments. Appendix~\ref{app:experimental_details:environments} fully specifies each environment, including its dynamics, observation and action spaces, low-level controllers (and the pre-training of the Go1 locomotion controller), reset distributions, predicates, labeling functions, robustness metrics, and persistence annotations. The template LTL specifications used for evaluation are listed in Table~\ref{tab:formulae} (Appendix~\ref{app:experimental_details:eval_ltl_specifications}). Network architectures for the Plato embedding module, all baselines, and both backbones are given in Appendix~\ref{app:experimental_details:training_details:networks}, and all PPO and P\textsuperscript{2}BRS hyperparameters are listed in Table~\ref{tab:ppo_params}. The training curriculum is detailed in Appendix~\ref{app:experimental_details:training_details:curriculum} and Table~\ref{tab:curriculum}, and the configurations for each ablation are described in Appendix~\ref{app:ablation_studies}. Finally, Section~\ref{sec:experiments} describes the evaluation protocol, including the number of seeds, evaluation episodes, and evaluation horizon, as well as the construction of confidence intervals.


\subsubsection*{Acknowledgments}
This work was supported by the EPSRC Centre for Doctoral Training in Autonomous Intelligent Machines and Systems [EP/S024050/1].

\clearpage

\bibliography{references}
\bibliographystyle{iclr2027_conference}

%%%%%%%%%%%%%%%%%%%%%%%%%%%%%%%%%%%%%%%%%%%%%%%%%%%%%%%%%%%%%%%%%%%%%%%%%%%%%%%
%%%%%%%%%%%%%%%%%%%%%%%%%%%%%%%%%%%%%%%%%%%%%%%%%%%%%%%%%%%%%%%%%%%%%%%%%%%%%%%
% APPENDIX
%%%%%%%%%%%%%%%%%%%%%%%%%%%%%%%%%%%%%%%%%%%%%%%%%%%%%%%%%%%%%%%%%%%%%%%%%%%%%%%
%%%%%%%%%%%%%%%%%%%%%%%%%%%%%%%%%%%%%%%%%%%%%%%%%%%%%%%%%%%%%%%%%%%%%%%%%%%%%%%

\newpage
\appendix

\section{Extended Related Work}
\label{app:extended_related_work}

\textbf{LTL-Guided RL.} Methods for learning LTL-guided policies can generally be classified into two approaches. The first approach converts the LTL formula into an automaton and learns a policy conditioned directly on the automaton's state or transitions \citep{icarte_using_2018, icarte_reward_2022, camacho_ltl_2019, yuan_modular_2019, hahn_omega-regular_2019, bozkurt_control_2020, bertrand_deep_2020, hasanbeig_certified_2023, cai_modular_2021, voloshin_eventual_2023, shah_ltl-constrained_2025, fan_imitation_2025, li_ground-compose-reinforce_2025}, either as a single policy or a modular set of sub-policies. However, such methods generally produce \textit{single-task} policies, since the automaton states and transitions correspond only to a particular specification. In contrast, the methods we refer to as \textit{backbones} condition the policy on a learned representation of the task as a goal, with the goal space shared across specifications \citep{vaezipoor_ltl2action_2021, leon_nutshell_2022, qiu_instructing_2023, zhang_exploiting_2023, zhang_exploiting_2024, yalcinkaya_compositional_2024, xu_generalization_2024, jackermeier_deepltl_2025, guo_one_2025, abate_semantically_2026, jackermeier_zero-shot_2026}. By training over a sufficiently wide distribution of such representations, one obtains \textit{multi-task} policies.

For example, LTL2Action \citep{vaezipoor_ltl2action_2021} encodes the syntax tree of the task LTL specification using a graph neural network (GNN; \citealp{zhou_graph_2020}), updating the specification via LTL progression as the agent progresses through the task. Meanwhile, DeepLTL \citep{jackermeier_deepltl_2025} extracts a reach-avoid sequence of sets of assignments for LTL satisfaction from its corresponding automaton, and uses a recurrent neural network (RNN; \citealp{sherstinsky_fundamentals_2020}) to embed the sequence, updating it as the agent satisfies each subgoal. StructLTL \citep{jackermeier_zero-shot_2026} improves upon this by representing sets of assignments as Boolean formulae, and learns \textit{structured} task representations to better generalize across particularly complex specifications.

\textbf{Learning Representations for Parameterized Specifications.} Backbones typically compose goal embeddings from a fixed, discrete vocabulary of propositions known a priori. This presents a challenge when faced with \textit{parameterized} specifications as is commonplace in high-dimensional environments such as robotics, since na\"{i}vely binning continuous parameters into discrete propositions scales poorly with resolution and parameter dimensions. The other na\"{i}ve solution is to append the parameters to the observation space; such an approach works well for single-task policies, i.e., those generalizing only over the parameters of a fixed formula structure \citep{jothimurugan_composable_2019, jothimurugan_compositional_2021, shukla_logical_2024}, but degrades in multi-task settings, since the policy must learn a complex association between propositions in the goal embedding and parameters in the state embedding. \cite{hatanaka_reinforcement_2025} go a step further and use Feature-wise Linear Modulation (FiLM; \citealp{perez_film_2018}) to transform the backbone's state embedding using the parameters, but the method suffers the same fundamental scaling issues once more than one proposition may be relevant at a given time. In comparison, our work models propositions as parameterized instances of \textit{atomic predicates} and injects the parameters into the goal embedding, which yields low-dimensional per-predicate embedding geometries and enables efficient generalization across parameters.

\citet{guo_one_2025} demonstrate generalization to unseen propositions (and thus new parameters) by remapping the observation space into universal ``reach'' and ``avoid'' observations. However, this assumes that the observation space can be partitioned into proposition-specific observations, produced by identical observation functions under output transformation with a shared structure amenable to fusion, which is generally not true (or requires significant environment-specific engineering) for high-dimensional robotics environments. For example, if a manipulator arm's observations include a table of the IDs and locations of all objects in the scene, it is not clear how to fuse conjunctions and disjunctions of propositions related to these objects into ``reach'' and ``avoid'' observations, even if the table can be partitioned into per-object rows. Our work makes no such assumptions.

\textbf{Exploration for LTL-Guided RL.} Another key challenge in LTL-guided RL is \textit{exploration efficiency}, since the reward function is typically \textit{sparse}. While a training curriculum \citep{jackermeier_deepltl_2025} or hindsight relabeling \citep{qiu_instructing_2023} can speed up learning to some extent, they do not provide explicit guidance towards states that satisfy the current subgoal, meaning that for complex subgoals (such as conjunctions of propositions) in high-dimensional environments, visits of those states can be impractically rare.

Some methods define a potential function over automaton states to guide the learning of multi-step tasks without biasing the optimal policy \citep{camacho_ltl_2019, icarte_reward_2022, bagatella_directed_2025}. While effective when individual subgoals are simple enough to be discovered without guidance, automaton-level potentials are by construction constant in the environment state, so the shaped reward differentiates progress only across (sparse) automaton transitions and provides no gradient \textit{within} a subgoal; it thus does not address complex subgoals in high-dimensional environments.

Like our approach, Ground-Compose-Reinforce (GCR; \citealp{li_ground-compose-reinforce_2025}) constructs potentials that vary with environment state for automaton-structured tasks (reward machines; \citealp{icarte_reward_2022}), rewriting each transition's formula in DNF and composing per-literal progress signals via a $\min$ over literals and a $\max$ over clauses. Its progress signals are value functions for reaching each proposition or its negation, learned via offline RL (alongside a labeling function) from a dataset of labeled trajectories, and combined with automaton-level values obtained via value iteration. Our work differs in several respects. First, GCR addresses a single-task setting, exploiting compositionality in the construction of rewards and potentials rather than in the policy; we instead shape the training of multi-task backbones that generalize zero-shot across specifications. Second, GCR assumes a fixed propositional vocabulary with a pair of value functions per proposition, whereas our metrics are defined per parameterized predicate. Third, GCR requires a \textit{dataset of labeled trajectories} with adequate coverage of the state space (e.g., spanning the parameter spaces of all predicates), which must be collected for each environment; our robustness metrics are simple priors requiring no pre-training data. Fourth, a plain $\min$ over literals provides no guidance on \textit{implicit} temporal orderings within a conjunction, and can also overestimate conjunctive values \citep{li_ground-compose-reinforce_2025}; our \textit{persistence gating} addresses this and proves essential for heterogeneous conjunctions (Appendix~\ref{app:ablation_studies:p2brs:persistence_gating}). Finally, GCR targets reward machine returns, and its authors report best results with an undiscounted shaping term that forgoes the invariance guarantee of PBRS; P\textsuperscript{2}BRS retains this guarantee (Theorem~\ref{thm:pbrs}) and targets LTL satisfaction, including \textit{persistence} ($\F \G$) tasks, via LDBAs. The two approaches are \textit{complementary}, as GCR's learned value functions could serve as robustness metrics for P\textsuperscript{2}BRS (Appendix~\ref{app:future_work}).

\cite{cai_overcoming_2023} employ a path-planning-guided dense reward scheme for single-task policies, but the planner-based guidance is geometric and does not transfer to multi-task policies. In contrast, our work provides \textit{dense} guidance in multi-task settings without biasing the optimal policy.

Other works propose using alternative specification languages with \textit{quantitative semantics} to densify the satisfaction signal; the most well-known is signal temporal logic (STL; \citealp{maler_monitoring_2004, donze_robust_2010}), with others including truncated LTL (TLTL; \citealp{li_reinforcement_2017}) and SPECTRL \citep{jothimurugan_composable_2019}. Such languages require \textit{accurate} per-predicate robustness metrics since task satisfaction is reasoned about directly from the quantitative semantics rather than Boolean labeling functions, and policy optimization often seeks to maximize the expected task robustness value, which is not equivalent to maximizing task satisfaction probability (thus biasing the policy away from it). Moreover, the difference in semantics means that these languages are incompatible with LTL-based backbones, and the bespoke RL-based methods for solving them typically learn single-task policies. Meanwhile, our robustness metrics act as \textit{simple priors} that represent coarse domain knowledge to guide learning, and the tasks we solve remain fundamentally LTL, ensuring compatibility with (multi-task) LTL-based backbones.

% Recently, Transformers have been used to generate task embeddings from LTL formulae \cite{zhang_exploiting_2023, zhang_exploiting_2024}, though it has been shown that, in model-free RL settings, Transformers often struggle to learn the complex interactions needed for LTL specifications \cite{giuri_zero-shot_2025}.

%%%%%%%%%%%%%%%%%%%%%%%%%%%%%%%%
% LTL SATISFACTION SEMANTICS
%%%%%%%%%%%%%%%%%%%%%%%%%%%%%%%%

\section{LTL Satisfaction Semantics}
\label{app:ltl_satisfaction_semantics}

Let $w[i] = \sigma_i$ and $w[i \dots] = (\sigma_i, \sigma_{i+1}, \sigma_{i+2}, \dots)$. The satisfaction relation $w \models \varphi$ is recursively defined as follows \citep{baier_principles_2008}:
\begin{align*}
    w & \models \true & & \\
    w & \models \mathsf{p} & & \iff \mathsf{p} \in w[0] \\
    w & \models \lnot \varphi & & \iff w \not\models \varphi \\
    w & \models \varphi \land \psi & & \iff w \models \varphi \; \land \; w \models \psi\\
    w & \models \X \varphi & & \iff w[1 \dots] \models \varphi \\
    w & \models \varphi \U \psi & & \iff \exists \; j \geq 0 \quad \mathrm{s.t.} \quad w[j \dots] \models \psi \quad \land \quad \forall \; 0 \leq i < j : w[i \dots] \models \varphi
\end{align*}

%%%%%%%%%%%%%%%%%%%%%%%%%%%%%%%%
% LDBAS AND LDBA CONTRACTION
%%%%%%%%%%%%%%%%%%%%%%%%%%%%%%%%

\section{Limit-Deterministic B\"{u}chi Automata}
\label{app:ldba}

We give here the formal conditions on the components of an LDBA $\mathcal{B} = \langle\mathcal{Q}, q_0, \Sigma, \delta, \mathcal{F}, \mathcal{E}\rangle$, as defined in Section~\ref{sec:background}. The state set $\mathcal{Q}$ is partitioned into two subsets $\mathcal{Q}_N$ (initial component) and $\mathcal{Q}_D$ (accepting component), i.e., $\mathcal{Q} = \mathcal{Q}_N \uplus \mathcal{Q}_D$, such that $q_0 \in \mathcal{Q}_N$ and $\mathcal{F} \subseteq \mathcal{Q}_D$. Subsets $\mathcal{Q}_N$ and $\mathcal{Q}_D$ are closed and deterministic under non-$\varepsilon$-transitions, i.e., $\delta(q, \sigma) \in \mathcal{Q}_D$ for all $q \in \mathcal{Q}_D$ and $\sigma \in \Sigma$, and $\delta(q, \sigma) \in \mathcal{Q}_N$ for all $q \in \mathcal{Q}_N$ and $\sigma \in \Sigma$. To transition from $\mathcal{Q}_N$ to $\mathcal{Q}_D$ requires taking a non-deterministic $\varepsilon$-transition $\varepsilon \in \mathcal{E}$; taking an $\varepsilon$-transition does not consume an assignment, and thus $\varepsilon$-transitions can be thought of intuitively as \textit{unobserved} jumps between two states. Note that there is at most one $\varepsilon$-transition for any path within the LDBA. Given an input $\omega$-word $w$, a \textit{run} $r = (q_0, q_1, q_2, \dots)$ of $\mathcal{B}$ is an infinite sequence of states $q \in \mathcal{Q}$ respecting the transition function $\delta$. An $\omega$-word $w$ is \textit{accepted} by $\mathcal{B}$ if there exists a run that infinitely often visits accepting states $q \in \mathcal{F}$.

Figure~\ref{fig:example_ldba:contracted} presents an LDBA for the formula $(\F \G \mathsf{a}) \lor \F \mathsf{b}$, which expresses a persistence requirement and a reachability specification. The automaton starts in state $q_0$ and transitions to the accepting state $q_1$ upon observing proposition $\mathsf{b}$. Once it has reached $q_1$, it stays there indefinitely. This captures the $\F \mathsf{b}$ (eventually satisfy $\mathsf{b}$) component of the specification. Alternatively, it can transition to the accepting state $q_2$ without consuming any input via the $\varepsilon$-transition $\varepsilon_{q_2}$. Once in $q_2$, the automaton accepts exactly the words where $\mathsf{a}$ is true at every step. This captures the $\F \G \mathsf{a}$ (eventually \textit{always} satisfy $\mathsf{a}$) component of the specification.

\begin{figure*}[t]
    \centering
    \subfloat[Contracted LDBA \label{fig:example_ldba:contracted}]{%
        \resizebox{0.393\linewidth}{!}{
            \begin{tikzpicture}[>={Latex}, shorten >=1pt, node distance=2cm, on grid, auto] 
               % States
               \node[state, initial, initial text=start] (q0) {$q_0$};
               \node[state, accepting] (q1) [right=of q0, yshift=1cm] {$q_1$};
               \node[state, accepting] (q2) [right=of q0, yshift=-1cm] {$q_2$};
               \node[state] (q3) [right=of q2] {$q_3$};
            
               % Self-loops
               \path[->] 
                   (q0) edge [loop above] node {$\lnot\mathsf{b}$} ()
                   (q1) edge [loop above] node {$\true$} ()
                   (q2) edge [loop above] node {$\mathsf{a}$} ()
                   (q3) edge [loop above] node {$\true$} ();
            
               % Transitions
               \path[->]
                   (q0) edge [] node {$\mathsf{b}$} (q1)
                   (q0) edge [] node[swap] {$\varepsilon_{q_2}$} (q2)
                   (q2) edge [] node {$\lnot\mathsf{a}$} (q3);
    
            \end{tikzpicture}
        }
    }
    \hfill
    \subfloat[Full LDBA ($\mathcal{Q}_N$ and $\mathcal{Q}_D$ are made distinct) \label{fig:example_ldba:full}]{%
        \resizebox{0.593\linewidth}{!}{
            \begin{tikzpicture}[>={Latex}, shorten >=1pt, node distance=2cm, on grid, auto] 
               % States
               \node[state, initial, initial text=start] (q0) {$q_0$};
               \node[state] (q1) [right=of q0, yshift=1cm] {$q_1$};
               \node[state, accepting] (q2) [right=of q0, xshift=3cm, yshift=-1cm] {$q_2$};
               \node[state] (q3) [right=of q2] {$q_3$};
               \node[state, accepting] (q4) [right=of q1, xshift=2cm] {$q_4$};
            
               % Self-loops
               \path[->] 
                   (q0) edge [loop above] node {$\lnot\mathsf{b}$} ()
                   (q1) edge [loop above] node {$\true$} ()
                   (q2) edge [loop above] node {$\mathsf{a}$} ()
                   (q3) edge [loop above] node {$\true$} ()
                   (q4) edge [loop right] node {$\true$} ();
            
               % Transitions
               \path[->]
                   (q0) edge [] node {$\mathsf{b}$} (q1)
                   (q0) edge [] node[swap, pos=0.8] {$\varepsilon_{q_2}$} (q2)
                   (q1) edge [] node[pos=0.35] {$\varepsilon_{q_4}$} (q4)
                   (q2) edge [] node {$\lnot\mathsf{a}$} (q3);
    
                % Subsets
                \begin{pgfonlayer}{background}
                    \tikzset{boxstyle/.style={draw=#1, dashed, ultra thick, rounded corners, inner sep=8pt}}
    
                    % --- Blue Shape Q_N ---
                    \node[boxstyle=blue, fit=(q0) (q1), label={[blue, font=\Large\bfseries]above:$\mathcal{Q}_N$}] (QNbox) {};
    
                    % --- Orange Shape Q_D ---
                    \node[boxstyle=orange, fit=(q2) (q3) (q4), label={[orange, font=\Large\bfseries]above:$\mathcal{Q}_D$}] (QDbox) {};
                \end{pgfonlayer}
            \end{tikzpicture}
        }
    }
    \caption{LDBA for the formula $(\F \G \mathsf{a}) \lor \F \mathsf{b}$.}
    \label{fig:example_ldba}
\end{figure*}

Note that Figure~\ref{fig:example_ldba:contracted} presents a contracted (yet equivalent) version of the full LDBA that strictly follows the LDBA definition in Section~\ref{sec:background}. The full LDBA is illustrated in Figure~\ref{fig:example_ldba:full}, with the initial component $\mathcal{Q}_N$ and accepting component $\mathcal{Q}_D$ made distinct. Note that this distinction is technically not possible in Figure~\ref{fig:example_ldba:contracted}'s contracted LDBA, yet the two automata are functionally equivalent since $\varepsilon$-transition $\varepsilon_{q_4}$ is trivial; in other words, the contracted LDBA simply encodes that the automaton should \textit{always} take $\varepsilon$-transition $\varepsilon_{q_4}$ if possible.

This \textit{LDBA contraction} is a practical optimization that removes all unnecessary $\varepsilon$-transitions, i.e., those corresponding to subformulae that do not require non-determinism (such as $\F \mathsf{b}$), and is done to reduce the number of nodes and edges while keeping LDBA behavior functionally equivalent. This helps to keep the reach-avoid sequences given to the agent as simple as possible, and also simplifies checking LTL satisfaction at run-time.

%%%%%%%%%%%%%%%%%%%%%%%%%%%%%%%%
% PRODUCT MDPs
%%%%%%%%%%%%%%%%%%%%%%%%%%%%%%%%

\section{Product MDPs}
\label{app:product_mdps}

Product MDPs are defined as follows:
\begin{definition}[Product MDP; \citealp{jackermeier_deepltl_2025}]
    \label{def:product_mdp}
    The \textit{product MDP} $\mathcal{M}^{\varphi}$ of an MDP $\mathcal{M}$ and an LDBA $\mathcal{B}^\varphi$ synchronizes the execution of $\mathcal{M}$ and $\mathcal{B}^\varphi$. Formally, $\mathcal{M}^\varphi$ is an MDP with state space $\mathcal{S}^\varphi = \mathcal{S} \times \mathcal{Q}^\varphi$, action space $\mathcal{A}^\varphi = \mathcal{A} \cup \mathcal{E}^\varphi$, initial state distribution $\mu^\varphi(s,q^\varphi) = \mu(s) \cdot \mathds{1}[q^\varphi = q^\varphi_0]$, and state-transition distribution
    \begin{equation*}
            p^\varphi ((s',q^{\prime\varphi}) \cond (s, q^\varphi), a) =
            \begin{cases}
                p(s' \cond s, a) & \text{if } a \in \mathcal{A} \land q^{\prime\varphi} = \delta^\varphi(q^\varphi, L^\varphi(s)), \\
                1 & \text{if } a = \varepsilon_{q^{\prime\varphi}} \land q^{\prime\varphi} = \delta^\varphi(q^\varphi, a) \land s' = s, \\
                0 & \text{else}.
            \end{cases}
    \end{equation*}
\end{definition}
The action space in $\mathcal{M}^\varphi$ is extended with $\mathcal{E}^\varphi$ to allow the policy to select $\varepsilon$-transitions in $\mathcal{B}^\varphi$ without acting in the MDP. Trajectories in $\mathcal{M}^\varphi$ are infinite sequences $\tau^\varphi = \big((s_0, q^\varphi_0), a_0, (s_1, q^\varphi_1), a_1, \dots \big)$, and the run $\tau^\varphi_q = (q^\varphi_0, q^\varphi_1, q^\varphi_2, \dots)$ is the projection of $\tau^\varphi$ onto $\mathcal{B}^\varphi$.

\section{Aside: Connection to Plato's Theory of Forms}
\label{app:plato_theory_of_forms}

Anecdotally, it is often challenging to construct an LTL specification that accurately matches the desired task outcome, in particular those tasks originally defined in natural language. This is because natural language tasks tend to implicitly assume constraints, dependencies, and edge cases that are imperfectly translated (or forgotten entirely) during the formulation of the LTL specification.

We can draw a connection to Plato's Theory of Forms \citep{plato_republic_1992}, wherein the ultimate ``Form of the Good'' (desired task outcome) is imperfectly represented as a composition of base concepts, or ``Forms'' (atomic predicates), which are instantiated as ``particulars'' (atomic propositions as predicate instances) to create an imperfect imitation of ``Good'' (the LTL specification).

While not scientifically relevant, we found this connection amusing and it inspired the name of our approach, and we believe it to be worthy of note.

%%%%%%%%%%%%%%%%%%%%%%%%%%%%%%%%
% FIRST-ORDER LOGIC
%%%%%%%%%%%%%%%%%%%%%%%%%%%%%%%%

\section{Connection to First-Order LTL}
\label{app:fo-ltl}

First-order LTL (FOLTL; \citealp{gigante_introduction_2025}) extends LTL with atomic predicates over variables (defined over some domain $D$), as well as $\exists$ (exists) and $\forall$ (for-all) quantifiers that operate over these variables. Unlike the parameterized specifications considered in this paper, the variables in FOLTL are not necessarily instantiated with specific values within a given specification.

\begin{example}
    \label{exm:fo-ltl_1}
    The FOLTL formula $\G(\exists \; x, y, z \;[ \mathsf{pos}(x,y,z) \land \mathsf{gt}(z, 0.0)])$ represents that the agent position must remain above the $z=0$ plane for all time. Literally, the formula can be interpreted as ``always, there exists an assignment of position variable values such that 1) the values match the agent position, and 2) the $z$-value is greater than 0''.
\end{example}

\begin{example}
    \label{exm:fo-ltl_2}
    The FOLTL formula $\forall \; r, g, b \; [\G(\mathsf{at}(r,g,b) \Rightarrow F(\mathsf{at}(b,g,r)))]$ represents that whenever the agent enters a colored zone, it must eventually enter a colored zone with the R- and B-values swapped. Literally, the formula can be interpreted as ``for all assignments of color variable values, whenever the agent enters a colored zone matching the color values, it must eventually enter a colored zone matching the same color values but with $r$ and $b$ swapped''.
\end{example}

To ground FOLTL specifications in an MDP, each predicate $\mathsf{f}$ requires a labeling function $L_{\mathsf{f}}(s, x_\mathsf{f}): \mathcal{S} \times \mathcal{X}_\mathsf{f} \to \{0, 1\}$, which returns its truth value for a given state $s \in \mathcal{S}$ and variable values $x_\mathsf{f} \in \mathcal{X}_\mathsf{f}$.

Unfortunately, FOLTL is undecidable in general \citep{gigante_introduction_2025}, and thus arbitrary FOLTL formulae cannot be translated into finite automata.\footnote{Recent work by \citet{olivieri_it_2026} trains policies to satisfy specifications of LTLfMT, a finite-trace fragment of FOLTL modulo theories, but the method is single-task and restricted to finite-horizon specifications.} However, under the following restrictions, the logic reduces to LTL:
\begin{enumerate}[itemsep=-1.5pt, topsep=-1.5pt]
    \item The $\exists$ and $\forall$ quantifiers are removed.
    \item All variables are instantiated with fixed values within a given specification.
\end{enumerate}
Under the first restriction, the syntax reduces to predicates and LTL's logical and temporal operators only. Under the second restriction, the predicates become \textit{predicate instances}, which are atomic propositions; given fixed parameters $x_\mathsf{f}$, the labeling function for each predicate instance becomes $L_{\mathsf{f}(x_\mathsf{f})}(s): \mathcal{S} \to \{0, 1\}$, and the labeling function across all predicate instances in a specification $\varphi$ becomes $L^\varphi(s): \mathcal{S} \to \Sigma^\varphi$. Thus, our parameterized specifications belong to a fragment of FOLTL that behaves exactly like LTL from a logic perspective; the only difference from standard LTL is the replacement of propositions with predicate instances, which is purely notational.

%%%%%%%%%%%%%%%%%%%%%%%%%%%%%%%%
% STRUCT LTL
%%%%%%%%%%%%%%%%%%%%%%%%%%%%%%%%

\section{Example Backbone: StructLTL}
\label{app:struct_ltl}

This appendix outlines core details regarding the StructLTL backbone, introduced by \cite{jackermeier_zero-shot_2026}, with the purpose of illustrating how a typical backbone interacts with our problem setting (Section~\ref{sec:problem_setting}), the Plato embedding module (Section~\ref{sec:plato}), and P\textsuperscript{2}BRS (Section~\ref{sec:p2brs}).

\subsection{Reward Function}
\label{app:structltl:reward_function}

StructLTL trains a sequence-conditioned policy $\pi_\beta(a \cond s, \boldsymbol{\beta}_\pm)$. Given an $n$-step reach-avoid sequence $\boldsymbol\beta_\pm = \big( (\beta^+_i, \beta^-_i) \big)^{n-1}_{i=0}$, StructLTL assigns a reward $r^\varphi = +1$ if the agent satisfies $\beta^+_0$ and $n=1$; it assigns $r^\varphi = -1$ if the agent satisfies $\beta^-_0$; it assigns $r^\varphi = 0$ otherwise. Furthermore, the sequence is progressed to $\big( (\beta^+_i, \beta^-_i) \big)^{n-1}_{i=1}$ if the agent satisfies $\beta^+_0$, keeping the task Markovian. By training over a distribution of such sequences, StructLTL learns a policy that generalizes across sequences, and therefore across LDBAs (via the sequences extracted from depth-first search).

This \textit{sparse} reward function $r^\varphi$ is used to approximately solve Equation~\ref{eqn:rl_problem} by training a policy that can execute paths to accepting states in an LDBA, thus implicitly maximizing LTL satisfaction probability by maximizing the number of visits to accepting states as a proxy. However, the trained policy is \textit{biased} towards earlier satisfaction due to the discount factor $\gamma$.

Thus, when we claim that the augmented reward function $r^\varphi_\mathrm{aug}(s, q^{\varphi}, a, s^\prime, q^{\prime\varphi}) = r^\varphi + \gamma \Phi(s^\prime, q^{\prime\varphi}) - \Phi(s, q^{\varphi})$ under P\textsuperscript{2}BRS does not bias the optimal policy, this is with respect to the optimal policy under the \textit{backbone's} reward function $r^\varphi$, not the hypothetical policy that maximizes Equation~\ref{eqn:rl_problem}.

\subsection{Model Architecture}
\label{app:structltl:model_architecture}

To compute an embedding $\mathbf{z}_{\boldsymbol{\beta}_\pm}$ for $\boldsymbol\beta_\pm$, StructLTL first computes an embedding $\mathbf{z}_{\beta}$ for each constituent Boolean formula $\beta$. Note that each Boolean formula in $\boldsymbol{\beta}^\varphi_\pm = \text{DFS}_{\mathcal{B}^\varphi}(q^\varphi)$ is in disjunctive normal form (DNF), i.e., $\beta = \bigvee_{k=1}^{n_\beta}\bigwedge_{j=1}^{m_{\beta k}} l_{kj}$, where all $l_{kj} \in \{\mathsf{p}, \lnot \mathsf{p}\}$ and $\mathsf{p} \in \mathit{AP}^\varphi$; see Appendix~\ref{app:reach_avoid_sequence_extraction} for details. StructLTL exploits DNF structure by building clause embeddings from proposition embeddings, and then building disjunction embeddings from those clause embeddings:
\begin{equation*}
  \mathbf{z}_{\beta} = f_{\lor}\left(\sum_{k=1}^{n_\beta} f_{\land}\left( \sum_{j=1}^{m_{\beta k}} \Pi\big(\phi(\mathsf{p}_{kj})\big)\right) \right),
\end{equation*}
where $\phi$ is a learned encoder that maps proposition tokens to embeddings, $\Pi$ denotes a learned linear projection if $l_{kj}$ is negated and the identity function otherwise, and $f_\lor$ and $f_\land$ are learned networks applied after sum pooling, such that each level forms a permutation-invariant DeepSets encoder \citep{zaheer_deep_2017}. An empty set (i.e., a clause with no literals, or a formula with no clauses; see Appendix~\ref{app:reach_avoid_sequence_extraction:boolean_formula_construction}) is mapped to a learned embedding in place of the sum. We see that StructLTL satisfies Assumption~\ref{asm:embedding}, and the learned mapping $\phi$ can be replaced with the Plato embedding module.

The embeddings for each $\beta^+_i$ and $\beta^-_i$ are concatenated to obtain the embedding $\mathbf{z}_i = \mathbf{z}_{\beta^+_i} || \mathbf{z}_{\beta^-_i}$ for each sequence step. The sequence embedding is then computed as that of the current subgoal contextualized by future subgoals; $\mathbf{z}_{\boldsymbol{\beta}_\pm} = \text{LayerNorm}(\mathbf{z}_0 + \mathbf{h})$~\citep{ba_layer_2016}, where $\mathbf{h}$ is computed using single-head dot-product attention \citep{vaswani_attention_2017}:
\begin{equation*}
  \mathbf{h} = \left[\softmax\left(\frac{\mathbf{q}K^\top}{\sqrt{d_k}}\right)V\right]W^O,
\end{equation*}
where $\mathbf{q} = \mathbf{z}_0 W^Q$ is the query vector, $W^O$ is the output projection matrix, and $K, V\in\mathbb R^{n\times d_k}$ are the key and value matrices obtained by projecting $\mathbf{z}_{0:n-1}$ using $W^K$ and $W^V$, respectively. StructLTL employs ALiBi \citep{press_train_2022} to bias the attention toward earlier subgoals, subtracting a linear penalty $m\cdot i$ from each unnormalized attention score before applying the $\softmax$, with $m$ being a fixed hyperparameter.

Finally, StructLTL concatenates $\mathbf{z}_{\boldsymbol{\beta}_\pm}$ with a learned encoding of the environment MDP state and passes this into an \textit{actor-critic} MLP to obtain a distribution over actions and a predicted value. To handle $\varepsilon$-transitions, the policy's action space is augmented with a special $\varepsilon$-action which triggers an $\varepsilon$-transition in the LDBA state without performing an action in the environment MDP. For discrete action spaces, the policy network outputs a categorical distribution with an additional logit for the $\varepsilon$-action, while for continuous action spaces the policy network outputs a mixed discrete-continuous distribution. Thus, the policy must learn to predict \textit{when} to trigger the $\varepsilon$-transition. The probability of selecting the $\varepsilon$-action is set to $0$ when no $\varepsilon$-transitions are available for the current LDBA state.

%%%%%%%%%%%%%%%%%%%%%%%%%%%%%%%%
% PBRS PROOF
%%%%%%%%%%%%%%%%%%%%%%%%%%%%%%%%

\section{Policy Invariance under P\textsuperscript{2}BRS}
\label{app:pbrs_proof}

P\textsuperscript{2}BRS relies on the classic policy invariance of PBRS \citep{ng_policy_1999}, which we restate over product states and prove for completeness:
\begin{theorem}
    \label{thm:pbrs}
    Given a bounded function $\Phi(s, q^{\varphi})\colon \mathcal{S} \times \mathcal{Q}^\varphi \to \mathbb{R}$, the optimal policy under the augmented reward function $r^\varphi_\mathrm{aug} = r^\varphi + \gamma \Phi(s^\prime, q^{\prime\varphi}) - \Phi(s, q^{\varphi})$ is also optimal under $r^\varphi$.
\end{theorem}
\begin{proof}
Consider the discounted return under $r^\varphi_\mathrm{aug} = r^\varphi + \gamma \Phi(s^\prime, q^{\prime\varphi}) - \Phi(s, q^{\varphi})$:
\begin{align*}
    R^\varphi_\mathrm{aug}(\tau^\varphi) & = \lim_{T \to \infty} \sum_{t=0}^{T-1} \gamma^t r^\varphi_t + \lim_{T \to \infty} \sum_{t=0}^{T-1} \gamma^{t+1} \Phi(s_{t+1}, q^\varphi_{t+1}) - \lim_{T \to \infty} \sum_{t=0}^{T-1} \gamma^{t} \Phi(s_t, q^\varphi_t) \\
    & = R^\varphi(\tau^\varphi) + \lim_{T \to \infty} \sum_{t=1}^{T} \gamma^{t} \Phi(s_{t}, q^\varphi_t) - \lim_{T \to \infty} \sum_{t=0}^{T-1} \gamma^{t} \Phi(s_t, q^\varphi_t) \\
    & = R^\varphi(\tau^\varphi) + \lim_{T \to \infty} \big(\gamma^{T} \Phi(s_{T}, q^\varphi_T)\big) - \Phi(s_0, q^\varphi_0) \\
    & = R^\varphi(\tau^\varphi) - \Phi(s_0, q^\varphi_0),
\end{align*}
where the limit vanishes since $\Phi$ is bounded. The offset
$\Phi(s_0, q^\varphi_0)$ does not depend on the actions taken, so it shifts the expected return of every policy equally (and, applied from any product state, shifts its value by $\Phi(s, q^\varphi)$), leaving the set of optimal policies unchanged.
\end{proof}

\paragraph{Finite Episodes.} When an episode terminates at $T < \infty$, the same telescoping yields $R^\varphi_{\mathrm{aug}}(\tau^\varphi) = R^\varphi(\tau^\varphi) + \gamma^T \Phi(s_T, q^\varphi_T) - \Phi(s_0, q^\varphi_0)$, where the residual term depends on the trajectory and would bias the optimal policy. The terminal override $\Phi(s_T, q^\varphi_T) = 0$ (Section~\ref{sec:p2brs:potential_function}) eliminates it, so the result of Theorem~\ref{thm:pbrs} extends to finite episodes.

\textbf{Interaction with Eventual Discounting}. Some backbones may employ \textit{eventual discounting} for their reward functions, only discounting time steps corresponding to visits to accepting LDBA states \citep{voloshin_eventual_2023}. In such cases, eventual discounting must \textit{not} be applied to the potential difference component of the augmented reward, which requires standard discounting for the return to telescope correctly.

%%%%%%%%%%%%%%%%%%%%%%%%%%%%%%%%
% PERSISTENCE GATING
%%%%%%%%%%%%%%%%%%%%%%%%%%%%%%%%

\section{Persistence Gating}
\label{app:persistence_gating}

This section gives the full definition of the Boolean formula robustness metric of Section~\ref{sec:p2brs:robustness:boolean_formulae}. For a formula $\beta$ satisfying Assumption~\ref{asm:dnf}, let $J^+_k = \{\, j : \mathsf{f}_{kj} \in \mathit{Pred}^+ \,\}$ index the persistent literals of clause $c_k$ and $J^-_k = \{1{:}m_{\beta k}\} \setminus J^+_k$ the rest, where $\mathsf{f}_{kj}$ is the predicate of literal $l_{kj}$. With the conventions $\min_{j \in \emptyset}(\cdot) = 1$ and $\max_{k \in \emptyset}(\cdot) = -1$, we compute
\begin{align*}
    \rho_{l_{kj}}(s) =
    \begin{cases}
        \hphantom{-}\rho_\mathsf{f}(s, x_\mathsf{f}) & \text{if } l_{kj} = \mathsf{f}(x_\mathsf{f}), \\
        -\rho_\mathsf{f}(s, x_\mathsf{f}) & \text{if } l_{kj} = \lnot\mathsf{f}(x_\mathsf{f}),
    \end{cases} \quad
    h_k(s) = \min_{j \in J^+_k} \rho_{l_{kj}}(s), \quad
    t_k(s) = \min_{j \in J^-_k} \rho_{l_{kj}}(s), \\
    \rho_{c_k}(s) =
    \begin{cases}
        h_k(s) \wedge_\sigma t_k(s) & \text{if } \emptyset \neq J^+_k \neq \{1{:}m_{\beta k}\}, \\
        \min\big(h_k(s), t_k(s)\big) & \text{otherwise},
    \end{cases} \qquad
    \rho_\beta(s) = \max_{k \in 1:n_\beta} \rho_{c_k}(s).
\end{align*}

\paragraph{Negated Literals.} Persistence is a property of the predicate rather than the literal: the truth value of a persistent predicate typically changes only through actions directed at it (e.g., grasping or releasing an object), so either value, once attained, tends to carry over while the agent works towards other literals. Gating therefore applies equally to negated literals; for example, for the clause $\lnot\mathsf{has}(obj) \land \mathsf{pos}(x,y,z)$, the agent is guided to first release the object and then move to the target position, remaining empty-handed along the way. Other orderings may also satisfy the clause, such as carrying the object to the target and releasing it there, since releasing does not move the gripper. Gating need not identify the only valid ordering, but rather one that is simple and reliable to guide towards; as with any potential function, it shapes exploration without restricting the policy to which the agent ultimately converges.

\paragraph{Ungated Clauses.} When $J^+_k = \emptyset$ or $J^-_k = \emptyset$, the empty-set convention makes one argument of $\wedge_\sigma$ identically $1$, so the staged conjunction would compress the whole unsatisfied range of the clause into a single stage's slice. We therefore fall back to the plain $\min$ over all literals of $c_k$.

\paragraph{Satisfaction.} Since $|\rho_\mathsf{f}| > 0$, the staged conjunction is negative whenever either argument is negative, and positive otherwise, so $\rho_\beta(s) > 0$ if and only if $\beta$ holds, mirroring $L_\mathsf{f} = \mathcal{H}(\rho_\mathsf{f})$ for individual predicates. Moreover, $\rho_\beta$ remains in $[-1, 1]$, keeping $\Phi$ bounded.

%%%%%%%%%%%%%%%%%%%%%%%%%%%%%%%%
% REACH-AVOID SEQUENCE EXTRACTION
%%%%%%%%%%%%%%%%%%%%%%%%%%%%%%%%

\section{Reach-Avoid Sequence Extraction}
\label{app:reach_avoid_sequence_extraction}

This appendix details the Boolean reach-avoid sequence extraction procedure abstracted as $\boldsymbol{\beta}^\varphi_\pm = \text{DFS}_{\mathcal{B}^\varphi}(q^\varphi)$ in the main text. The full procedure was first introduced by \cite{jackermeier_zero-shot_2026} but is outlined again here for clarity. The procedure comprises three stages: extraction of all accepting runs from $q^\varphi$ (Appendix \ref{app:reach_avoid_sequence_extraction:computing_paths_to_accepting_cycles}), followed by construction of $\boldsymbol{\beta}^\varphi_\pm$ in minimal disjunctive normal form (Appendix \ref{app:reach_avoid_sequence_extraction:boolean_formula_construction}), and finally selection of the optimal sequence (Appendix \ref{app:reach_avoid_sequence_extraction:optimal_sequence_selection}).

\subsection{Computing Paths to Accepting Cycles}
\label{app:reach_avoid_sequence_extraction:computing_paths_to_accepting_cycles}

The DFS algorithm to compute all paths to accepting cycles in $\mathcal{B}^\varphi$ from $q^\varphi$, i.e., all accepting runs from $q^\varphi$, is presented in Algorithm \ref{alg:dfs}. Note that the reach-avoid sequences are infinite, and so we approximate the looping part of the accepting cycle by truncation such that the truncated sequence visits an accepting state $k$ times, where $k$ is a hyperparameter (set to $k=2$ for our experiments).

% Furthermore, as noted by \citet{jackermeier_deepltl_2025}, in practice it can prove too restrictive for the trained policy to avoid all LDBA states that are not the next state in the selected accepting run (or the current state). Therefore, we mark as ``avoid'' only those LDBA states that are \textit{sink states} (from which there does not exist any valid path to an accepting cycle) and \textit{regressive states} that lead the agent back along the current path (and thus represent regressing back to an earlier stage of the task).

\begin{algorithm*}
\caption{Computing paths to accepting cycles \citep{jackermeier_deepltl_2025}}
\label{alg:dfs}
\begin{algorithmic}[1]
\Require An LDBA $\mathcal{B} = \langle \mathcal{Q}, q_0, \Sigma, \delta, \mathcal{F}, \mathcal{E} \rangle$ and current state $q$

\Function{DFS}{$q, p, i$}
    \Comment{$i$ is the index of the last seen accepting state, or $-1$ otherwise}
    \State $P \gets \emptyset$
    \If{$q \in \mathcal{F}$}
        \State $i \gets |p|$
    \EndIf
    \ForAll{$\sigma \in \Sigma \cup \mathcal{E}$}
        \State $p' \gets [p, q]$
        \State $q' \gets \delta(q, \sigma)$
        \If{$q' \in p$}
            \If{$\Call{Index}{q', p} \le i$}
                \State $P \gets P \cup \{p'\}$
            \EndIf
        \Else
            \State $P \gets P \cup \Call{DFS}{q', p', i}$
        \EndIf
    \EndFor
    \State \Return $P$
\EndFunction

\State $i \gets \begin{cases}
    0 & \text{if } q \in \mathcal{F} \\
    -1 & \text{else}
\end{cases}$

\State \Return \Call{DFS}{$q, [], i$}
\end{algorithmic}
\end{algorithm*}

\subsection{Boolean Formula Construction}
\label{app:reach_avoid_sequence_extraction:boolean_formula_construction}

For each $n$-step prefix of an accepting run $\mathbf{q}^\varphi = (q^\varphi_i)^{n}_{i=0}$, where $q^\varphi_0 = q^\varphi$, we construct a reach-avoid sequence of Boolean formulae $\boldsymbol\beta^\varphi_\pm =  \big((\beta^{\varphi+}_i, \beta^{\varphi-}_i)\big)^{n-1}_{i=0}$ such that
\begin{align*}
    & \forall \sigma^\varphi \in \Sigma^\varphi: \sigma^\varphi \models \beta^{\varphi+}_i \iff \delta^\varphi(q^\varphi_i, \sigma^\varphi) = q^\varphi_{i+1}, \\
    & \forall \sigma^\varphi \in \Sigma^\varphi: \sigma^\varphi \models \beta^{\varphi-}_i \iff \delta^\varphi(q^\varphi_i, \sigma^\varphi) \not\in \{q^\varphi_i, q^\varphi_{i+1}\}.
\end{align*}
In other words, $\beta^{\varphi+}_i$ represents the set of assignments that trigger the transition from $q^\varphi_i$ to $q^\varphi_{i+1}$, while $\beta^{\varphi-}_i$ represents the set of assignments that trigger a transition to any other state (excluding self-loops). Sequence $\boldsymbol\beta^\varphi_\pm$ is thus a reach-avoid sequence to achieve accepting run $\mathbf{q}^\varphi$.

The construction is done using the Quine-McCluskey algorithm \citep{quine_problem_1952, mccluskey_minimization_1956} to obtain each $\beta_i^{\varphi+}$ and $\beta_i^{\varphi-}$ in minimal disjunctive normal form (DNF). Given a set of satisfying assignments $S \subseteq \Sigma^\varphi$ and a set of non-satisfying assignments $N \subseteq \Sigma^\varphi$, the algorithm produces a Boolean formula that evaluates to true for assignments $\sigma^\varphi \in S$ and to false for assignments $\sigma^\varphi \in N$. The produced formula $\beta$ is \textit{DNF} because it is a disjunction of clauses of possibly negated propositions, i.e., $\beta = \bigvee_{k=1}^{n_\beta}\bigwedge_{j=1}^{m_{\beta k}} l_{kj}$, where all $l_{kj} \in \mathit{AP}^\varphi \cup \{\lnot \mathsf{p}: \mathsf{p} \in \mathit{AP}^\varphi\}$. The formula is also \textit{minimal} in the number of clauses and, among those, the number of literals. For $\beta_i^{\varphi+}$ we set $S = \{\sigma^\varphi \in \Sigma^\varphi : \delta^\varphi(q^\varphi_i, \sigma^\varphi) = q^\varphi_{i+1}\}$, for $\beta_i^{\varphi-}$ we set $S = \big\{\sigma^\varphi \in \Sigma^\varphi : \delta^\varphi(q^\varphi_i, \sigma^\varphi) \not\in \{q^\varphi_i, q^\varphi_{i+1}\}\big\}$, and for both cases we use $N = \Sigma^\varphi \setminus S$. The degenerate cases $S = \emptyset$ and $S = \Sigma^\varphi$ yield the trivial formulae $\false$ and $\true$, respectively; we treat $\false$ as a degenerate DNF with no clauses ($n_\beta = 0$), and we treat $\true$ as a degenerate DNF with a single clause of no literals ($n_\beta = 1$, $m_{\beta 1} = 0$). In particular, $\beta^{\varphi-}_i = \false$ whenever no transition from $q^\varphi_i$ leads to a state other than $q^\varphi_i$ or $q^\varphi_{i+1}$, i.e., there is nothing to avoid. Under the conventions of Appendix~\ref{app:persistence_gating}, their bounded robustness metrics are $\rho_\true = 1$ and $\rho_\false = -1$, corresponding to $g = \pm\infty$.

If a run $\mathbf{q}^\varphi$ requires an $\varepsilon$-transition from $q^\varphi_i$ to $q^\varphi_{i+1}$, we set $\beta^{\varphi+}_i = \varepsilon$ in the corresponding reach-avoid sequence, where $\varepsilon$ is a shared token used to represent any $\varepsilon$-transition. While there may be multiple $\varepsilon$-transitions in a given LDBA, there is at most one $\varepsilon$-transition for any path within the LDBA, and thus the rest of the ``reach'' component of $\boldsymbol\beta^\varphi_\pm$ gives sufficient context to uniquely identify its $\varepsilon$-transition. Meanwhile, we ignore any $\varepsilon$-transitions from $q^\varphi_i$ to any other state, representing that the agent should never choose to take those $\varepsilon$-transitions.

In practice, both the DFS (run from every automaton state) and the Boolean formula construction can be performed alongside LDBA construction prior to training, yielding a cached set of reach-avoid sequences for every state in each formula structure's LDBA.

% This means that there is no need for multiple $\varepsilon$-actions in the policy; the single $\varepsilon$-action corresponds to whichever $\varepsilon$-transition is represented by the $\varepsilon$ token in the ``reach'' component of $\boldsymbol\beta^\varphi_\pm$ (if present).

\subsection{Optimal Sequence Selection}
\label{app:reach_avoid_sequence_extraction:optimal_sequence_selection}

The DFS may yield a set of multiple reach-avoid sequences, $\mathit{RA}_{q^\varphi} = \{\boldsymbol\beta^\varphi_\pm\}_{q^\varphi}$, requiring us to choose among them upon changing LDBA state. We note that for backbones such as StructLTL, the policy's learned value function (or ``critic'') is an approximate lower bound on the (discounted) probability of LTL satisfaction, and thus choosing the reach-avoid sequence that maximizes critic value also maximizes (predicted) LTL satisfaction; see \citet{jackermeier_deepltl_2025} for details. Thus, our heuristic for sequence selection chooses the sequence that is predicted to maximize value.

To obtain a sequence for P\textsuperscript{2}BRS at training time, we select the sequence that maximizes the sequence-conditioned potential function $\Phi_\beta$, i.e., $\boldsymbol\beta^{\varphi*}_\pm = \argmax_{\boldsymbol\beta^\varphi_\pm \in \mathit{RA}_{q^\varphi}} \Phi_\beta(s, \boldsymbol\beta^\varphi_\pm)$; this is because the potential function acts as an a priori initialization of the value estimates \citep{wiewiora_potential-based_2003}. At evaluation time, given the sequence-conditioned policy $\pi_\beta(a \cond s, \boldsymbol\beta_\pm)$, StructLTL selects the optimal sequence $\boldsymbol\beta^{\varphi*}_\pm = \argmax_{\boldsymbol\beta^\varphi_\pm \in \mathit{RA}_{q^\varphi}} V^{\pi_\beta}(s, \boldsymbol\beta^\varphi_\pm)$, where $V^{\pi_\beta}$ is the policy's value function. In both cases, a new sequence is only selected upon episode reset or changing LDBA state, i.e., $q^{\prime\varphi} \not= q^\varphi$, to ensure that the policy can commit to a sequence once selected.

%%%%%%%%%%%%%%%%%%%%%%%%%%%%%%%%
% EXPERIMENTAL DETAILS
%%%%%%%%%%%%%%%%%%%%%%%%%%%%%%%%

\section{Experimental Details}
\label{app:experimental_details}

\subsection{Implementation}
\label{app:experimental_details:implementation}

Our implementation is built in JAX \citep{bradbury_jax_2018}. We use MuJoCo XLA (MJX; \citealp{todorov_mujoco_2012}) for robotics simulation with models from the MuJoCo Menagerie \citep{zakka_mujoco_2022}, and we use Rabinizer 4 \citep{chockler_rabinizer_2018} to convert LTL specifications into LDBAs. All experiments were run on an NVIDIA GeForce RTX 5090 GPU.

\subsection{Environments}
\label{app:experimental_details:environments}

\subsubsection{Go1RGBZoneEnv}
\label{app:experimental_details:environments:rgb_zone_env}

\textit{Go1RGBZoneEnv} expands upon \textit{ZoneEnv} \citep{vaezipoor_ltl2action_2021}, requiring an MJX-simulated Unitree Go1 quadruped robot to navigate between colored zones. The arena is a square of side \SI{6.6}{\metre} containing eight colored circular zones of radius \SI{0.4}{\metre}: four colors, with two zones per color. At the start of each episode, the zone centers are sampled uniformly over the central region of the arena, at least \SI{1.1}{\metre} apart, and the robot is initialized in its standing pose at a uniformly sampled position at least \SI{1.1}{\metre} from every zone center, with a uniformly sampled heading. The episode terminates if the robot leaves the arena or falls, i.e., its torso rolls past horizontal or its height drops below \SI{0.10}{\metre}; episodes are otherwise truncated after $1000$ policy steps (\SI{60}{\second} of simulated time) during training. See Figure~\ref{fig:environments:go1} for visualizations of example reset configurations.

\textbf{Observations \& Actions.} The observation comprises the torso-frame planar velocity, the yaw rate, the last commanded forward speed, and a 32-bin RGB LiDAR. The LiDAR partitions bearings relative to the robot's heading into evenly spaced angular bins; each bin returns the color $(r, g, b)$ of the nearest zone whose center lies within it, an intensity $e^{-k d}$ (with $k = \SI{0.5}{\per\metre}$) that decays with the distance $d$ to that zone's center, and a detection flag, with empty bins reading zero. All components are normalized to $[-1, 1]$. The action is a two-dimensional velocity command: forward speed $v_x \in [-1.5, 1.5]$\,\si{\metre\per\second} and yaw rate $\omega \in [-1.5, 1.5]$\,\si{\radian\per\second}.

\textbf{Low-Level Control.} The velocity command is tracked by a pre-trained locomotion controller, which is frozen during task training and executed deterministically. The controller is an MLP with hidden layers $(512, 256, 128)$, trained with PPO \citep{schulman_proximal_2017} on a velocity-tracking joystick task whose hyperparameters and reward terms follow MuJoCo Playground's Go1 joystick task \citep{zakka_mujoco_2025}, over a command envelope of $\pm$\SI{1.8}{\metre\per\second} and $\pm$\SI{2.0}{\radian\per\second}. It maps a 47-dimensional proprioceptive input --- torso-frame linear and angular velocity, the projected gravity direction, joint angles relative to the standing pose, joint velocities, the previous controller action, and the current command --- to position targets for the twelve joints, expressed as offsets from the standing pose. The physics timestep is \SI{4}{\milli\second} and the controller runs at \SI{50}{\hertz}; each policy step holds the command for three controller steps, inducing a \SI{0.06}{\second} task-level control period.  Although the task-level policy does not observe the Go1's joint-level proprioceptive state, the closed-loop behavior of the low-level controller forms part of the environment dynamics, which the policy must learn to account for through its partial observation of the torso state.

\textbf{Predicates.} The environment provides a single predicate $\mathsf{at}$ with color parameters $(r, g, b) \in [0, 1]^3$; specifications may name up to four instances (matching the zone colors). A predicate instance $\mathsf{at}(r,g,b)$ holds if and only if the robot's planar torso position lies within the radius of a zone whose color is within Euclidean distance $0.25$ of $(r,g,b)$. Its robustness metric is the signed distance to the boundary of the nearest such color-matching zone, normalized by the zone radius (positive inside the zone); if no zone matches, the metric is the constant $-1$. The predicate is annotated as \textit{non-persistent}. The world's four zone colors are drawn jointly and uniformly over the color cube, subject to a pairwise Euclidean separation of at least $0.7$; the specification's $\mathsf{at}$ predicate instances take their parameters from this draw, so every named color is realized by two zones in the world. Since the separation exceeds twice the labeling tolerance, the tolerance balls around distinct instances are disjoint, and a zone matches at most one instance.

\subsubsection{X2DronesEnv}
\label{app:experimental_details:environments:x2_drones_env}

\textit{X2DronesEnv} loosely expands upon \textit{FlatWorld} \citep{voloshin_eventual_2023}, requiring two MJX-simulated Skydio X2 quadcopters to navigate between spherical position targets in a shared airspace: a cube of side \SI{20}{\metre} with no ground plane, in which the drones fly under gravity and aerodynamic drag. At the start of each episode, spherical position targets are placed throughout the airspace, and each drone is initialized at rest and level at a uniformly sampled position at least \SI{5.5}{\metre} from every target center, with a uniformly sampled heading. The episode terminates if any drone leaves the airspace or if the two drones collide, i.e., their centers come within \SI{0.72}{\metre} of one another (twice the radius of a sphere bounding each drone's propeller tips); episodes are otherwise truncated after $1000$ policy steps (\SI{50}{\second} of simulated time) during training. See Figure~\ref{fig:environments:x2} for visualizations of example reset configurations.

\textbf{Observations \& Actions.} A single centralized policy controls both drones. The observation comprises, for each drone: its position, normalized to $[-1, 1]^3$; its attitude, as the first two columns of its rotation matrix (in the continuous 6D representation described by \citealp{zhou_continuity_2019}); its linear velocity, expressed in its heading frame (i.e., the world frame rotated by the drone's yaw) and normalized by the maximum commanded speed; and two proximity readings $e^{-kc}$ on the clearance $c$ to the nearest wall ($k = \SI{1}{\per\metre}$) and to the other drone ($k = \SI{0.75}{\per\metre}$). The action comprises, for each drone, a velocity command $(v_x, v_y, v_z) \in [-4, 4]^3$\,\si{\metre\per\second} and a yaw rate $\omega \in [-2, 2]$\,\si{\radian\per\second}, expressed in the same heading frame, so that a commanded $v_x$ means the same horizontal direction whether or not the drone is tilting to accelerate.

The environment itself supports an arbitrary number of drones, but we deliberately study the two-drone configuration. In preliminary experiments with three drones, none of the methods we evaluated learned a strong policy. We attribute this to the joint observation and action spaces growing with the swarm while the policy remains a single centralized network with no weight sharing or permutation equivariance across drones, so the drone-to-target assignment implicit in wide conjunctions must be learned without any structural support --- a setting more naturally addressed by multi-agent RL (MARL; \citealp{albrecht_multi-agent_2024}), which has been explored in combination with temporal-logic- and automaton-guided RL \citep{hammond_multi-agent_2021, neary_reward_2021}. Such an extension is beyond the scope of this paper: two drones already suffice to exhibit the homogeneous conjunctions this environment is designed to study, and we leave swarm-scale variants to future work.

\textbf{Low-Level Control.} Each drone's velocity command is tracked by an identical model-based cascaded controller: a proportional velocity loop produces a desired thrust vector, whose tilt from vertical is capped at \SI{30}{\degree}; a geometric attitude controller \citep{lee_geometric_2010} computes the body torques that align the drone with this thrust vector; and the resulting thrust and torques are allocated to the four rotors within the actuator limits. The physics timestep is \SI{10}{\milli\second} and the controller runs at every physics step (\SI{100}{\hertz}); each policy step holds the command for five physics steps, inducing a \SI{0.05}{\second} task-level control period. Although the policy does not observe the drones' angular velocities or rotor thrusts, the closed-loop behavior of the controller (i.e., its bounded tilt and its lag in tracking aggressive commands) forms part of the environment dynamics, which the policy must learn to account for through its observation of each drone's attitude and velocity.

\textbf{Predicates.} The environment provides a single predicate $\mathsf{pos}$ with position parameters $(x, y, z) \in [-10, 10]^3$\,\si{\metre}. A predicate instance $\mathsf{pos}(x, y, z)$ holds if and only if some drone is within Euclidean distance \SI{2}{\metre} of $(x, y, z)$. Its robustness metric is the signed distance of the nearest drone to the boundary of the sphere, normalized by the sphere radius (positive inside the sphere). The predicate is annotated as \textit{non-persistent}. The target positions are drawn jointly and uniformly over the central region of the airspace, subject to a pairwise separation of at least \SI{5.5}{\metre}; the specification's $\mathsf{pos}$ instances take their parameters directly from this draw. Since the separation exceeds twice the sphere radius, targets are disjoint: a single drone cannot satisfy two instances at once, and a conjunction $\mathsf{pos}_i \wedge \mathsf{pos}_j$ can only be satisfied by the two drones splitting across the corresponding spheres.

\subsubsection{FrankaCubesEnv}
\label{app:experimental_details:environments:franka_cubes_env}

\textit{FrankaCubesEnv} loosely expands upon \textit{WarehouseEnv} \citep{jackermeier_zero-shot_2026}, requiring an MJX-simulated Franka Panda manipulator arm with a parallel-jaw gripper to pick and place colored cubes of side length \SI{5}{\centi\metre}. The arm is commanded at its grasp center (the point between the fingertips), which is confined to a box-shaped workspace in front of the base. At the start of each episode, position and orientation targets for the grasp center are drawn, three cubes of distinct colors are placed on the floor at least \SI{0.12}{\metre} apart with uniformly sampled yaws, and the arm is initialized at rest with the gripper open, at a uniformly sampled reachable pose at least \SI{0.18}{\metre} from every position target, at least \SI{0.8}{\radian} from every orientation target, and clear of every cube. No condition terminates an episode early: the arm is confined to its workspace, and no collision is catastrophic. Episodes are truncated after $1000$ policy steps (\SI{50}{\second} of simulated time) during training.  See Figure~\ref{fig:environments:franka} for visualizations of example reset configurations.

\textbf{Observations \& Actions.} The observation comprises: the actual and commanded grasp-center poses, each as a normalized position and a 6D orientation \citep{zhou_continuity_2019}, such that the action displaces the \textit{commanded} pose and the two differ by however far the joint servos lag; the arm's joint angles (as sines and cosines) and joint velocities; the finger opening, with flags reporting whether the gripper is driven open or closed, how long until it may be reversed, whether the last pose increment was feasible, whether a cube is held, and whether a cube sits between the fingertip pads; a scene table listing each cube's position and color; and a pre-processed wrist camera looking along the gripper's approach axis (range \SI{0.4}{\metre}, half-angle \SI{45}{\degree}) that reports the color of one cube (the held cube, or else the nearest cube in view), its offset from the grasp center, its relative orientation, and a detection flag. All components are normalized to $[-1, 1]$. The action comprises a Cartesian pose increment on the commanded grasp-center pose (effectively a twist command, given the fixed control period), of up to \SI{2}{\centi\metre} of translation and \SI{0.1}{\radian} of rotation per step, together with a binary open/close gripper command; a change of gripper state is honored at most once every ten steps (\SI{0.5}{\second}).

\textbf{Low-Level Control.} The commanded pose is resolved once per policy step into targets for the seven arm joints by closed-form inverse kinematics \citep{he_analytical_2021}, which gravity-compensated joint position servos then track. The arm's redundancy is resolved by sweeping the final wrist joint about its current angle and taking the solution requiring the smallest joint motion; solutions moving any joint by more than \SI{0.5}{\radian} (e.g., across a singularity) are rejected, and if no admissible solution exists the previous command is held and the observation flags the increment as infeasible. The physics timestep is \SI{5}{\milli\second}; each policy step holds the joint targets for ten physics steps, inducing a \SI{0.05}{\second} task-level control period. The fingers are driven by a position servo whose closing force is capped at \SI{20}{\newton}, so a closed gripper presses on whatever it meets with a fixed force. A grasp is registered when the closed fingers stall against a cube sitting squarely between the fingertip pads; the grasped cube is then held rigidly to the hand until the gripper opens, standing in for the stiction that MJX's velocity-level friction model lacks.

\textbf{Predicates.} The environment provides three predicates. A predicate instance $\mathsf{pos}(x, y, z)$, with position parameters $(x, y, z)$ ranging over the workspace box $[0.25, 0.77] \times [-0.32, 0.32] \times [0.02, 0.50]$\,\si{\metre}, holds if and only if the grasp center is within Euclidean distance \SI{0.07}{\metre} of $(x, y, z)$; its robustness metric is the signed distance to the boundary of that sphere, normalized by the sphere radius (positive inside the sphere). A predicate instance $\mathsf{rot}(\phi, \theta, \zeta)$, with roll--pitch--yaw parameters $(\phi, \theta, \zeta) \in [-\pi, \pi]^3$, holds if and only if the geodesic angle on $\mathrm{SO}(3)$ between the grasp-center orientation and the target orientation is within a tolerance of \SI{0.3}{\radian}; its robustness metric is the tolerance minus this angle, normalized by the tolerance. A predicate instance $\mathsf{has}(r, g, b)$, with color parameters $(r, g, b) \in [0, 1]^3$, holds if and only if the gripper is holding a cube whose color is within Euclidean distance $0.25$ of $(r, g, b)$; its robustness metric is $+1$ when the instance holds, and otherwise the negative distance from the grasp center to the nearest cube of matching color; if no cube matches, the metric is the negative of the workspace diagonal. $\mathsf{pos}$ and $\mathsf{rot}$ are annotated as \textit{non-persistent}; $\mathsf{has}$ is \textit{persistent}, since a grasped cube travels with the gripper and so remains held through the motion that goes on to satisfy a $\mathsf{pos}$ or $\mathsf{rot}$ conjunct.

Position and orientation targets are drawn uniformly, subject to pairwise separations of \SI{0.18}{\metre} and \SI{0.8}{\radian} respectively and to a joint reachability filter guaranteeing that every drawn position is reachable at every drawn orientation, since a specification may conjoin any pair; positions are drawn over the central region of the workspace, whose bottom face sits at a resting cube's center height. The world's three cube colors are drawn jointly and uniformly over the color cube, subject to a pairwise separation of at least $0.7$; the specification's $\mathsf{has}$ instances take their parameters from this draw, so every named color is stocked by a cube. Since each separation exceeds twice the corresponding tolerance, distinct instances of the same predicate cannot hold simultaneously for the single gripper; satisfiable conjunctions instead mix predicates, e.g., $\mathsf{has}_i \wedge \mathsf{pos}_j \wedge \mathsf{rot}_k$, which requires holding a cube of a named color at a named position and orientation.

\subsubsection{FalloutWorld}
\label{app:experimental_details:environments:fallout_world}

\begin{figure*}[t]
    \centering
    \newcommand\allenvswidth{0.994}
    \newcommand\perenvwidth{0.244}
    \subfloat[\textit{Go1RGBZoneEnv}. The legend marks the predicate instance for each color.\label{fig:environments:go1}]{%
        \makebox[\allenvswidth\linewidth]{%
        \includegraphics[width=\perenvwidth\linewidth]{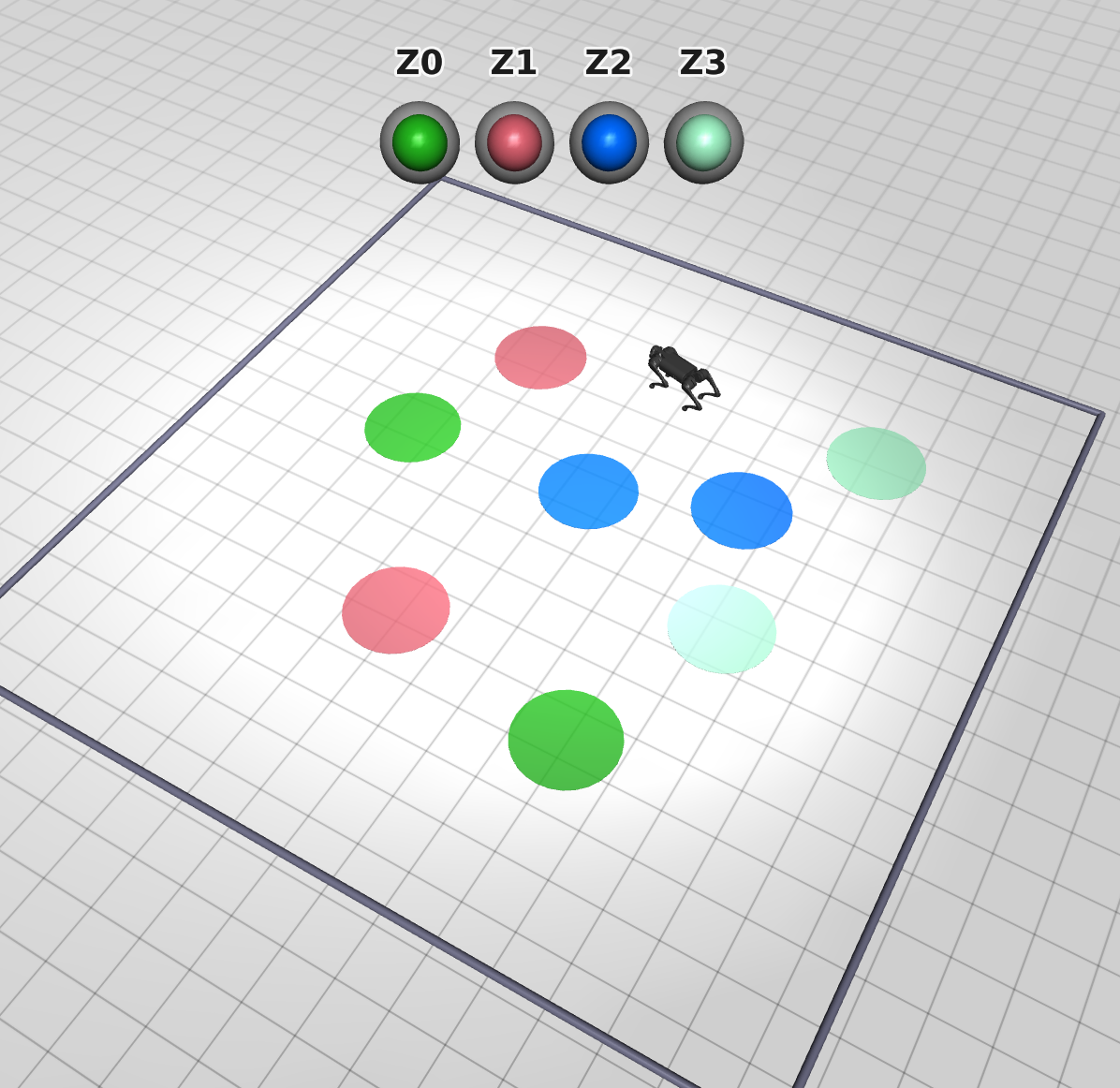}%
        \hfill
        \includegraphics[width=\perenvwidth\linewidth]{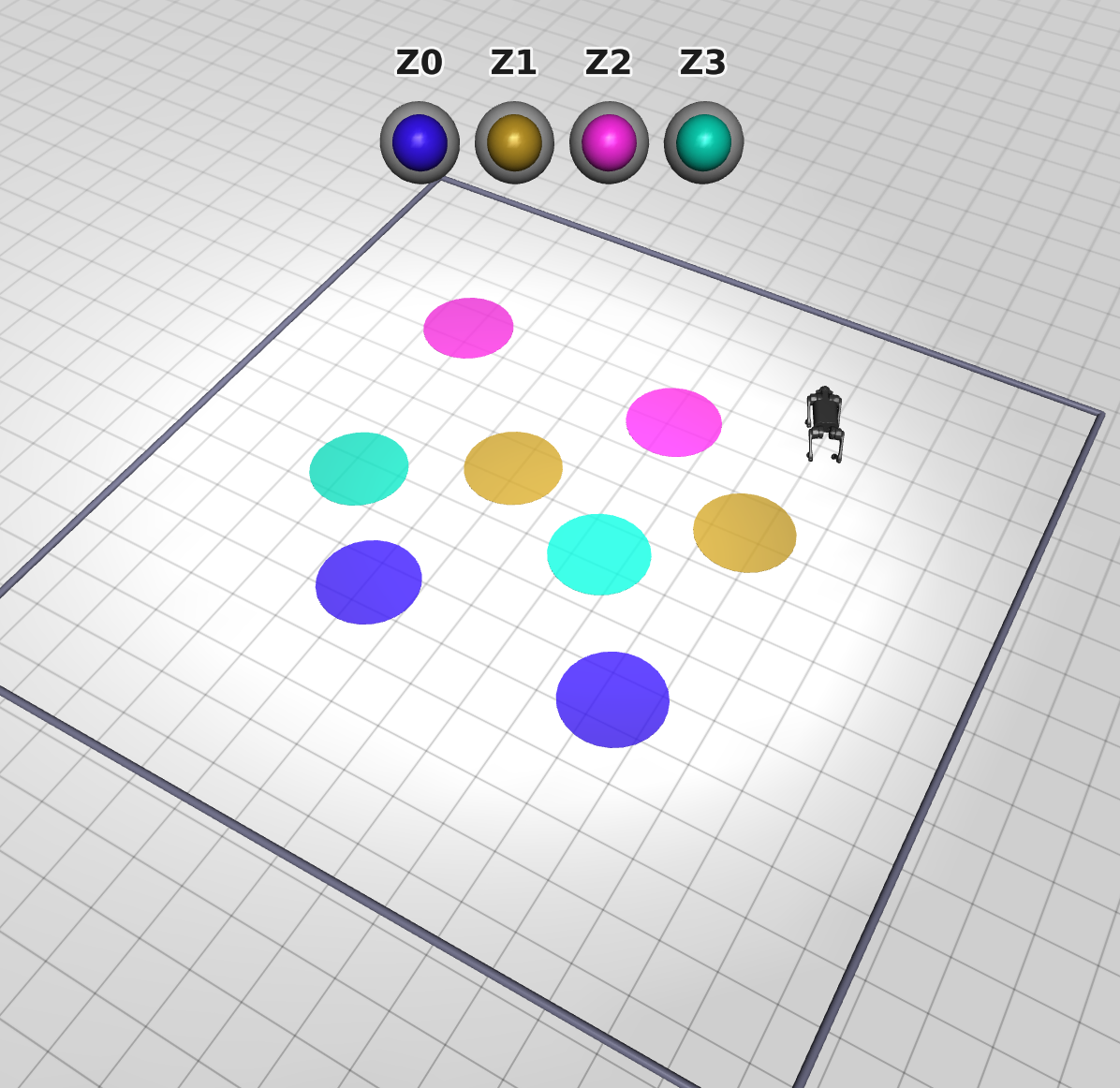}%
        \hfill
        \includegraphics[width=\perenvwidth\linewidth]{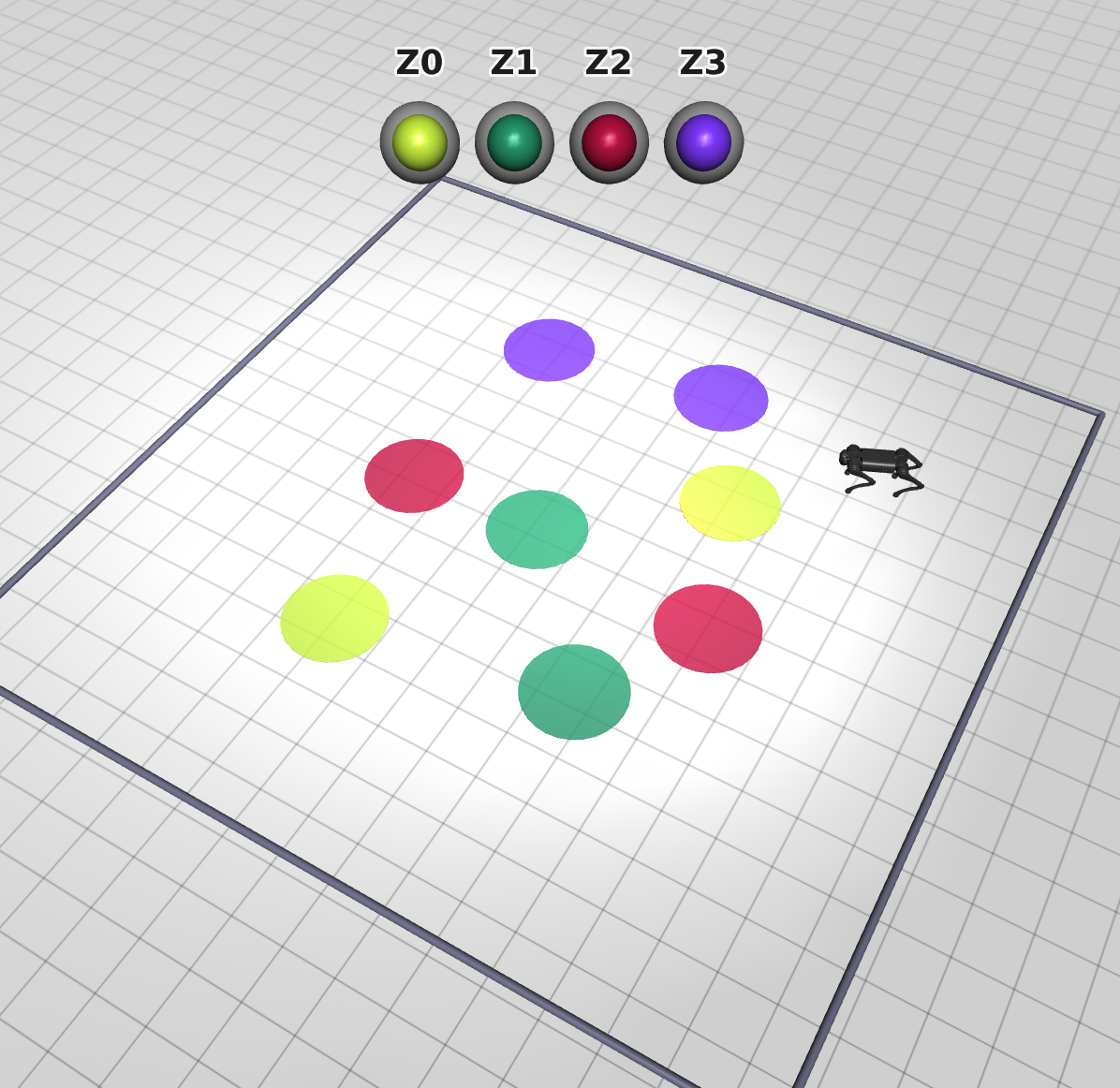}%
        \hfill
        \includegraphics[width=\perenvwidth\linewidth]{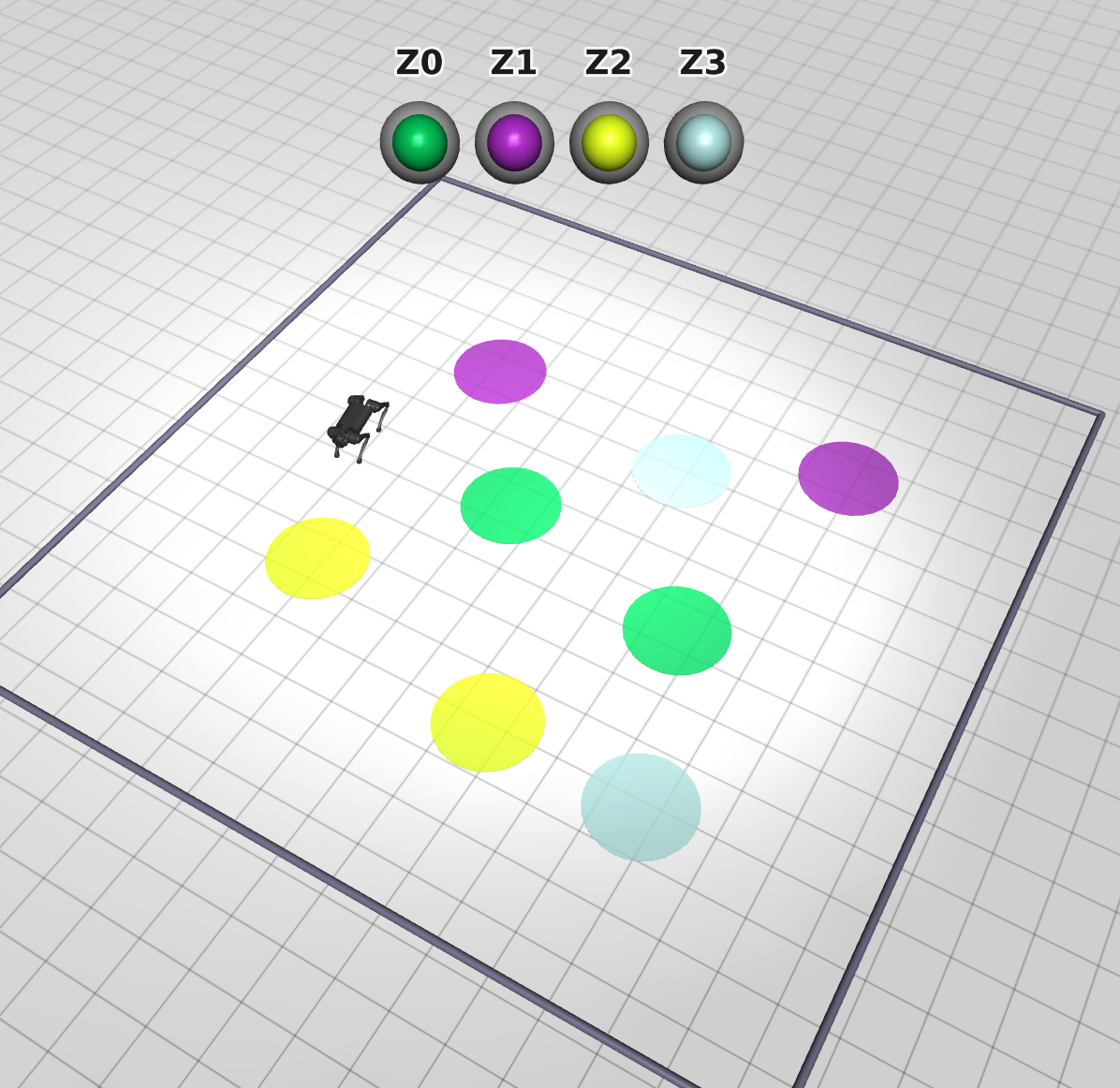}%
        }
    }
    \par\bigskip
    \subfloat[\textit{X2DronesEnv}. The blue and orange circles highlight the two drones.\label{fig:environments:x2}]{%
        \makebox[\allenvswidth\linewidth]{%
        \includegraphics[width=\perenvwidth\linewidth]{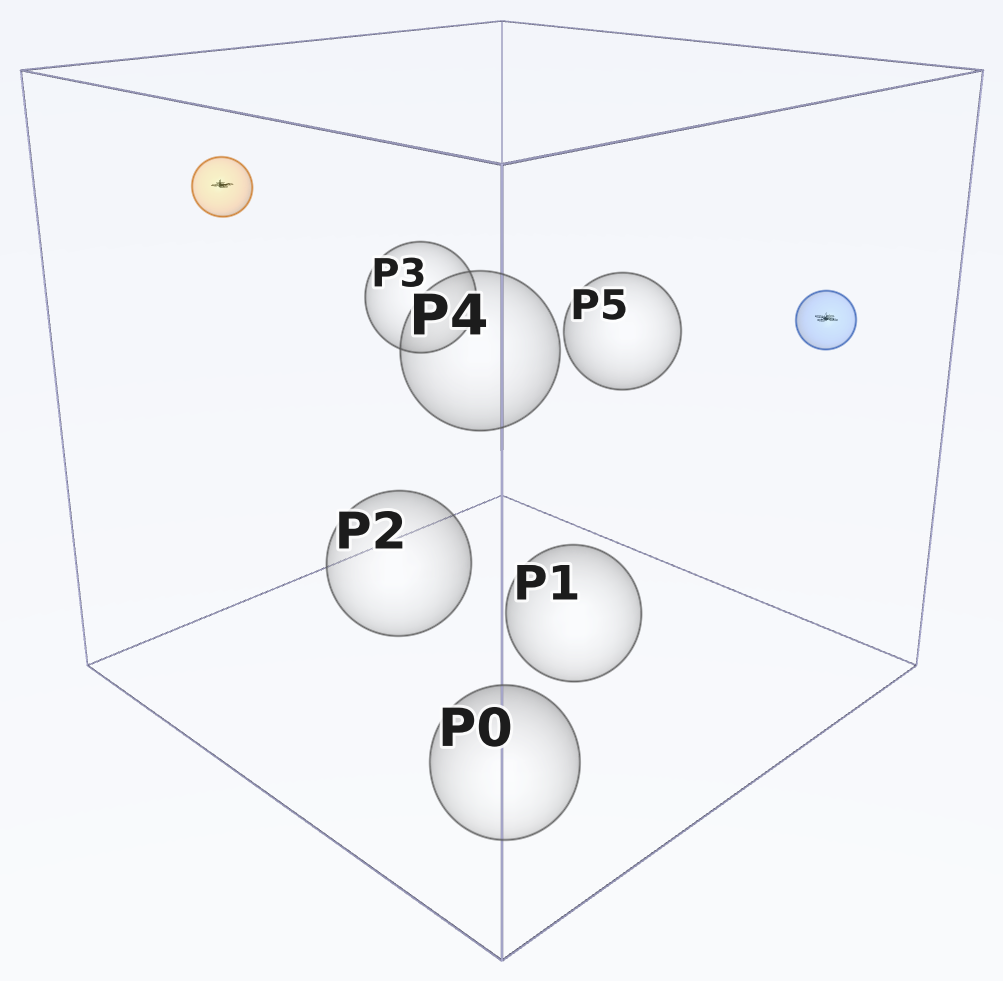}%
        \hfill
        \includegraphics[width=\perenvwidth\linewidth]{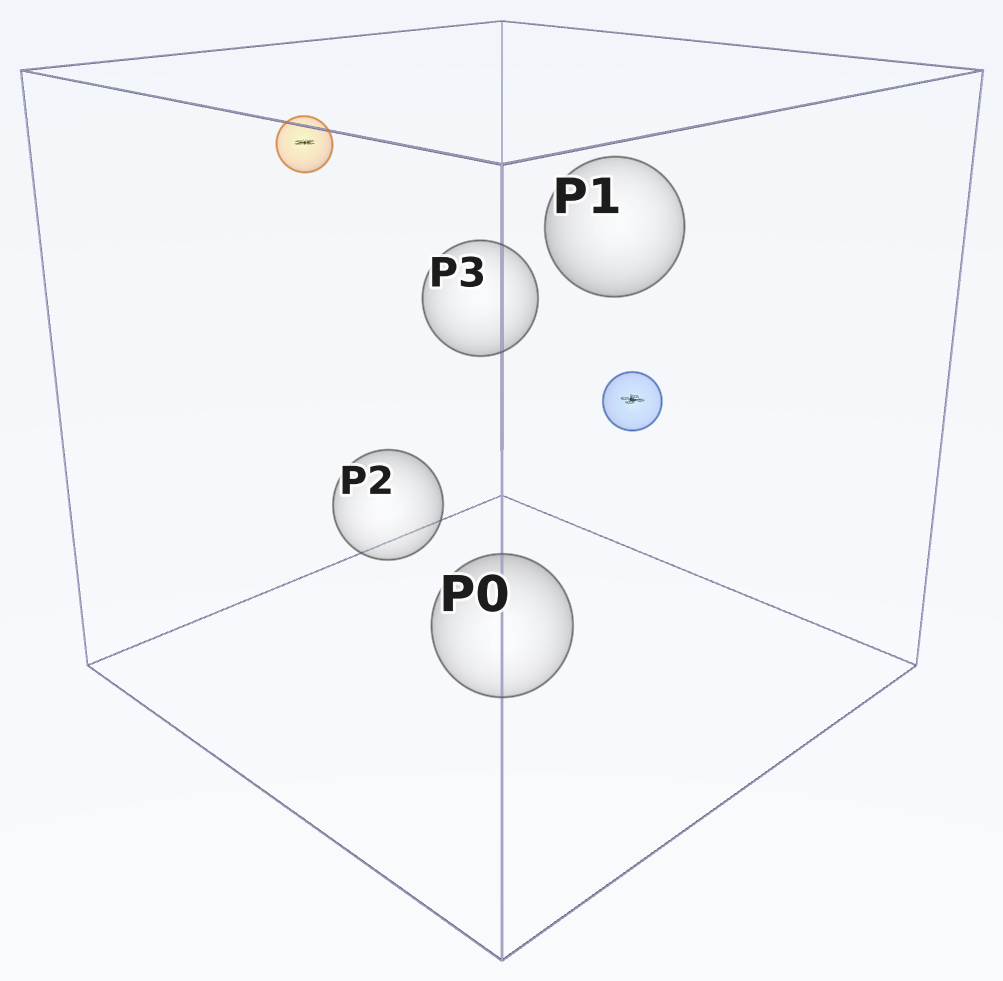}%
        \hfill
        \includegraphics[width=\perenvwidth\linewidth]{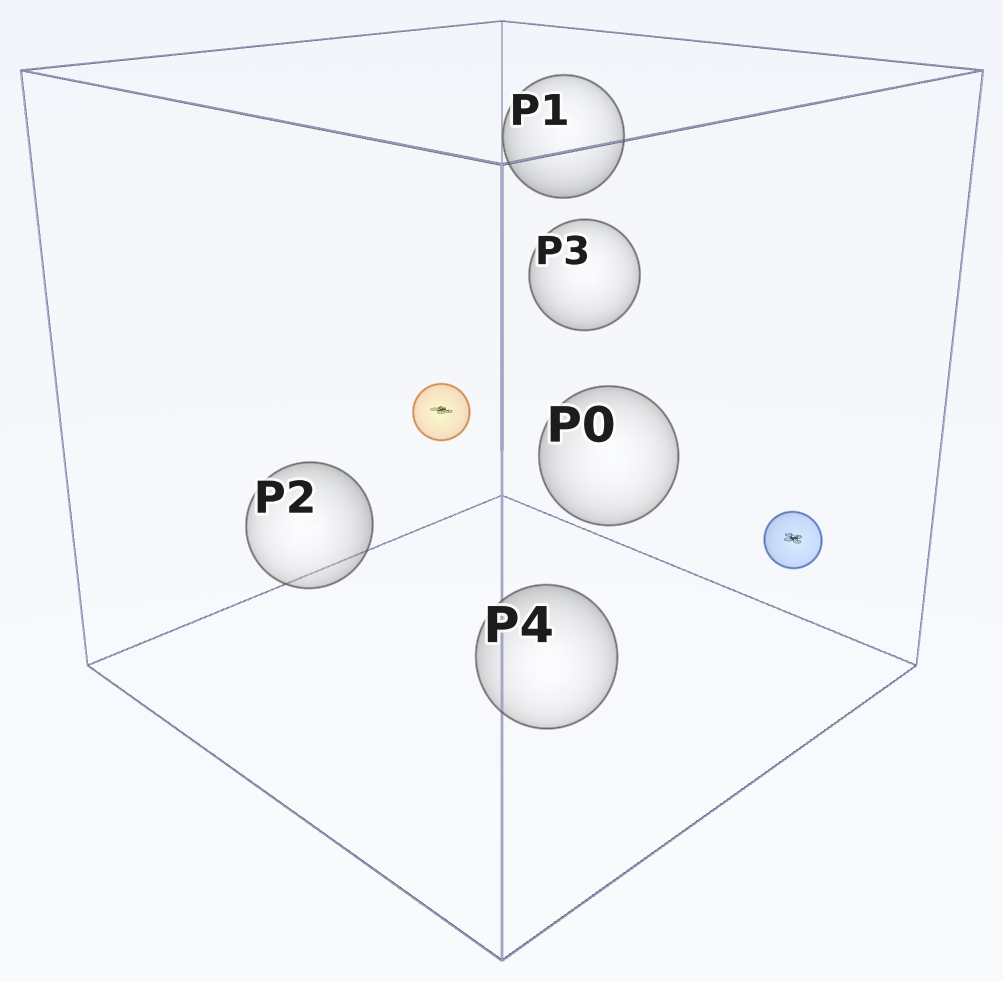}%
        \hfill
        \includegraphics[width=\perenvwidth\linewidth]{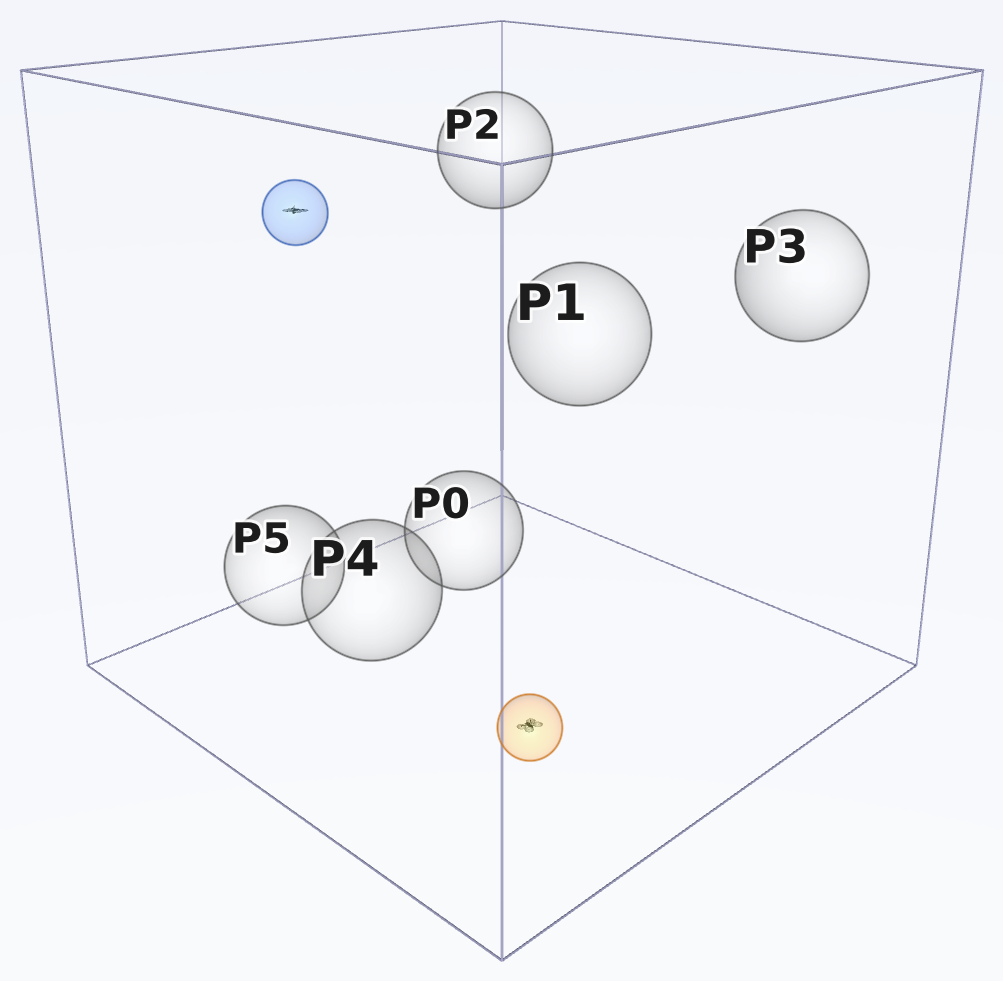}%
        }
    }
    \par\bigskip
    \subfloat[\textit{FrankaCubesEnv}. The legend marks the predicate instance for each color and orientation.\label{fig:environments:franka}]{%
        \makebox[\allenvswidth\linewidth]{%
        \includegraphics[width=\perenvwidth\linewidth]{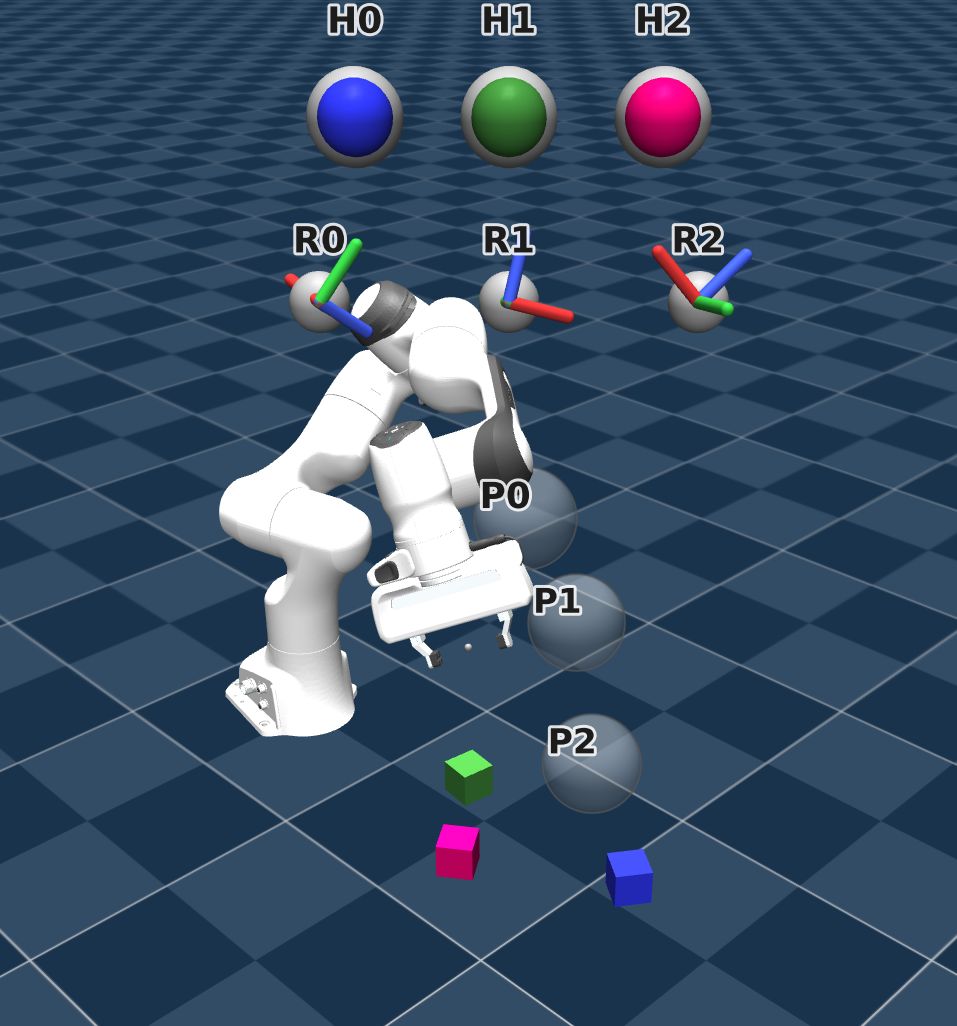}%
        \hfill
        \includegraphics[width=\perenvwidth\linewidth]{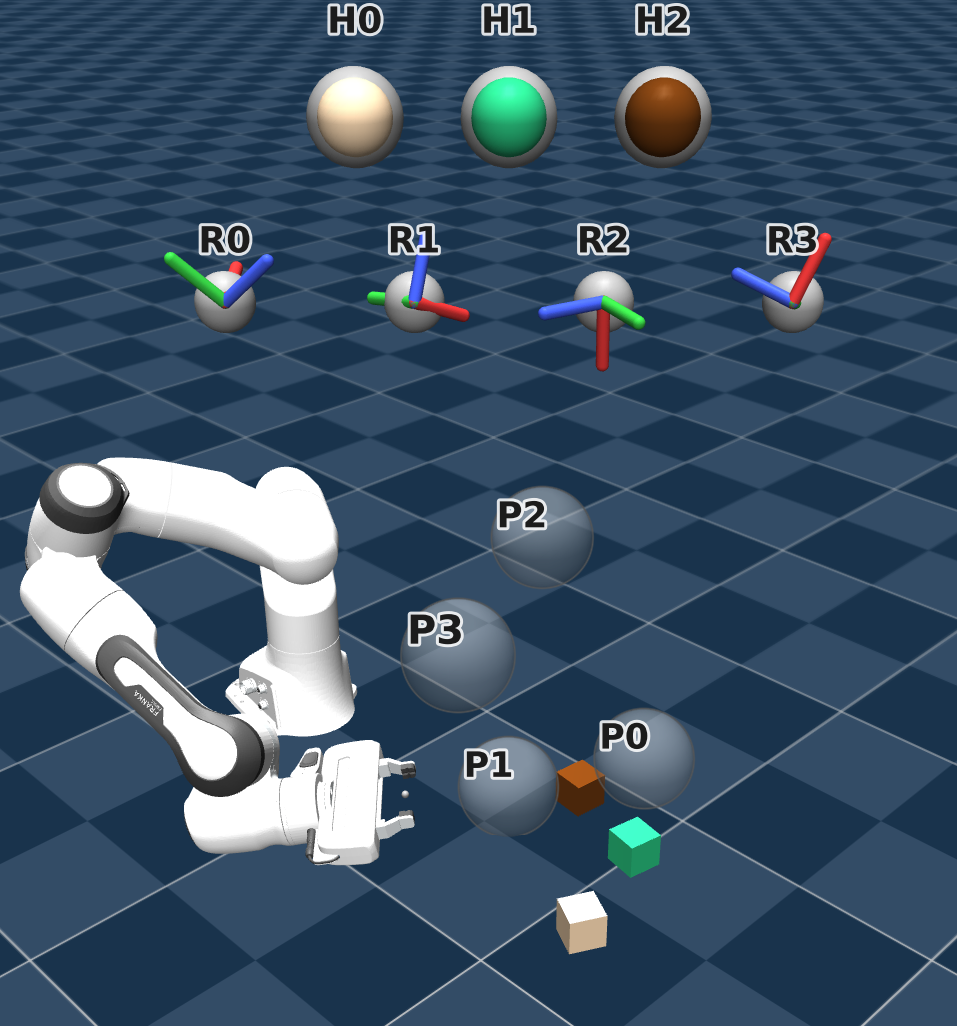}%
        \hfill
        \includegraphics[width=\perenvwidth\linewidth]{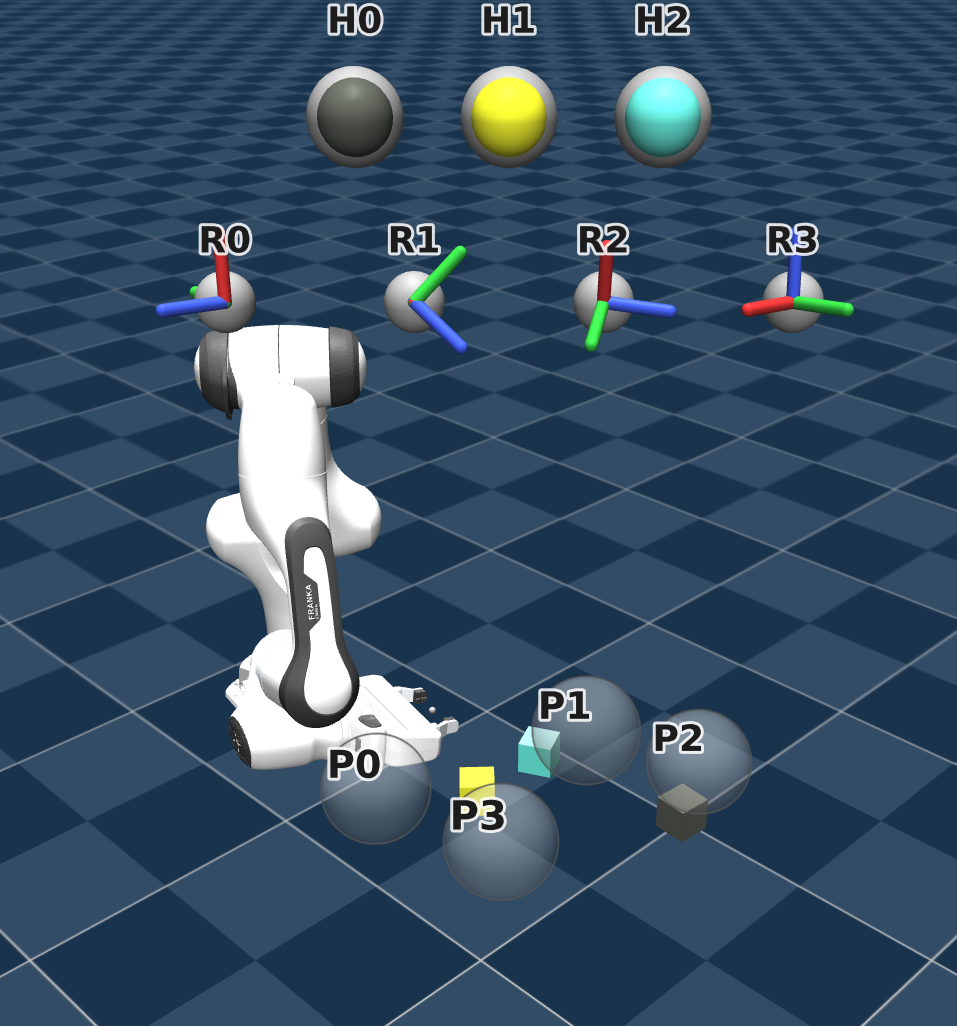}%
        \hfill
        \includegraphics[width=\perenvwidth\linewidth]{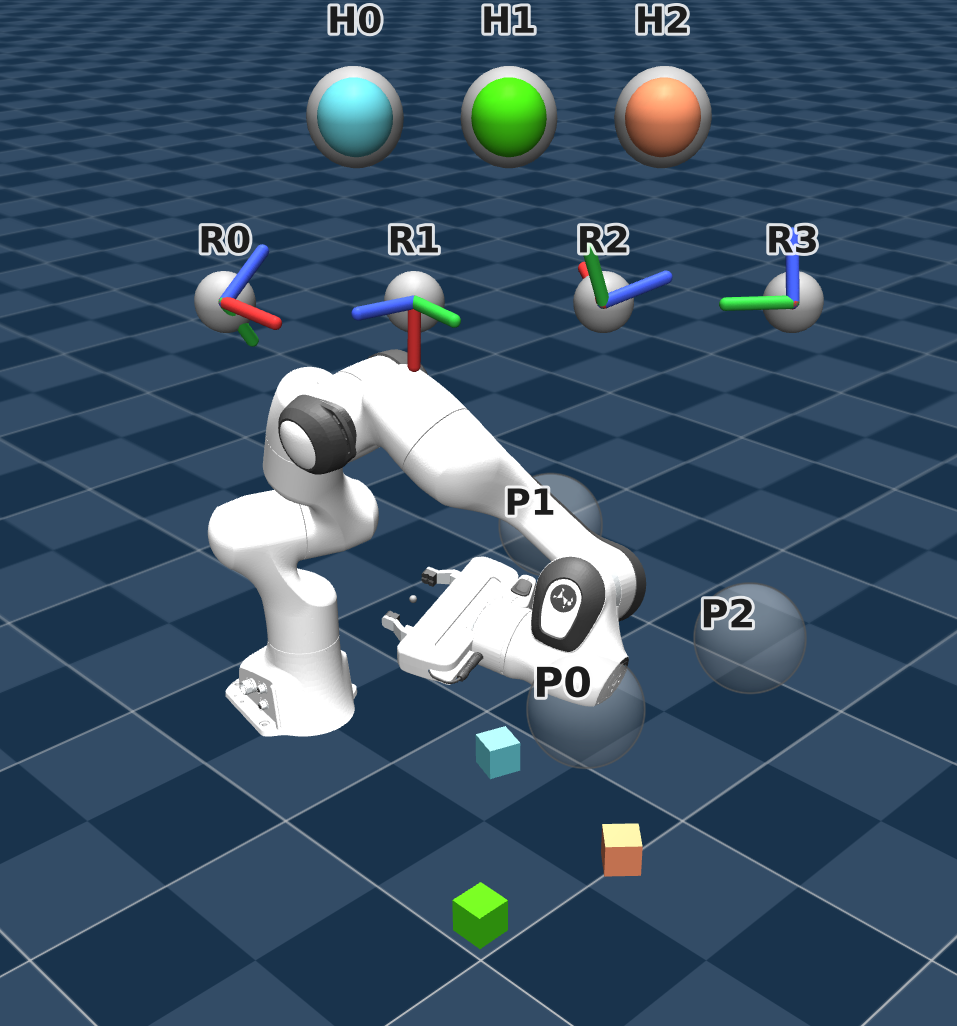}%
        }
    }
    \par\bigskip
    \subfloat[\textit{FalloutWorld}. The blue diamond marks the agent and the dashed borders represent the $\mathsf{rad}$ threshold.\label{fig:environments:fw}]{%
        \makebox[\allenvswidth\linewidth]{%
        \includegraphics[width=\perenvwidth\linewidth]{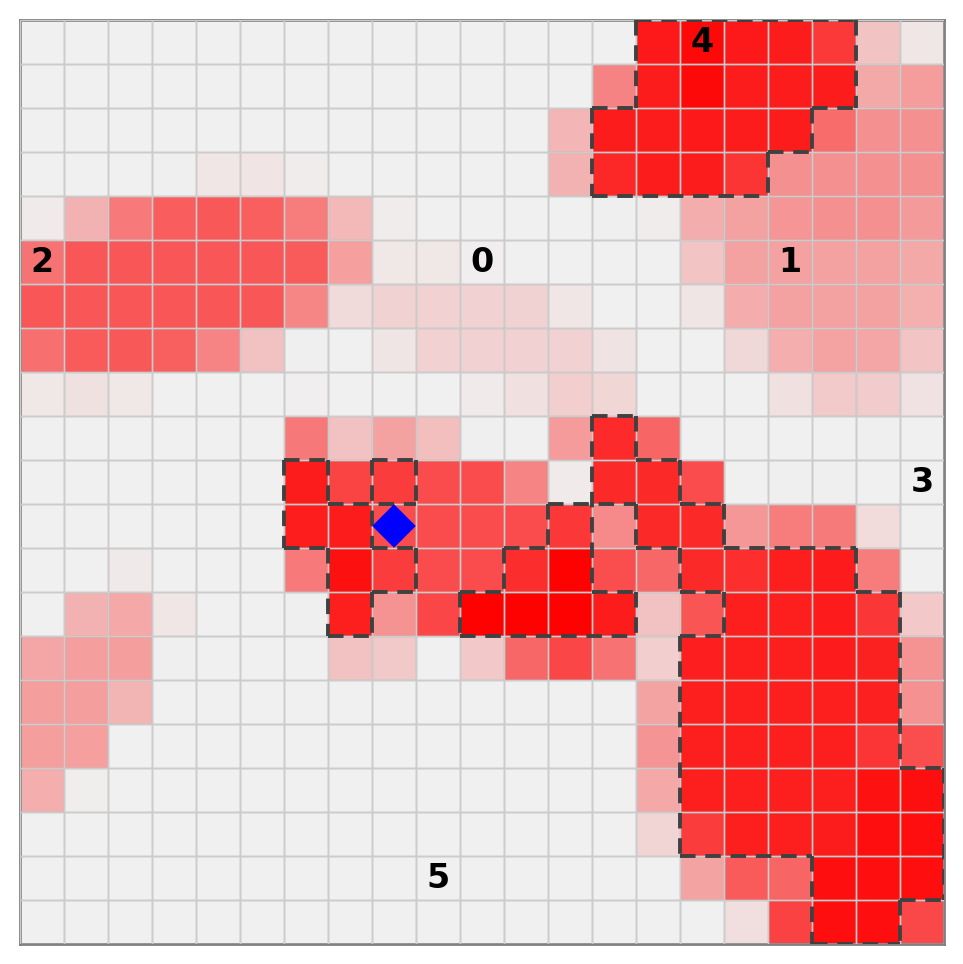}%
        \hfill
        \includegraphics[width=\perenvwidth\linewidth]{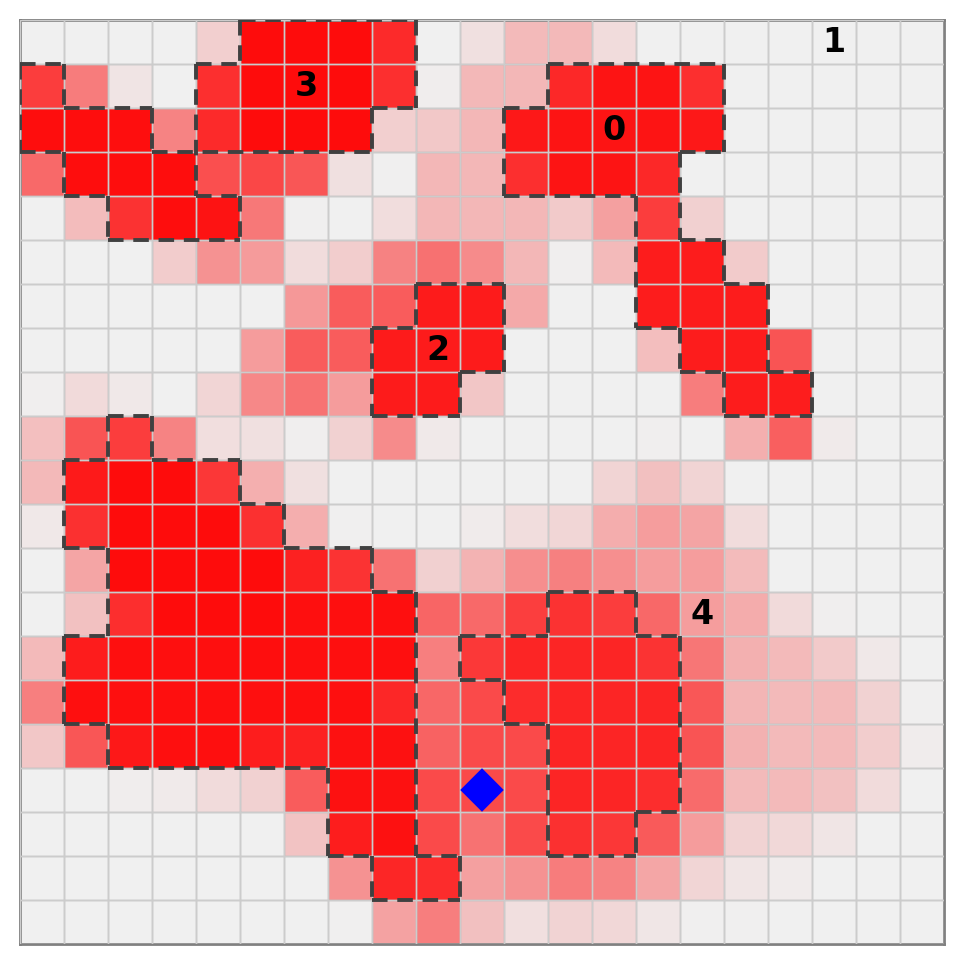}%
        \hfill
        \includegraphics[width=\perenvwidth\linewidth]{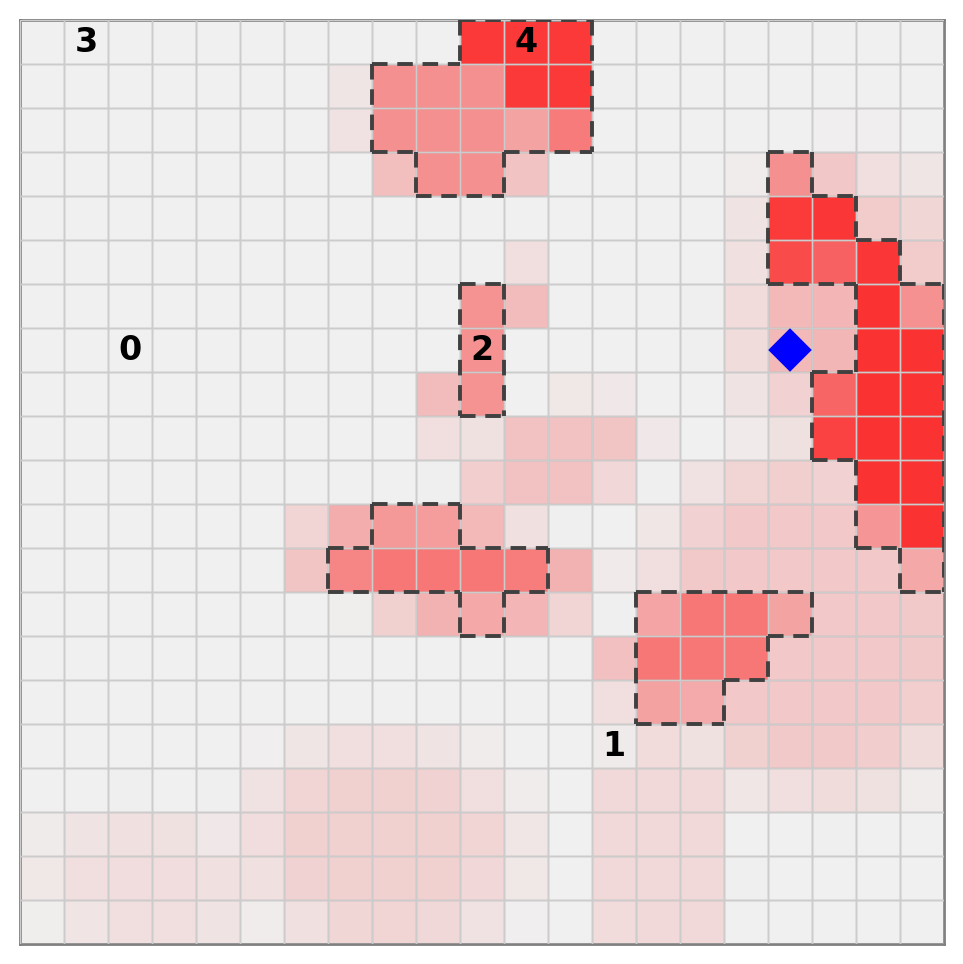}%
        \hfill
        \includegraphics[width=\perenvwidth\linewidth]{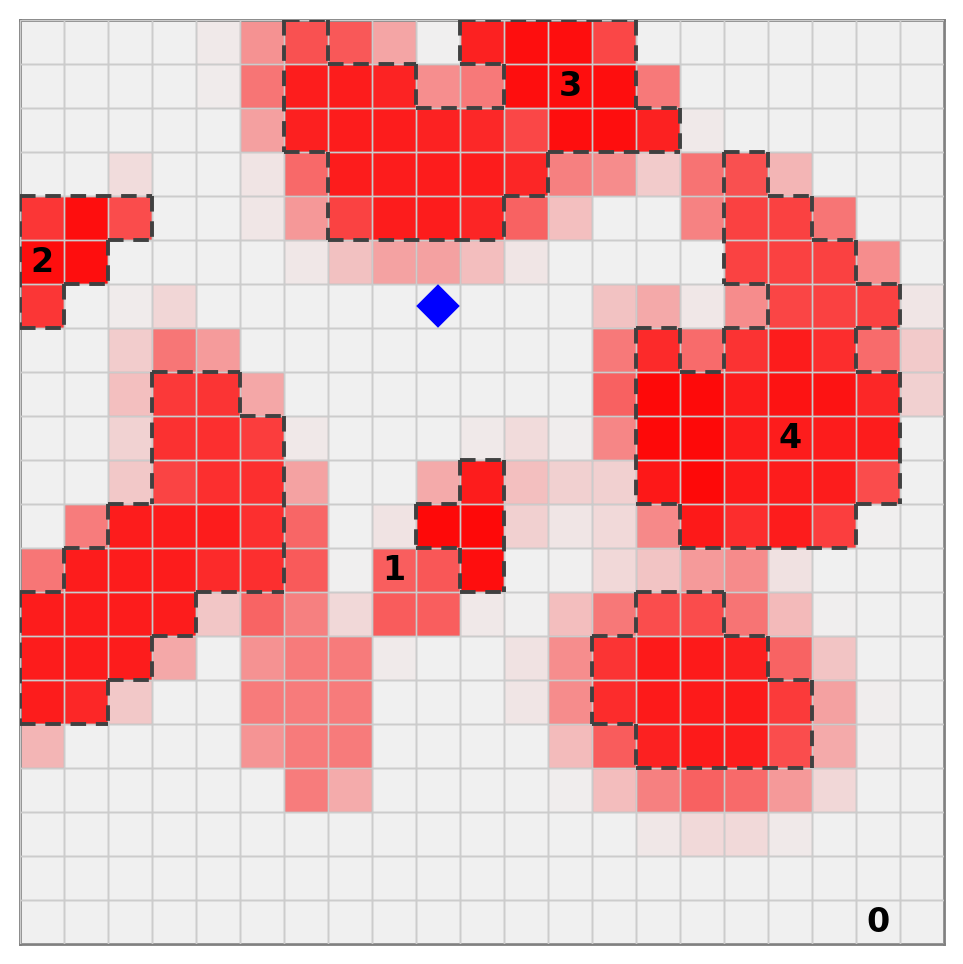}%
        }
    }
    \caption{Visualizations of example environment reset configurations, including predicate instances.}
    \label{fig:environments}
\end{figure*}

\textit{FalloutWorld} expands upon \textit{LetterWorld} \citep{vaezipoor_ltl2action_2021}, requiring an agent to navigate a $21 \times 21$ grid overlaid with a radiation field that assigns each cell a level in $[0, 1]$. The agent occupies a single cell and steps between adjacent cells; specifications direct it to visit target cells in particular orders while avoiding, or seeking, radiation above a named threshold. At the start of each episode, the target cells and the radiation threshold are drawn, a fresh radiation field is generated around them, and the agent spawns at a below-threshold cell. No condition terminates an episode early: the grid is bounded, and radiation affects the labeling function rather than the dynamics. Episodes are truncated after $500$ steps during training.  See Figure~\ref{fig:environments:fw} for visualizations of example reset configurations.

\textbf{Observations \& Actions.} The observation comprises two top-down images: the full grid, with one channel marking the agent's cell and one carrying the radiation field; and a $9 \times 9$ crop of the radiation field centered on the agent, paired with a mask marking which of its cells lie on the grid (off-grid cells read zero), which presents nearby radiation in coordinates relative to the agent. All channels take values in $[0, 1]$. The action selects one of five discrete moves: a step to the adjacent cell in one of the four cardinal directions, or remaining in place; a step off the grid leaves the agent in place.

\textbf{Predicates.} The environment provides two predicates. A predicate instance $\mathsf{loc}(x, y)$, with cell parameters $(x, y)$ ranging over the grid, holds if and only if the agent occupies the cell $(x, y)$; its robustness metric is the negated Manhattan distance from the agent's cell to $(x, y)$. A predicate instance $\mathsf{rad}(\eta)$, with threshold parameter $\eta \in [0.2, 0.8]$, holds if and only if the radiation level at the agent's cell is at least $\eta$; its robustness metric is the negated Manhattan distance to the nearest cell whose radiation level is at least $\eta$. Both predicates are annotated as \textit{non-persistent}.

The target cells are drawn jointly and uniformly over the grid, subject to a pairwise Manhattan separation of at least $7$ cells, and the specification's $\mathsf{loc}$ instances take their parameters directly from this draw; the agent's spawn cell is drawn with the same separation from every target, and the threshold uniformly over its range. Each target cell is then independently assigned, with equal probability, to lie above or below the threshold, and the radiation field is generated to realize this assignment. The field is a composition of smooth elliptical plateaus: ambient plateaus exceeding the threshold populate the grid, corrective plateaus force each target cell to its assigned side of the threshold, and decoy plateaus peaking below the threshold ensure that the presence of a plateau does not by itself reveal whether it exceeds the threshold. The generator guarantees that the spawn cell lies below the threshold, that every below-threshold target is reachable from the spawn without crossing the threshold (a below-threshold corridor is carved through the field where necessary), and that some cell exceeds the threshold, so specifications that avoid radiation and specifications that seek it remain feasible in every world. Since the agent occupies a single cell, distinct $\mathsf{loc}$ instances cannot hold simultaneously; a conjunction $\mathsf{loc}_i \wedge \mathsf{rad}$ is satisfiable in a given world if and only if target $i$ is assigned above the threshold, and holds upon arrival there.

\subsection{Evaluation LTL Specifications}
\label{app:experimental_details:eval_ltl_specifications}

Tables~\ref{tab:formulae:finite} and \ref{tab:formulae:infinite} list the template \textit{finite}-horizon and \textit{infinite}-horizon specifications, respectively. The evaluation specifications for \textit{Go1RGBZoneEnv} are adapted from those used for evaluation in \textit{ZoneEnv} by \citet{jackermeier_deepltl_2025}. For the other environments, the specifications are new and designed to capture the complexity of the new environments, including conjunctions with the same predicates for \textit{X2DronesEnv}, and conjunctions with different predicates for \textit{FrankaCubesEnv}.

For \textit{FrankaCubesEnv}, tasks $\varphi_{22}$--$\varphi_{24}$ require placing cubes at target positions on the floor ($\mathsf{pos}_0$ and $\mathsf{pos}_1$). Accordingly, for those tasks we fix the $z$-parameter for $\mathsf{pos}_0$ and $\mathsf{pos}_1$ to $0.025$, corresponding to the center offset height of a \qty{5}{\cm} cube when resting on the floor.

\subsection{Training Details}
\label{app:experimental_details:training_details}

\subsubsection{Neural Networks}
\label{app:experimental_details:training_details:networks}

\textbf{Proposition Embeddings.} Plato's MLPs $\phi_{\mathsf{f}}$ have a single hidden layer of size $32$ with ReLU activation and a linear output layer of size $n_\phi$. Discretize replaces them with one embedding table per predicate, holding a row of size $n_\phi$ for every cell of the discretized parameter space; parameters map to the nearest bin center under the corresponding domain's metric. The position and color parameter spaces of $\mathsf{pos}$, $\mathsf{at}$, and $\mathsf{has}$ are discretized to $11$ bins per dimension ($11^3 = 1331$ rows each); the orientation space of $\mathsf{rot}$ in \textit{FrankaCubesEnv} to $2048$ orientations selected by farthest-point sampling under the geodesic distance; and in \textit{FalloutWorld}, $\mathsf{loc}$ ranges over all $441$ grid cells and $\mathsf{rad}$ over $13$ threshold bins. Concatenate and SIAMS instead retain learned proposition token embeddings of size $n_\phi$ for the backbone and route the parameters to the observation encoder (described below), gathered into a fixed-width block with one row per predicate instance in the alphabet, zero-padded to a common width, and with the rows of instances absent from the current sequence zeroed. Concatenate appends this block to the observation encoder's input (to the proprioceptive input for \textit{Go1RGBZoneEnv}, and to the flattened convolutional outputs for \textit{FalloutWorld}), while SIAMS passes it through a FiLM MLP with one hidden layer (of size $512$ for \textit{Go1RGBZoneEnv} and \textit{FalloutWorld}, and $256$ for \textit{X2DronesEnv} and \textit{FrankaCubesEnv}) that outputs an element-wise scale and offset applied to the observation encoder's output.

\textbf{StructLTL.} The sequence encoder for the StructLTL backbone (Appendix~\ref{app:struct_ltl}) uses proposition embeddings of size $n_\phi = 32$, a learned vector of the same size embeds the $\varepsilon$-transition, and the negation projection $\Pi$ is a linear layer of size $32$. The post-pooling networks $f_\land$ and $f_\lor$ are each a single linear layer of size $32$ with ReLU activation, and each has a separate learned embedding for the empty set. The reach and avoid embeddings of each sequence step are concatenated into a step embedding of size $64$, and the attention mechanism projects queries, keys, and values to $d_k = 64$, with an ALiBi slope of $m = 5$.

\textbf{Observation Encoders.} For \textit{Go1RGBZoneEnv}, the observation encoder uses a 1D CNN for the 32-bin RGB LiDAR observations, with $[32, 64]$ channels, a kernel size of $5$, a stride of $1$, circular padding, and an average-pooling layer with kernel size and stride $2$ after each convolution; a linear layer of size $64$ embeds the proprioceptive observations, and the concatenated outputs pass through a linear layer of size $512$. For \textit{X2DronesEnv} and \textit{FrankaCubesEnv}, the observation encoder is an MLP with hidden layer sizes $[256, 256]$ and an output size of $256$. For \textit{FalloutWorld}, the observation encoder applies two independent 2D CNNs, one to the full-grid image, with $[8, 8, 2]$ channels, and one to the egocentric crop, with $[16, 16, 4]$ channels; both use $3 \times 3$ kernels with stride $1$ and unit zero-padding except for a final $1 \times 1$ convolution, and their flattened outputs are concatenated. All observation encoders use ReLU internal activations.

\textbf{Actor-Critic.} For all environments, the actor and critic are MLPs with hidden layer sizes $[512, 512]$ and ReLU internal activations, operating on the concatenation of the observation embedding and the goal embedding from the backbone. In the continuous environments, the actor has a $\tanh$ output layer for the mean and standard deviation of a diagonal Gaussian distribution over the action space (the standard deviation then passes through a softplus activation to remain strictly positive), alongside a logit for the $\varepsilon$-action. For \textit{FrankaCubesEnv}, whose action space pairs a continuous grasp-center pose command with a discrete gripper command, the actor adds a categorical output layer for the gripper, and the policy factorizes over the two distributions. In the discrete \textit{FalloutWorld}, the actor has a softmax output layer for a categorical distribution of size $|\mathcal{A}| + 1$, where the additional logit represents the $\varepsilon$-action. The critic outputs the predicted value of the value function.

\textbf{LTL2Action.} The formula encoder for the LTL2Action backbone (Appendix~\ref{app:ablation_studies:ltl2action}) uses proposition and operator token embeddings of size $n_\phi = 32$, and the syntax-tree encoder is an RGCN with $8$ message-passing rounds, weights shared across rounds, three edge relation types, and $\tanh$ activations; the root node's embedding after the final round serves as the specification embedding. The observation encoders, actor, and critic match those above, except that the actor omits the $\varepsilon$-logit, since LTL progression admits no $\varepsilon$-transitions. The proposition embeddings under Plato and the baselines are produced as for StructLTL.

\textbf{Model Size.} For our environments, each MLP $\phi_{\mathsf{f}}$ has $1120$--$1184$ parameters, and the complete Plato embedding module between $1184$ (\textit{Go1RGBZoneEnv} and \textit{X2DronesEnv}, one predicate each) and $3552$ (\textit{FrankaCubesEnv}, three predicates). The overall policy networks comprise between \qty{1.0}{M} and \qty{1.9}{M} parameters, dominated by the actor, critic, and observation encoders; the StructLTL sequence encoder accounts for roughly $\num{20000}$. The Plato embedding module thus makes up less than $0.4\%$ of the overall policy network in every environment, in line with the claim of Section~\ref{sec:plato}.

\subsubsection{Hyperparameters}
\label{app:experimental_details:training_details:hyperparameters}

We use PPO \citep{schulman_proximal_2017} for training. The hyperparameters for PPO and P\textsuperscript{2}BRS are listed in Table~\ref{tab:ppo_params}; all methods use the same hyperparameters. We use Adam \citep{kingma_adam_2015} for all experiments. The choice of RL algorithm is not critical to PlatoLTL; while other algorithms could be used, we opted for PPO as it is the algorithm originally used for the StructLTL and LTL2Action backbones, and it generally achieves fast wall-clock training time.

\begin{table*}[t]
    \centering
    \caption{Hyperparameters for PPO and P\textsuperscript{2}BRS. Values spanning multiple columns indicate identical values across those environments.}
    \label{tab:ppo_params}

    \newcommand{\linefill}{%
      \leavevmode\leaders\hrule height 2.4pt depth -2.0pt\hfill\kern0pt%
    }
    \newcommand{\spanval}[1]{\linefill\enspace #1\enspace\linefill}

    \begin{scriptsize}
    \begin{tabular}{llcccc}
        \toprule
        & & \multicolumn{3}{c}{Continuous} & Discrete \\
        \cmidrule(lr){3-5} \cmidrule(lr){6-6}
        & Parameter & Go1RGBZoneEnv & X2DronesEnv & FrankaCubesEnv & FalloutWorld \\
        \midrule

        \multirow{14}{*}{\rotatebox[origin=c]{90}{PPO}}
        
        & Number of parallel envs & \multicolumn{4}{c}{\spanval{$2048$}} \\
        & Steps per env per update & \multicolumn{4}{c}{\spanval{$64$}}  \\
        & Epochs & \multicolumn{3}{c}{\spanval{$5$}} & $4$ \\
        & Batch size & \multicolumn{3}{c}{\spanval{$32768$}} & $8192$ \\
        & Discount factor & \multicolumn{3}{c}{\spanval{$0.995$}} & $0.99$ \\
        & GAE $\lambda$ & \multicolumn{4}{c}{\spanval{$0.95$}} \\
        & Entropy coefficient & \multicolumn{3}{c}{\spanval{$0.003$}} & $0.03$ \\
        & Value loss coefficient & \multicolumn{4}{c}{\spanval{$0.5$}} \\
        & Max gradient norm & \multicolumn{4}{c}{\spanval{$1.0$}} \\
        & Clip $\epsilon$ & \multicolumn{4}{c}{\spanval{$0.2$}} \\
        & Adam learning rate & \multicolumn{4}{c}{\spanval{\num{5e-4}}} \\
        & Adam $\epsilon$ & \multicolumn{4}{c}{\spanval{\num{1e-8}}} \\
        & Total env interactions & \multicolumn{2}{c}{\spanval{\num{1e8}}} & \num{5e8} & \num{1e8} \\
        & Maximum episode length & \multicolumn{3}{c}{\spanval{$1000$}} & $500$ \\
        
        \midrule

        \multirow{8}{*}{\rotatebox[origin=c]{90}{P\textsuperscript{2}BRS}}
        
        & $r_\mathrm{rch}$ & \multicolumn{4}{c}{\spanval{$0.4$}} \\
        & $r_\mathrm{avd}$ & \multicolumn{4}{c}{\spanval{$0.1$}} \\
        & $\sigma$ & \multicolumn{4}{c}{\spanval{$0.5$}} \\
        & $q_g$ / $q_\kappa$ & \multicolumn{4}{c}{\spanval{$0.1$}} \\
        & $n_g$ / $n_\kappa$ & \multicolumn{4}{c}{\spanval{$131072$}} \\
        & $\lambda_\kappa$ & \multicolumn{4}{c}{\spanval{$0.03$}} \\
        & $\kappa_\mathrm{min}$ & \multicolumn{4}{c}{\spanval{$0.1$}} \\
        & $\kappa_\mathrm{max}$ & \multicolumn{4}{c}{\spanval{$10.0$}} \\
        \bottomrule
    \end{tabular}
    \end{scriptsize}
    \vskip -0.1in
\end{table*}

\subsubsection{Curriculum}
\label{app:experimental_details:training_details:curriculum}

For the StructLTL backbone, we use \textit{curriculum learning} to gradually introduce more challenging sequences during training, starting with one-step reach tasks and increasing to multi-step reach-avoid and reach-stay tasks by the final curriculum stage. Table~\ref{tab:curriculum} presents our generic $7$-stage training curriculum shared by all environments. Reach-avoid sequences take the form $\big((\beta^+_i, \beta^-_i)\big)^{n-1}_{i=0}$, while reach-stay sequences take the form $\big((\varepsilon, \beta^-), (\beta^+, \lnot \beta^+), (\beta^+, \lnot \beta^+), \dots\big)$. The curriculum structure is deliberate: stage $1$ introduces reaching propositions, stage $2$ extends this to clauses, stage $3$ introduces a ``narrow'' set of clauses to avoid, stage $4$ extends this to a ``wide'' set of disjunctions, stage $5$ introduces multi-step sequences, stage $6$ introduces reach-stay tasks, and stage $7$ serves as the final training distribution. For environments without persistence ($\F \G$) tasks in the evaluation set, stages $6$ and $7$ are excluded from training. To progress to the next stage, success rate must be greater than or equal to $0.95$, measured over a sliding window of the $512$ most recent completed episodes.

Because LTL2Action (used for Appendix~\ref{app:ablation_studies:ltl2action}) conditions the policy on the task formula directly, progressing it after each environment step, it cannot consume the reach-avoid sequences our standard curriculum produces; for \textit{Go1RGBZoneEnv} and \textit{X2DronesEnv} we use small bespoke curricula instead. These roughly mirror the standard ladder's structure. Tasks are sequential reach-avoid formulae of the form $\mathsf{F}\big(r_1 \wedge \mathsf{F}(r_2 \wedge \cdots)\big)$, where any $\mathsf{F}(r_i \wedge \cdot)$ may be replaced by $\neg a_i \mathsf{U} (r_i \wedge \cdot)$ to place a guard $a_i$ on the $i$-th step. The subgoals $r_i$ and guards $a_i$ are sampled from propositions and disjunctions of propositions, plus conjunctions of propositions for \textit{X2DronesEnv}, with increasing depth and number of disjunctions (and conjunctions) in later stages.

%%%%%%%%%%%%%%%%%%%%%%%%%%%%%%%%
% INFINITE-HORIZON RESULTS
%%%%%%%%%%%%%%%%%%%%%%%%%%%%%%%%

\section{Evaluation Results for Infinite-Horizon LTL Specifications}
\label{app:infinite}

Table~\ref{tab:results_infinite} presents the average number of visits for \textit{infinite}-horizon LTL specifications. Note that specifications $\psi_{1}$--$\psi_{4}$ yield hundreds of visits on average; this is because these involve \textit{persistence} ($\F \G$) tasks, and once the policy makes the $\varepsilon$-transition, it completes a visit at every timestep so long as the specification is not violated. The remaining infinite-horizon LTL specifications involve \textit{recurrence} ($\G \F$) tasks, for which visits occur much more sparsely. Note also that $\psi_{11}$, $\psi_{18}$, $\psi_{19}$, and $\psi_{24}$ have a constant offset of $+1$ to average visits; this is a quirk of the LDBAs constructed for these tasks by Rabinizer, which yield a visit upon the first automaton transition regardless of MDP state.

As with the \textit{finite} tasks, we see strong performance on \textit{infinite} tasks from Plato in \textit{Go1RGBZoneEnv} and \textit{FalloutWorld} with considerably poorer performance from the baselines, as well as non-zero visits by Plato in \textit{X2DronesEnv} while the baselines make no progress. Applying P\textsuperscript{2}BRS to Plato significantly increases average visits in \textit{X2DronesEnv}, and yields even more dramatic improvement in \textit{FrankaCubesEnv}, achieving high numbers of average visits for tasks with otherwise near zero visits under unguided Plato. Again, we see little to no gains from P\textsuperscript{2}BRS in \textit{Go1RGBZoneEnv} and \textit{FalloutWorld}, with the exception of persistence tasks ($\psi_{1}$--$\psi_{4}$), which see a mild boost on average, likely because the dense signal guides the agent to stay within the colored zones.

\begin{table*}[t]
    \caption{Evaluation results for \textit{infinite}-horizon LTL specifications.}
    \label{tab:results_infinite}
    \begin{center}
    \begin{tiny}

    % Reduce column spacing
    % Default is 6pt. 2pt is usually tight enough to fit 12 cols.
    \setlength{\tabcolsep}{6pt}

    \makeatletter
    \newcommand{\tightscripts}{%
      \fontdimen14\textfont2=1.4pt   % superscript shift up
      \fontdimen16\textfont2=1.0pt   % subscript shift down
      \fontdimen17\textfont2=1.0pt   % subscript shift down (with superscript)
      \fontdimen8\textfont3=0.12pt   % min sup/sub gap = 4x this
    }
    \makeatother

    \newcommand{\res}[3]{\ensuremath{\tightscripts #1^{+#2}_{-#3}}}

    \setlength{\aboverulesep}{0.3ex}
    \setlength{\belowrulesep}{0.3ex}

    \begin{tabular}{ll ccccc}
        \toprule

        & & \multicolumn{5}{c}{Average Visits ($\uparrow$)} \\
        \cmidrule(lr){3-7}

        & $\psi$ & Discretize & Concatenate & SIAMS & Plato & Plato + P\textsuperscript{2}BRS \\
        \midrule

        % ================= Go1RGBZoneEnv =================
        \multirow{6}{*}{\rotatebox[origin=c]{90}{Go1RGBZoneEnv}}
        % psi_1 = FG _at_0_
        & $\psi_{1}$
        & \res{23.1}{3.5}{2.2} & \res{303.0}{206.5}{148.6} & \res{32.1}{29.6}{9.1} & \res{837.9}{59.3}{57.3} & \res{\mathbf{871.8}}{76.6}{78.4} \\

        % psi_2 = FG _at_0_ & F (_at_1_ & F _at_2_)
        & $\psi_{2}$
        & \res{22.3}{3.3}{1.9} & \res{293.4}{205.6}{145.7} & \res{30.2}{27.0}{8.1} & \res{808.8}{56.1}{59.6} & \res{\mathbf{875.6}}{81.9}{73.9} \\

        % psi_3 = FG _at_0_ & G !_at_1_
        & $\psi_{3}$
        & \res{11.7}{1.7}{0.9} & \res{253.2}{185.8}{127.6} & \res{26.1}{27.3}{8.2} & \res{816.3}{71.6}{53.8} & \res{\mathbf{848.1}}{82.1}{81.7} \\

        % psi_4 = G ((_at_0_ | _at_1_) => F _at_2_) & FG (_at_0_ | _at_3_)
        & $\psi_{4}$
        & \res{18.5}{2.0}{1.4} & \res{213.6}{155.2}{109.5} & \res{34.8}{25.3}{7.8} & \res{787.1}{57.3}{65.9} & \res{\mathbf{836.4}}{73.4}{89.5} \\

        % psi_5 = GF _at_0_ & GF _at_1_
        & $\psi_{5}$
        & \res{10.2}{0.6}{0.7} & \res{16.2}{1.7}{1.0} & \res{12.8}{1.2}{1.7} & \res{29.2}{1.5}{2.1} & \res{\mathbf{29.9}}{0.8}{0.8} \\

        % psi_6 = GF _at_0_ & GF _at_1_ & GF _at_2_ & G !_at_3_
        & $\psi_{6}$
        & \res{0.3}{0.1}{0.1} & \res{5.5}{0.5}{0.5} & \res{3.1}{0.6}{0.6} & \res{13.6}{0.7}{0.8} & \res{\mathbf{14.2}}{0.3}{0.7} \\

        \midrule

        % ================= X2DronesEnv =================
        \multirow{6}{*}{\rotatebox[origin=c]{90}{X2DronesEnv}}
        % psi_7 = G F _pos_0_ & G F _pos_1_ & G F _pos_2_ & G F _pos_3_ & G F _pos_4_ & G F _pos_5_
        & $\psi_{7}$
        & \res{0.0}{0.0}{0.0} & \res{0.0}{0.0}{0.0} & \res{0.0}{0.0}{0.0} & \res{4.0}{0.2}{0.2} & \res{\mathbf{4.5}}{0.2}{0.2} \\

        % psi_8 = G F (_pos_0_ & _pos_1_) & G F _pos_2_ & G !(_pos_3_ | _pos_4_)
        & $\psi_{8}$
        & \res{0.0}{0.0}{0.0} & \res{0.0}{0.0}{0.0} & \res{0.0}{0.0}{0.0} & \res{3.3}{0.7}{1.0} & \res{\mathbf{6.3}}{0.3}{0.4} \\

        % psi_9 = G F (_pos_0_ & X F (_pos_1_ & X F (_pos_2_ & X F (_pos_1_ & X F _pos_0_)))) & G !_pos_3_
        & $\psi_{9}$
        & \res{0.0}{0.0}{0.0} & \res{0.0}{0.0}{0.0} & \res{0.0}{0.0}{0.0} & \res{5.6}{0.8}{1.4} & \res{\mathbf{8.0}}{0.2}{0.4} \\

        % psi_10 = G F (_pos_0_ & _pos_1_) & G F (_pos_2_ & _pos_3_) & F (_pos_4_ & F (_pos_0_ & F _pos_2_)) & G !_pos_5_
        & $\psi_{10}$
        & \res{0.0}{0.0}{0.0} & \res{0.0}{0.0}{0.0} & \res{0.0}{0.0}{0.0} & \res{1.5}{0.5}{0.5} & \res{\mathbf{5.0}}{0.2}{0.3} \\

        % psi_11 = G F _pos_0_ & G F _pos_1_ & G F _pos_2_ & G ((_pos_3_ | _pos_4_) => !_pos_0_ U _pos_1_)
        & $\psi_{11}$
        & \res{1.0}{0.0}{0.0} & \res{1.0}{0.0}{0.0} & \res{1.0}{0.0}{0.0} & \res{3.8}{0.5}{0.5} & \res{\mathbf{4.7}}{0.2}{0.2} \\

        % psi_12 = G F (_pos_0_ & _pos_1_) & G F (_pos_2_ & _pos_3_) & G !(_pos_0_ & _pos_2_)
        & $\psi_{12}$
        & \res{0.0}{0.0}{0.0} & \res{0.0}{0.0}{0.0} & \res{0.0}{0.0}{0.0} & \res{1.6}{0.5}{0.6} & \res{\mathbf{5.1}}{0.2}{0.2} \\

        \midrule

        % ================= FrankaCubesEnv =================
        \multirow{6}{*}{\rotatebox[origin=c]{90}{FrankaCubesEnv}}
        % psi_13 = GF _pos_0_ & GF _pos_1_ & GF _pos_2_ & GF _pos_3_
        & $\psi_{13}$
        & \res{0.1}{0.0}{0.0} & \res{0.0}{0.0}{0.0} & \res{0.0}{0.0}{0.0} & \res{19.6}{1.9}{2.0} & \res{\mathbf{29.7}}{1.2}{1.7} \\

        % psi_14 = GF _pos_0_ & GF _pos_1_ & GF _pos_2_ & G !_pos_3_ & (!_pos_1_ U _pos_0_) & (!_pos_2_ U _pos_1_)
        & $\psi_{14}$
        & \res{0.1}{0.0}{0.0} & \res{0.0}{0.0}{0.0} & \res{0.0}{0.0}{0.0} & \res{14.6}{4.5}{6.1} & \res{\mathbf{33.9}}{2.2}{1.6} \\

        % psi_15 = GF (_pos_0_ & F (_rot_0_ & F (_pos_1_ & F _rot_1_))) & G !_pos_2_
        & $\psi_{15}$
        & \res{0.0}{0.0}{0.0} & \res{0.0}{0.0}{0.0} & \res{0.0}{0.0}{0.0} & \res{0.7}{0.3}{0.2} & \res{\mathbf{12.9}}{1.2}{0.7} \\

        % psi_16 = (!_has_0_ U (_pos_0_ | _pos_1_)) & GF (_has_0_ & _pos_2_) & GF (_has_0_ & _pos_3_) & G !(_has_0_ & (_pos_0_ | _pos_1_))
        & $\psi_{16}$
        & \res{0.0}{0.0}{0.0} & \res{0.0}{0.0}{0.0} & \res{0.0}{0.0}{0.0} & \res{0.0}{0.2}{0.0} & \res{\mathbf{39.4}}{3.8}{5.8} \\

        % psi_17 = GF (_has_0_ & _pos_0_) & GF (_has_1_ & _pos_1_)
        & $\psi_{17}$
        & \res{0.0}{0.0}{0.0} & \res{0.0}{0.0}{0.0} & \res{0.0}{0.0}{0.0} & \res{0.0}{0.0}{0.0} & \res{\mathbf{4.9}}{1.0}{0.7} \\

        % psi_18 = GF _has_0_ & GF !_has_0_ & G (_has_0_ => F (_pos_0_ & F _pos_1_))
        & $\psi_{18}$
        & \res{1.0}{0.0}{0.0} & \res{1.0}{0.0}{0.0} & \res{1.0}{0.0}{0.0} & \res{1.2}{0.6}{0.2} & \res{\mathbf{7.4}}{0.4}{0.5} \\

        \midrule

        % ================= FalloutWorld =================
        \multirow{6}{*}{\rotatebox[origin=c]{90}{FalloutWorld}}
        % psi_19 = G (_rad_0_ => F (_loc_0_ & F _loc_1_)) & GF _loc_2_ & GF _loc_3_
        & $\psi_{19}$
        & \res{27.7}{0.4}{0.4} & \res{6.2}{1.0}{1.1} & \res{1.0}{0.0}{0.0} & \res{\mathbf{30.8}}{0.2}{0.2} & \res{29.9}{0.3}{0.2} \\

        % psi_20 = GF _loc_0_ & GF _loc_1_ & GF _loc_2_ & G !_rad_0_
        & $\psi_{20}$
        & \res{7.0}{0.4}{0.3} & \res{0.4}{0.1}{0.1} & \res{0.0}{0.0}{0.0} & \res{\mathbf{15.4}}{0.5}{0.5} & \res{\mathbf{15.4}}{0.4}{0.4} \\

        % psi_21 = GF (_loc_0_ & F (_loc_1_ & F (_loc_2_ & F _loc_3_))) & G !_rad_0_
        & $\psi_{21}$
        & \res{3.8}{0.2}{0.3} & \res{0.1}{0.0}{0.0} & \res{0.0}{0.0}{0.0} & \res{10.0}{0.4}{0.4} & \res{\mathbf{10.2}}{0.3}{0.3} \\

        % psi_22 = GF _loc_0_ & GF _loc_1_ & GF _loc_2_ & GF _loc_3_ & GF _loc_4_ & GF _loc_5_
        & $\psi_{22}$
        & \res{18.1}{0.3}{0.3} & \res{0.8}{0.3}{0.2} & \res{0.0}{0.0}{0.0} & \res{\mathbf{19.6}}{0.1}{0.1} & \res{19.2}{0.1}{0.1} \\

        % psi_23 = GF (_rad_0_ & F (_loc_0_ & F (!_rad_0_ & F _loc_1_)))
        & $\psi_{23}$
        & \res{52.1}{1.0}{0.8} & \res{18.0}{2.5}{2.0} & \res{1.3}{0.6}{0.4} & \res{\mathbf{58.9}}{0.5}{0.4} & \res{57.4}{0.7}{0.7} \\

        % psi_24 = GF _rad_0_ & GF _loc_0_ & GF _loc_1_ & G (_rad_0_ => (!(_loc_0_ | _loc_1_) U _loc_2_))
        & $\psi_{24}$
        & \res{4.0}{0.4}{0.4} & \res{1.0}{0.0}{0.0} & \res{1.0}{0.0}{0.0} & \res{4.9}{0.7}{0.5} & \res{\mathbf{5.3}}{0.5}{0.4} \\

        \bottomrule
    \end{tabular}
    \end{tiny}
    \end{center}
    \vskip -0.1in
\end{table*}

%%%%%%%%%%%%%%%%%%%%%%%%%%%%%%%%
% TRAJECTORY VISUALIZATIONS
%%%%%%%%%%%%%%%%%%%%%%%%%%%%%%%%

\section{Trajectory Visualizations}
\label{app:trajectory_visualizations}

\begin{figure*}[t]
    \centering
    \captionsetup[subfloat]{justification=centering}
    \newcommand\alltrajswidth{0.994}
    \newcommand\pertrajwidth{0.244}
    \subfloat[\textit{Go1RGBZoneEnv}; $\psi_{2} = \F \G \mathsf{at}_0 \land \F (\mathsf{at}_1 \land \F \mathsf{at}_2)$.\label{fig:trajectories:go1}]{%
    \makebox[\alltrajswidth\linewidth]{%
        \includegraphics[width=\pertrajwidth\linewidth]{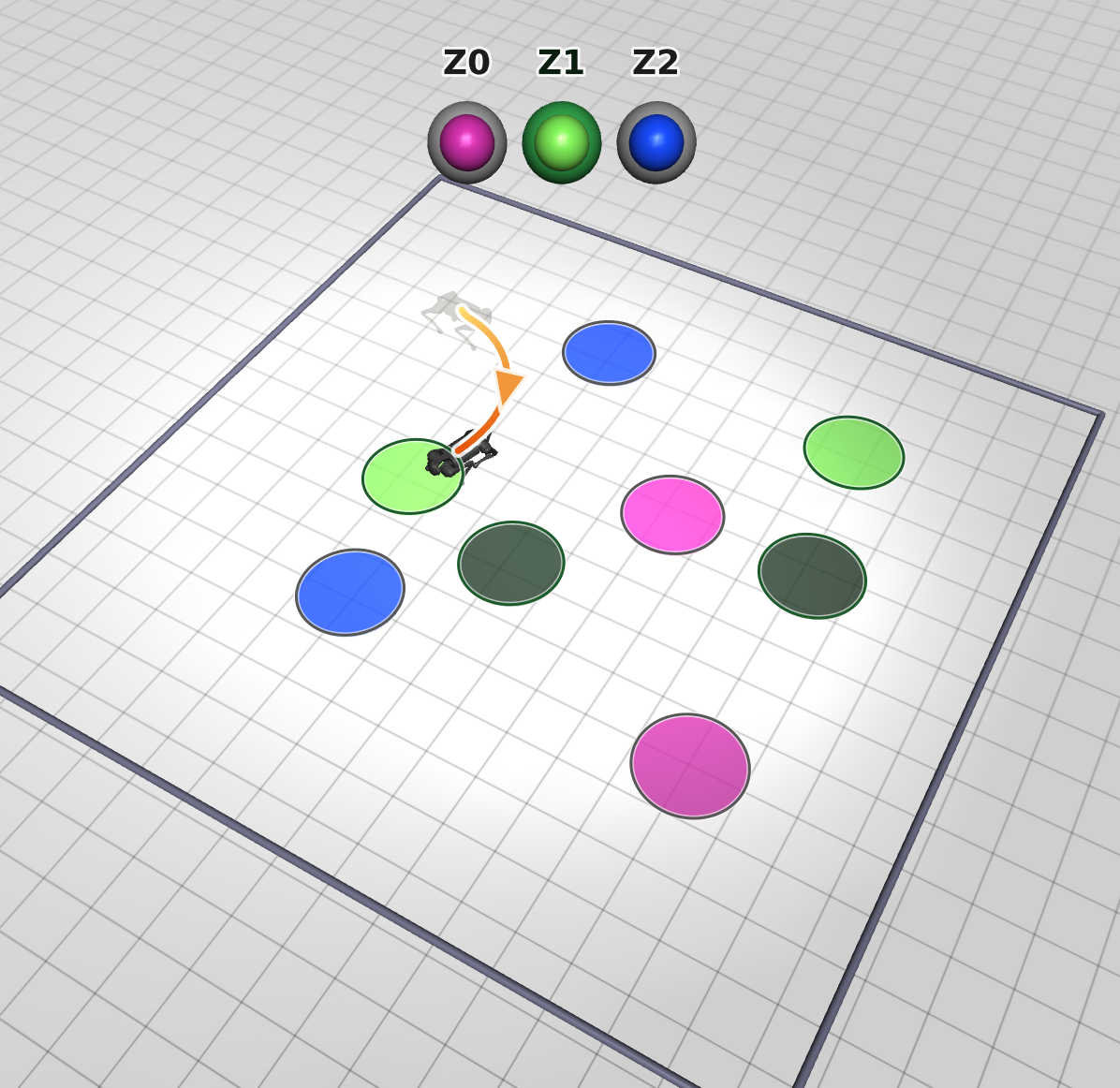}%
        \hfill
        \includegraphics[width=\pertrajwidth\linewidth]{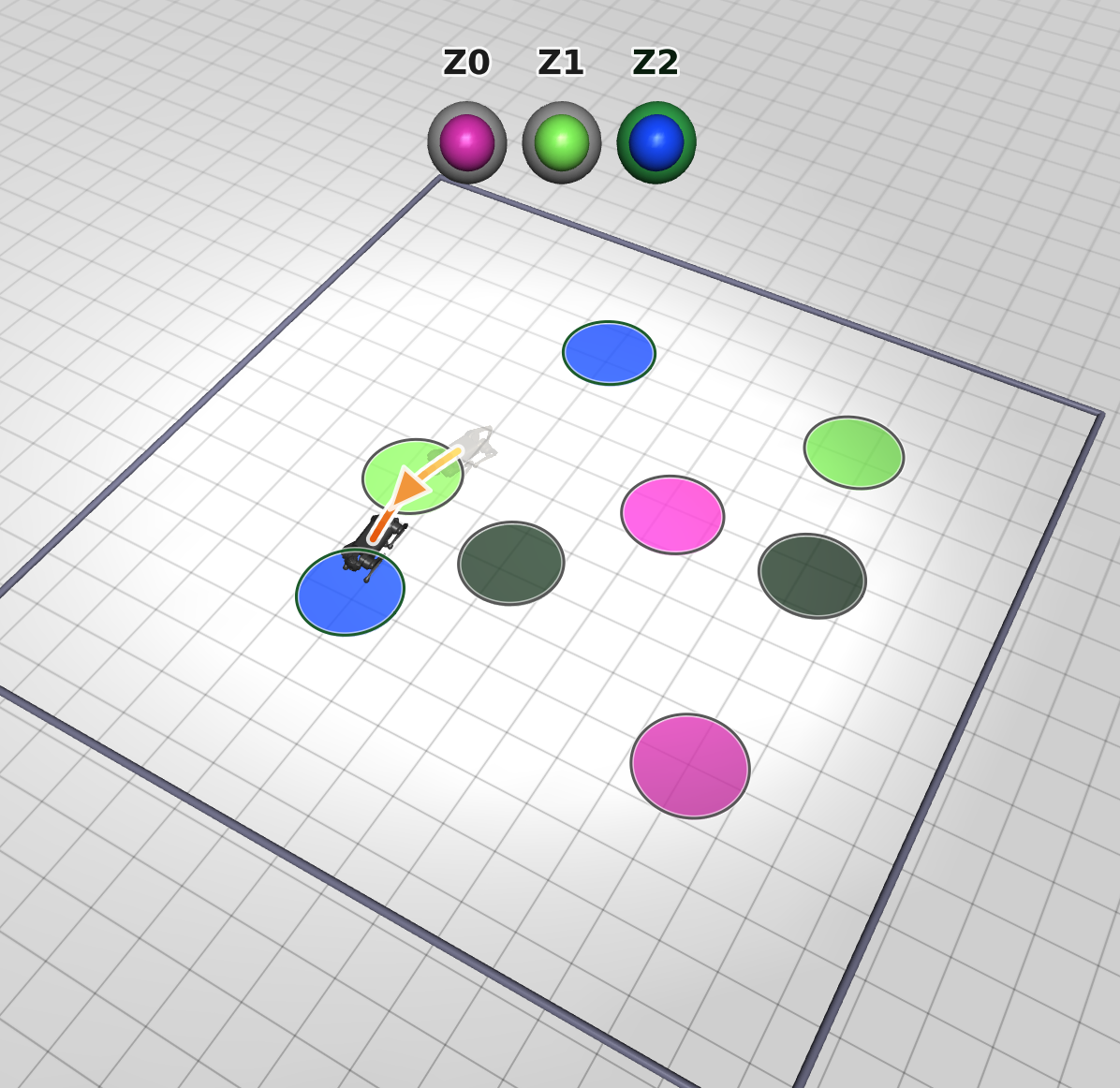}%
        \hfill
        \includegraphics[width=\pertrajwidth\linewidth]{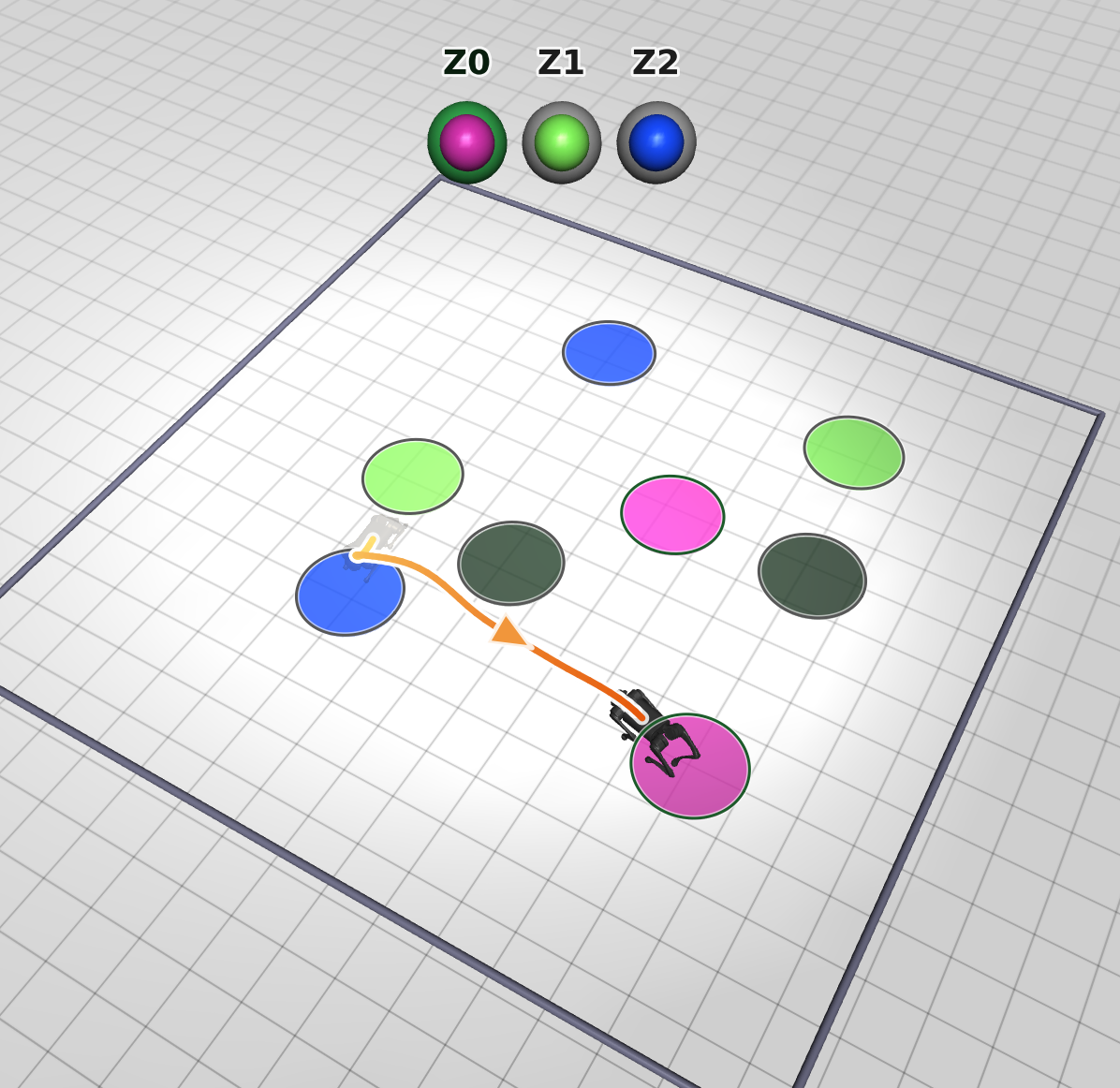}%
        \hfill
        \includegraphics[width=\pertrajwidth\linewidth]{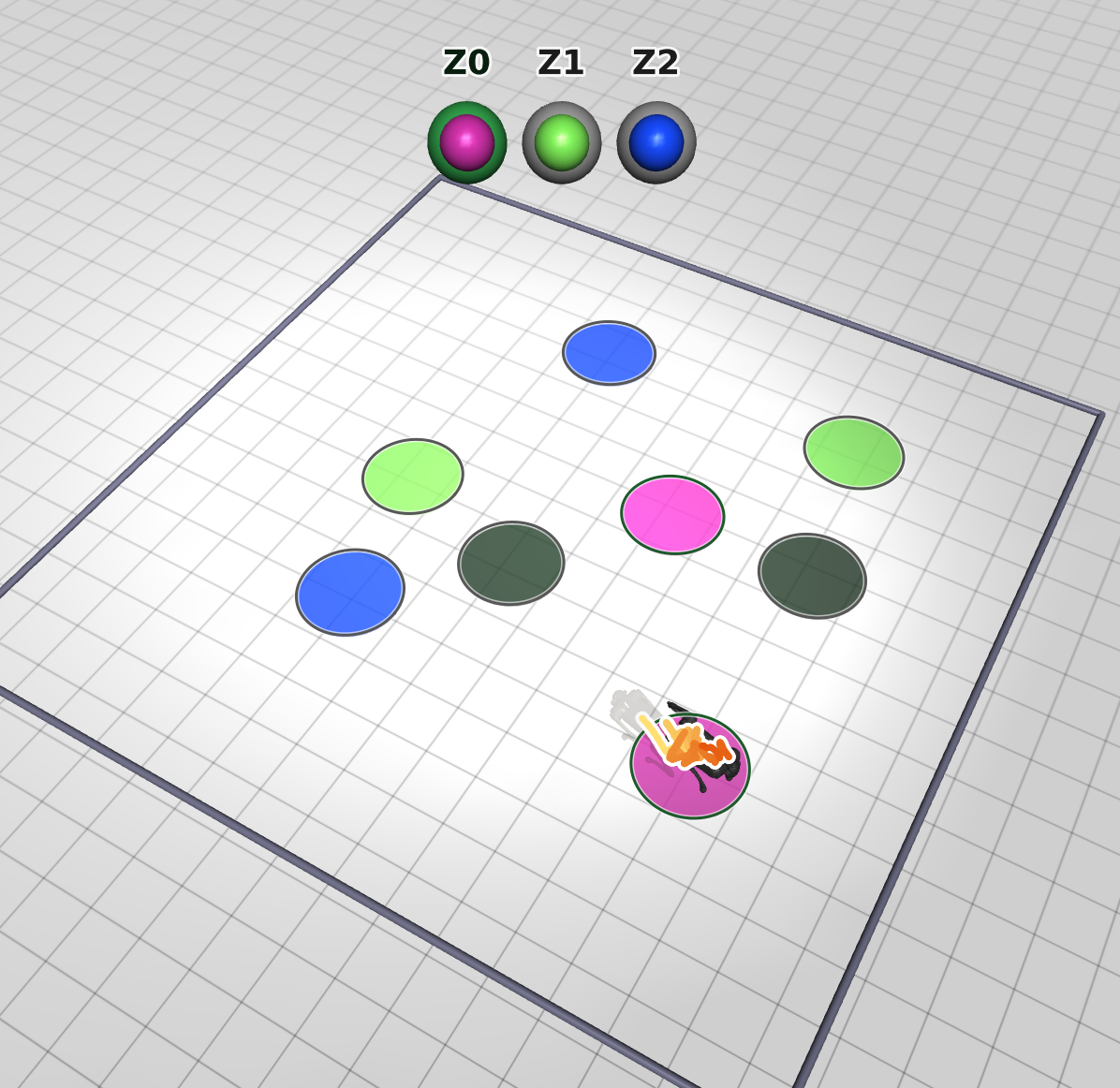}%
        }
    }
    \par\bigskip
    \subfloat[\textit{X2DronesEnv}; \\ \rule{0pt}{3ex} $\varphi_{12} = \F (\mathsf{pos}_0 \land \F ((\mathsf{pos}_1 \land \mathsf{pos}_2) \land \F ((\mathsf{pos}_2 \land \mathsf{pos}_3) \land (\lnot (\mathsf{pos}_0 \lor \mathsf{pos}_1) \U (\mathsf{pos}_3 \land \mathsf{pos}_4)))))$.\label{fig:trajectories:x2}]{%
    \makebox[\alltrajswidth\linewidth]{%
        \includegraphics[width=\pertrajwidth\linewidth]{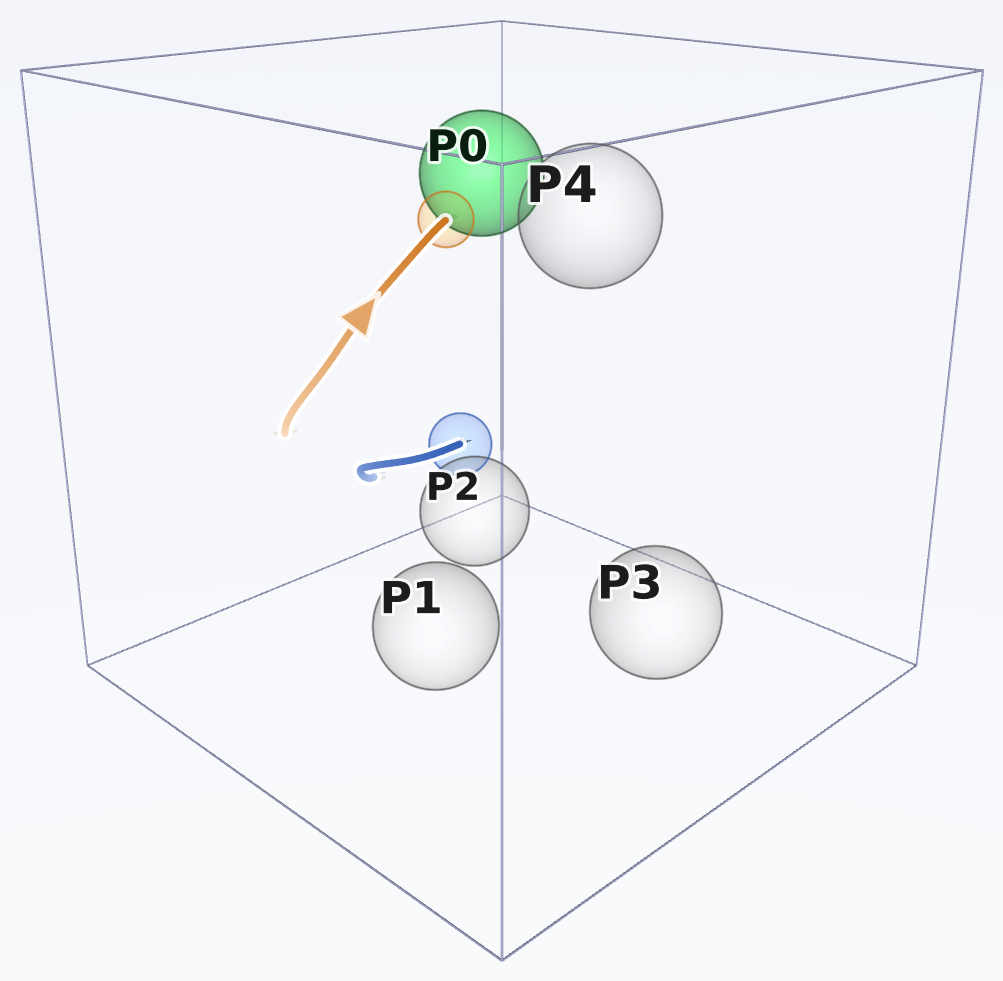}%
        \hfill
        \includegraphics[width=\pertrajwidth\linewidth]{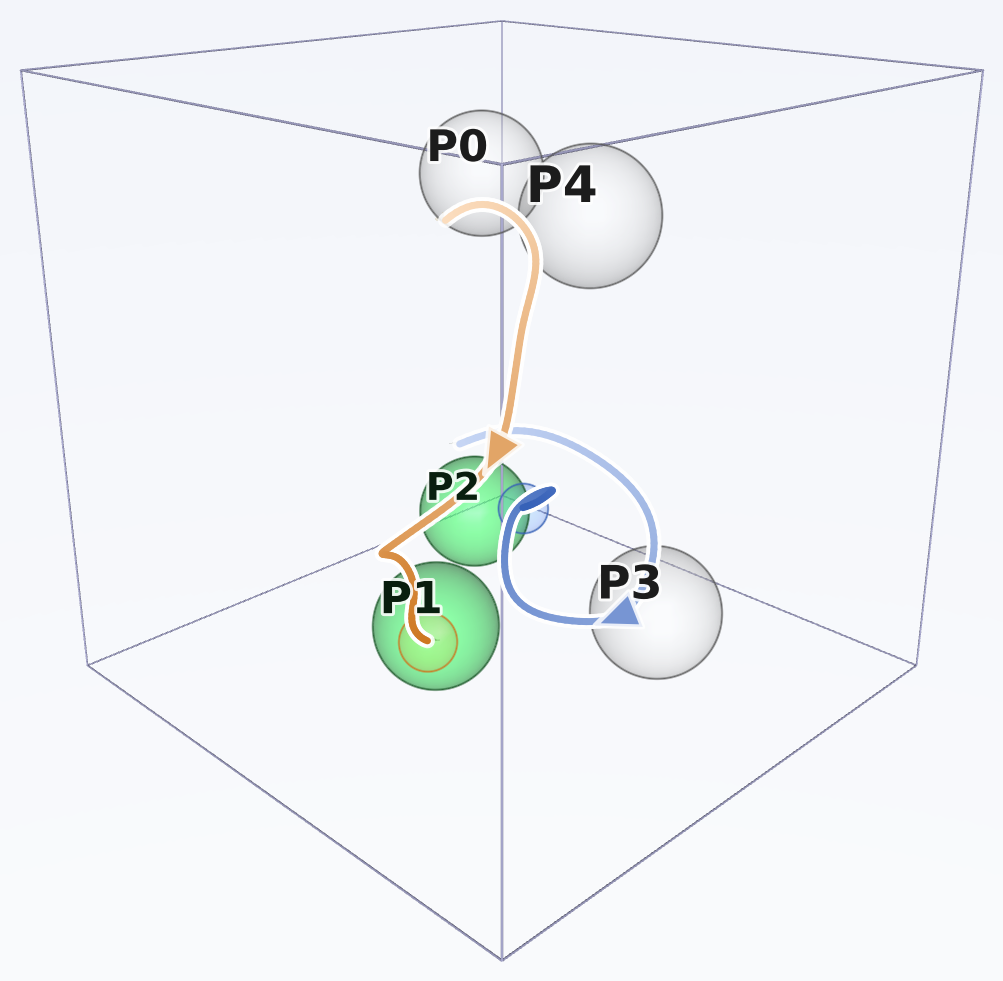}%
        \hfill
        \includegraphics[width=\pertrajwidth\linewidth]{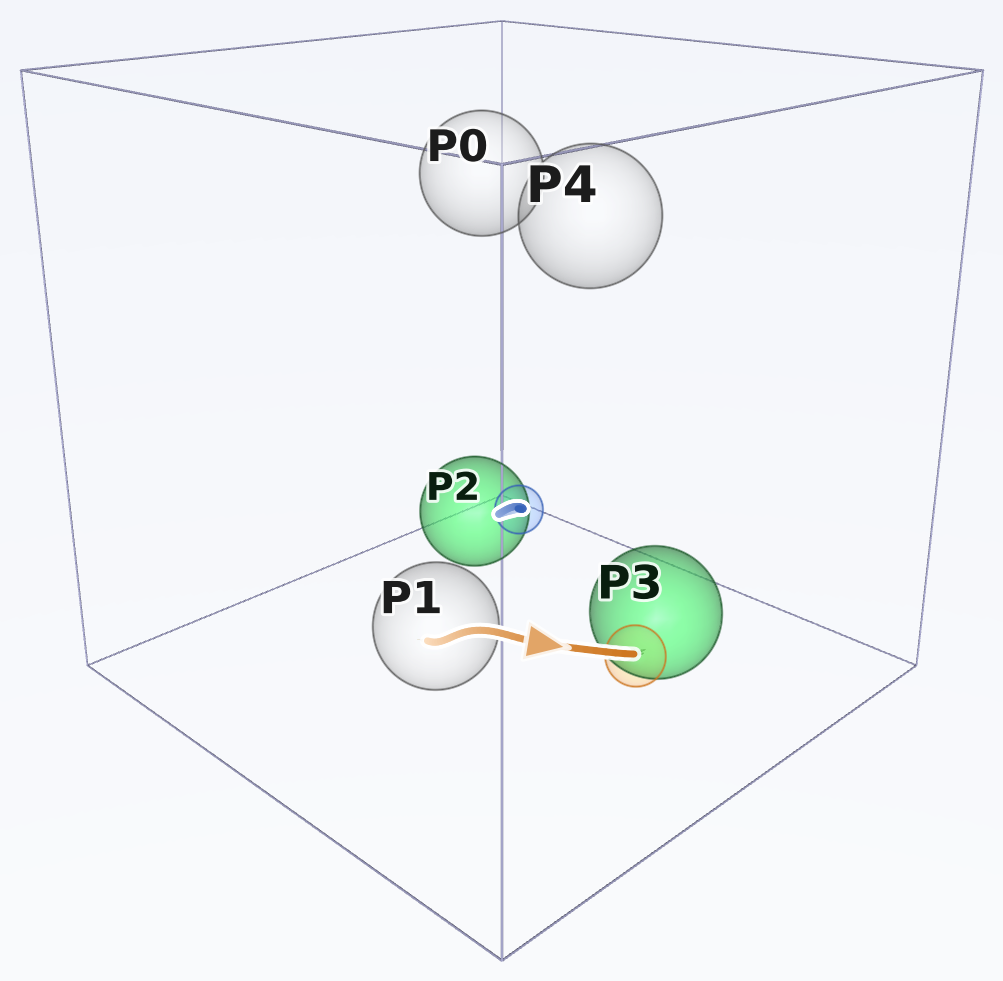}%
        \hfill
        \includegraphics[width=\pertrajwidth\linewidth]{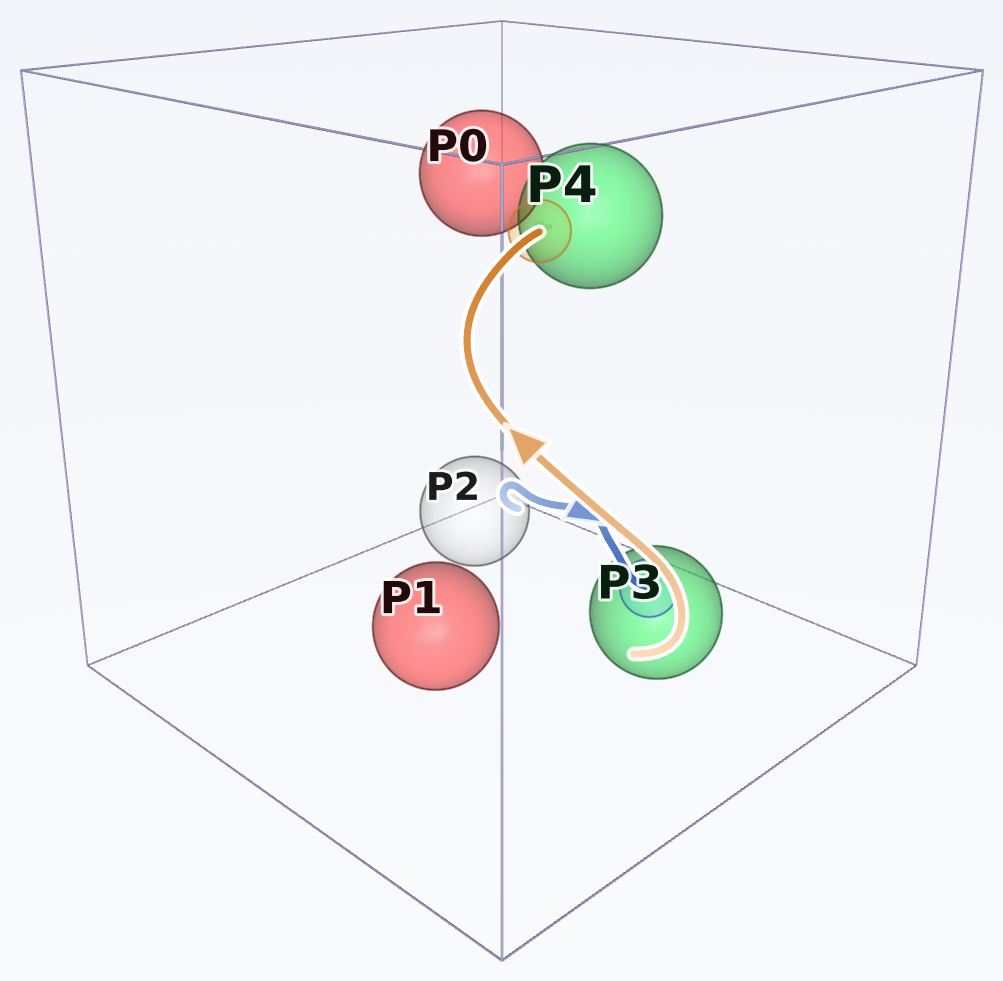}%
        }
    }
    \par\bigskip
    \subfloat[\textit{FrankaCubesEnv}; \\ \rule{0pt}{3ex} $\varphi_{22} = \F ((\mathsf{has}_0 \land \mathsf{pos}_0) \land (\mathsf{pos}_0 \U (\lnot \mathsf{has}_0 \land \mathsf{pos}_0))) \land \F ((\mathsf{has}_1 \land \mathsf{pos}_1) \land (\mathsf{pos}_1 \U (\lnot \mathsf{has}_1 \land \mathsf{pos}_1)))$.\label{fig:trajectories:franka}]{%
    \makebox[\alltrajswidth\linewidth]{%
        \includegraphics[width=\pertrajwidth\linewidth]{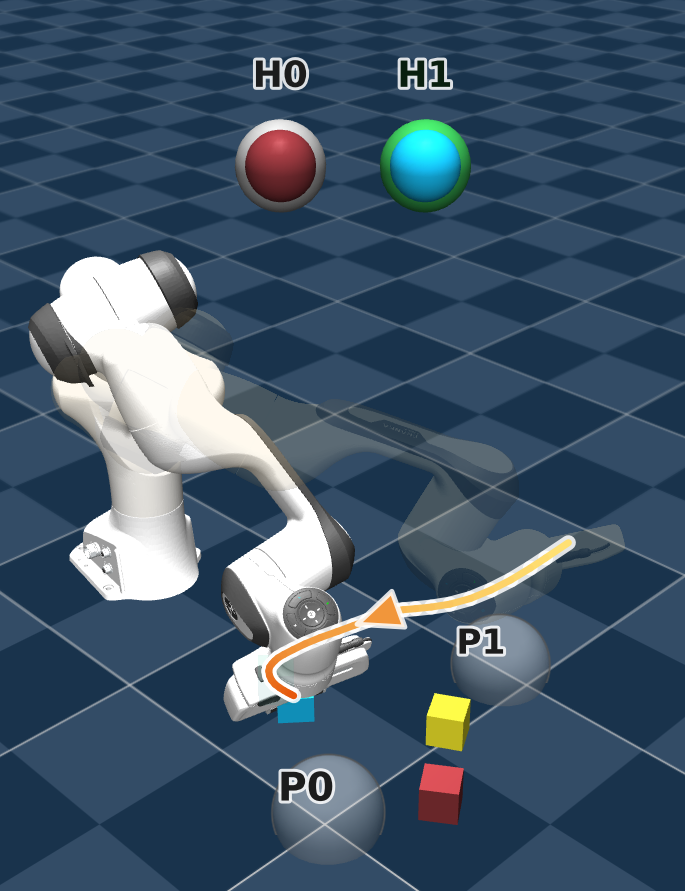}%
        \hfill
        \includegraphics[width=\pertrajwidth\linewidth]{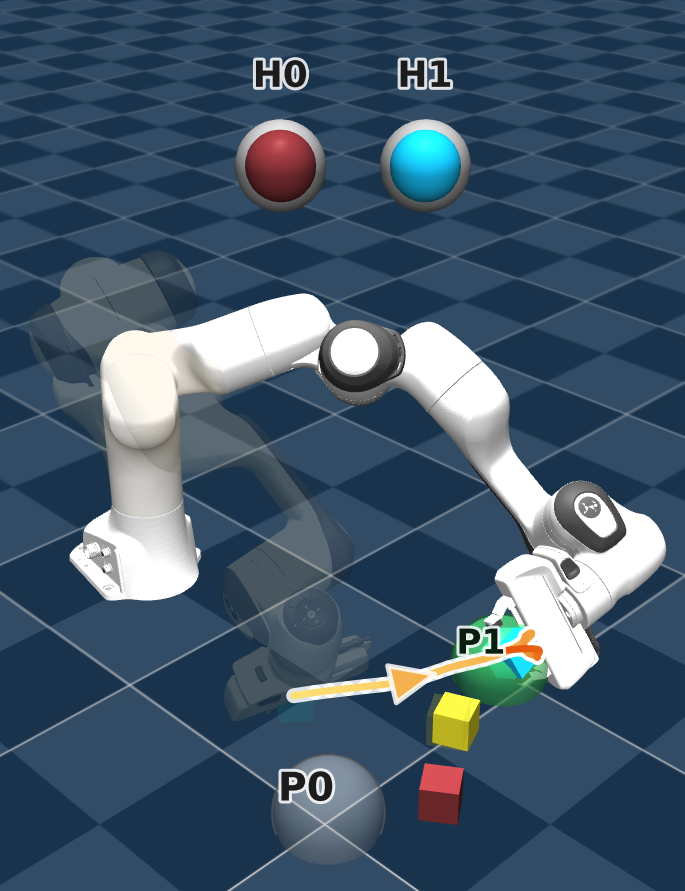}%
        \hfill
        \includegraphics[width=\pertrajwidth\linewidth]{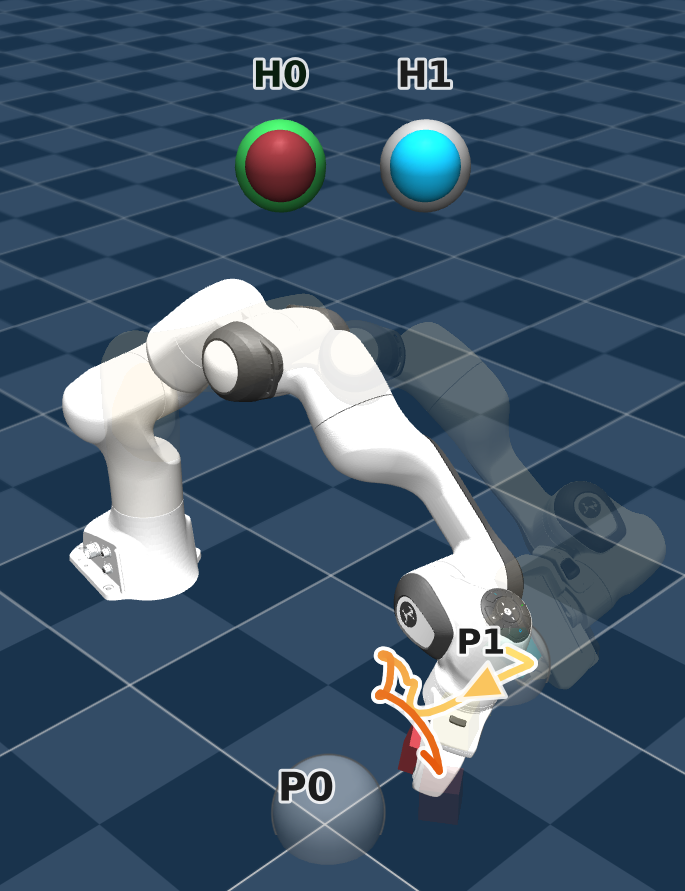}%
        \hfill
        \includegraphics[width=\pertrajwidth\linewidth]{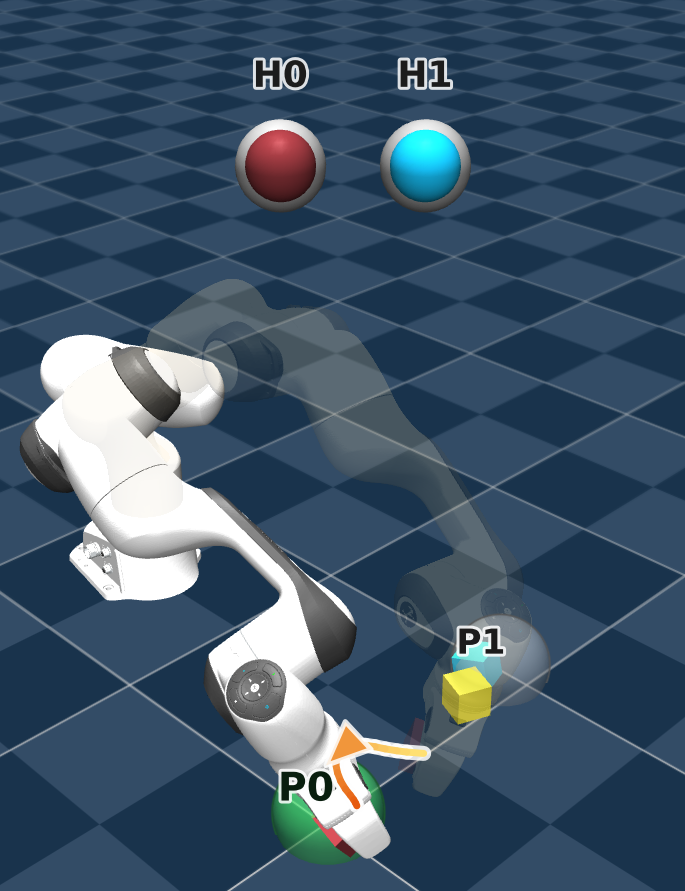}%
        }
    }
    \par\bigskip
    \subfloat[\textit{FalloutWorld}; $\psi_{20} = \G \F \mathsf{loc}_0 \land \G \F \mathsf{loc}_1 \land \G \F \mathsf{loc}_2 \land \G \lnot \mathsf{rad}_0$.\label{fig:trajectories:fw}]{%
    \makebox[\alltrajswidth\linewidth]{%
        \includegraphics[width=\pertrajwidth\linewidth]{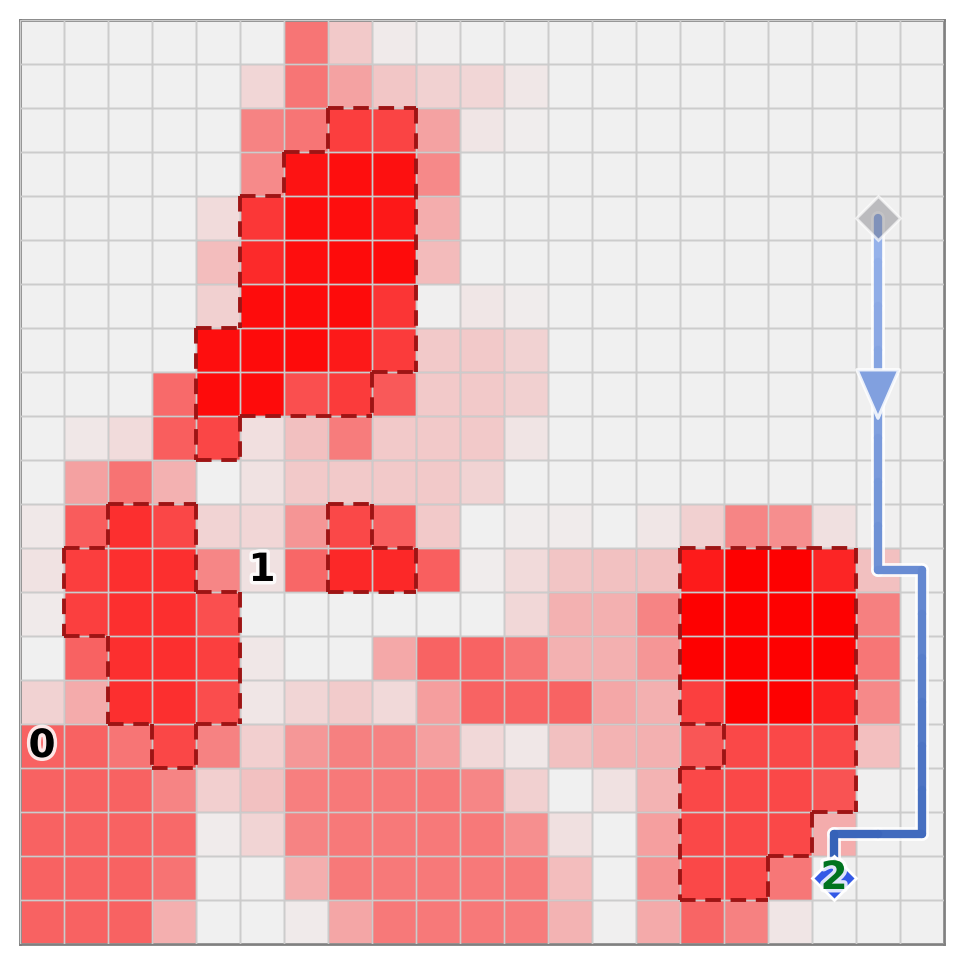}%
        \hfill
        \includegraphics[width=\pertrajwidth\linewidth]{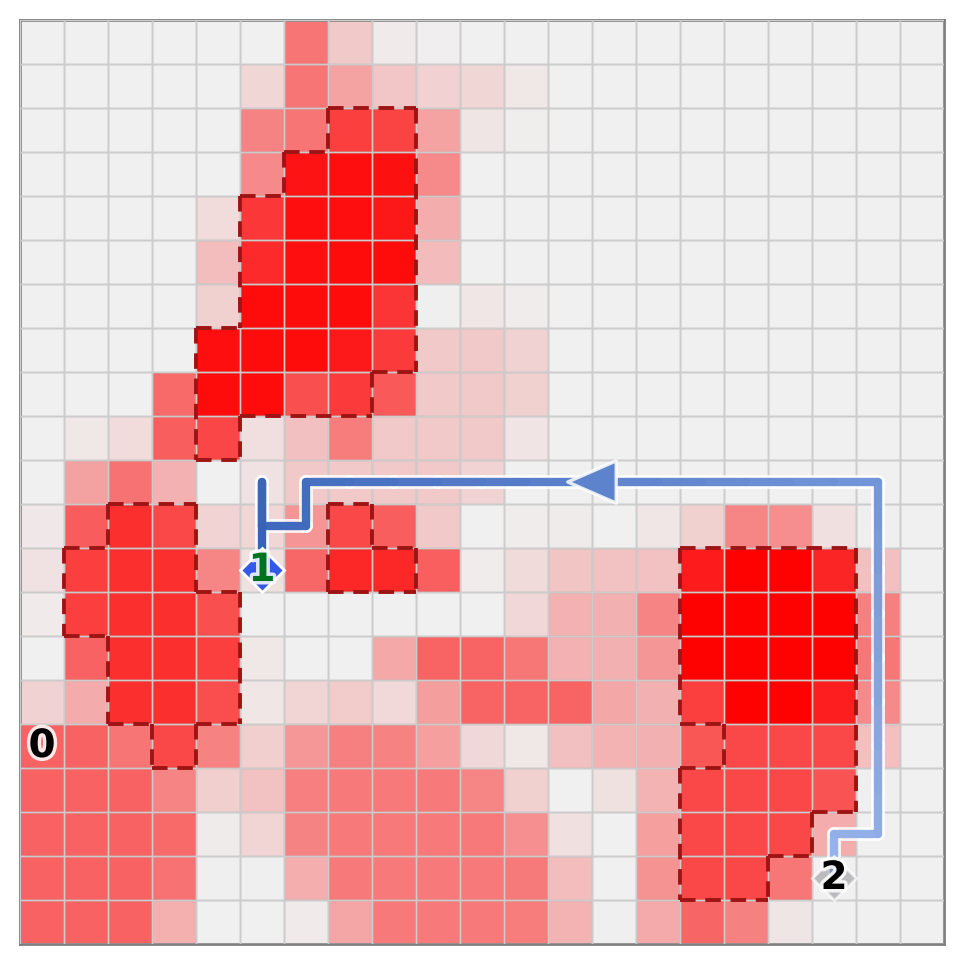}%
        \hfill
        \includegraphics[width=\pertrajwidth\linewidth]{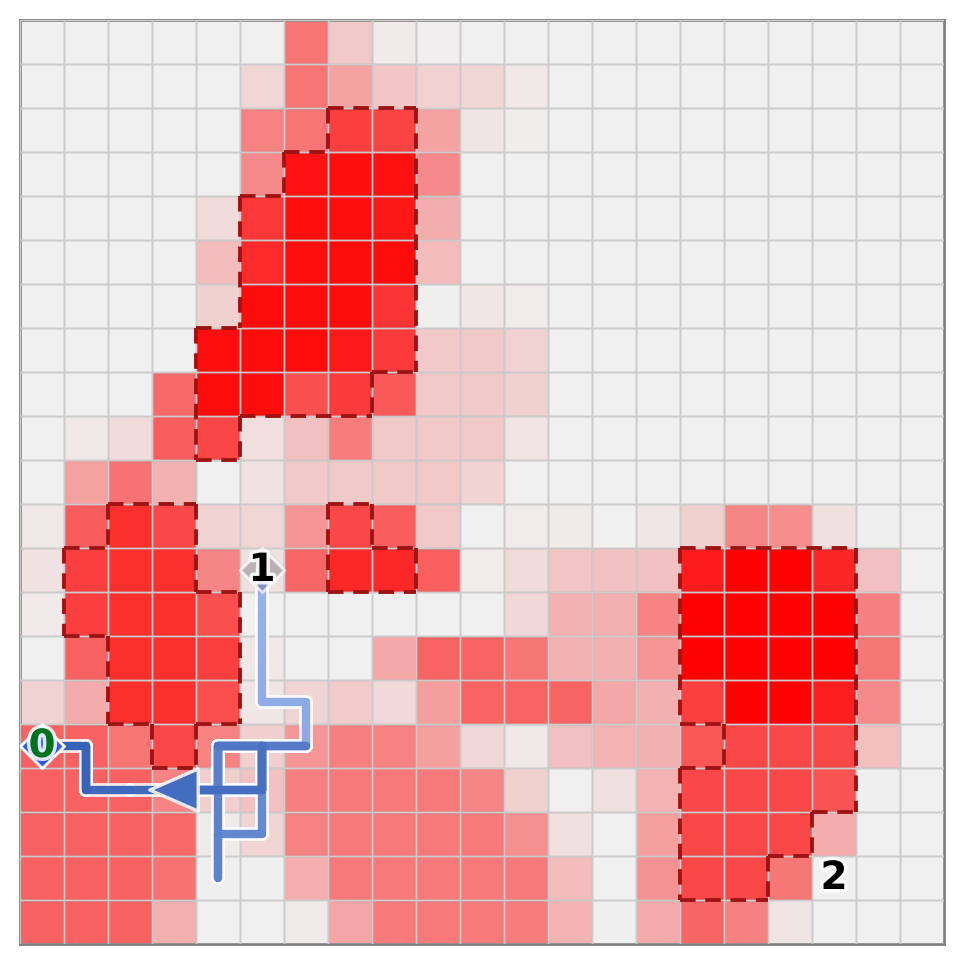}%
        \hfill
        \includegraphics[width=\pertrajwidth\linewidth]{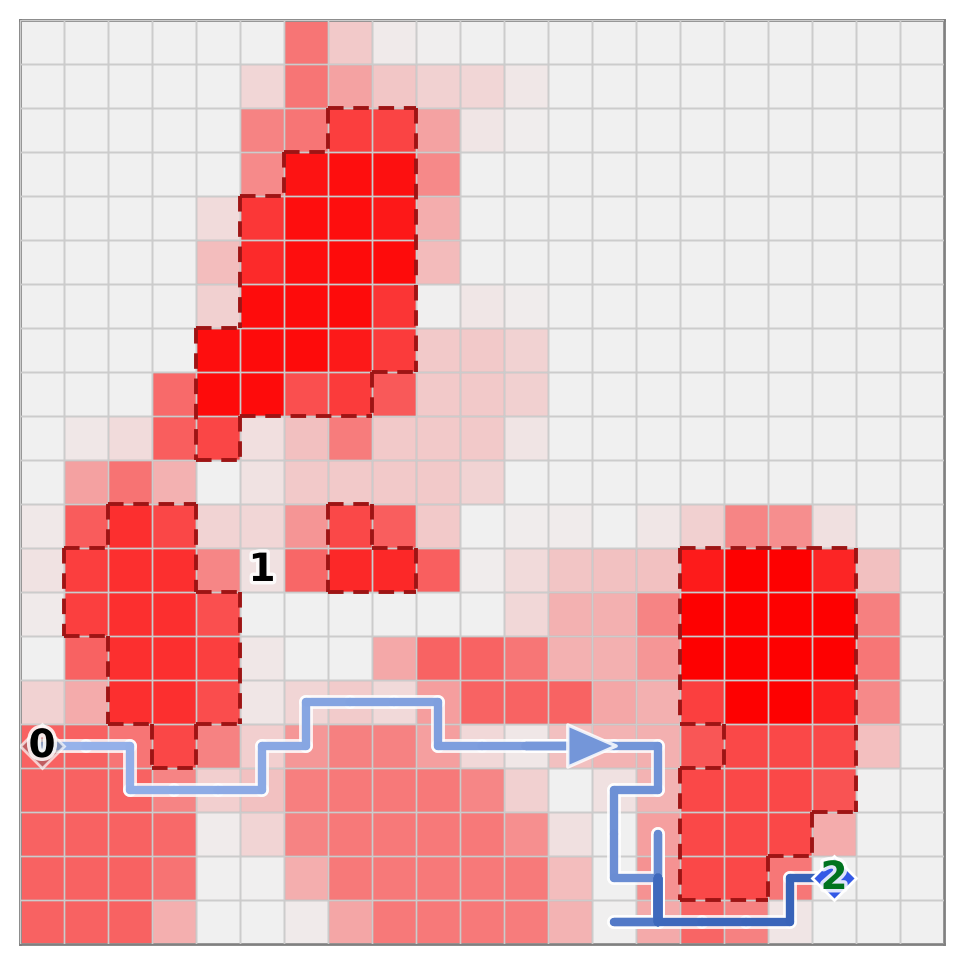}%
        }
    }
    \caption{Example trajectories for Plato with P\textsuperscript{2}BRS, truncated to the first 4 subgoals (left to right).}
    \label{fig:trajectories}
\end{figure*}

Figure~\ref{fig:trajectories} presents example trajectories for Plato with P\textsuperscript{2}BRS evaluated on \textit{finite} and \textit{infinite} tasks, truncated to the first 4 subgoals.

%%%%%%%%%%%%%%%%%%%%%%%%%%%%%%%%
% ABLATIONS
%%%%%%%%%%%%%%%%%%%%%%%%%%%%%%%%

\section{Ablation Studies}
\label{app:ablation_studies}

\subsection{Principal Component Analysis of Proposition Embeddings}
\label{app:ablation_studies:pca_prop_embeddings}

This ablation study tests our claim that the Plato embedding module learns low-dimensional per-predicate geometries in embedding space. We consider the two environments with multiple predicates: \textit{FrankaCubesEnv} (with $\mathsf{pos}(x,y,z)$, $\mathsf{rot}(\phi,\theta,\zeta)$, and $\mathsf{has}(r,g,b)$) and \textit{FalloutWorld} (with $\mathsf{rad}(\eta)$ and $\mathsf{loc}(x,y)$). We compare policies learned using Plato and Discretize (the other baselines do not learn unique embeddings for each predicate instance); we also include Plato with P\textsuperscript{2}BRS in \textit{FrankaCubesEnv}. For each environment, we sample $1024$ predicate instances for each predicate and perform a series of principal component analyses (PCAs) on the corresponding learned proposition embeddings under each method. We perform a combined PCA over all predicates as well as per-predicate PCAs, measuring the explained variance ratios $\lambda_i \in [0,1]$ for each component $i$ ($\Sigma_i \lambda_i = 1$). For each predicate we also compute the \textit{participation ratio} $\mathrm{PR} = \sfrac{(\Sigma_i \lambda_i)^2}{\Sigma_i \lambda^2_i}$, which measures the effective dimensionality of its embedding distribution.

\begin{figure*}[t]
    \centering
    \subfloat[Discretize]{\includegraphics[width=0.32\linewidth]{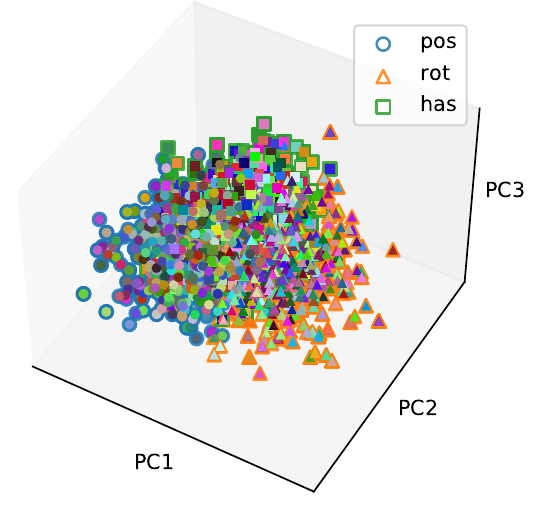} \label{fig:pca_franka:discretize_all}}
    \hfil
    \subfloat[Plato]{\includegraphics[width=0.32\linewidth]{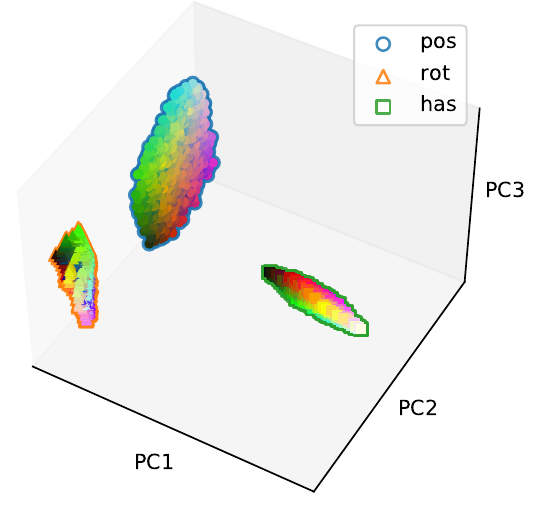} \label{fig:pca_franka:plato_all}}
    \hfil
    \subfloat[Plato + P\textsuperscript{2}BRS]{\includegraphics[width=0.32\linewidth]{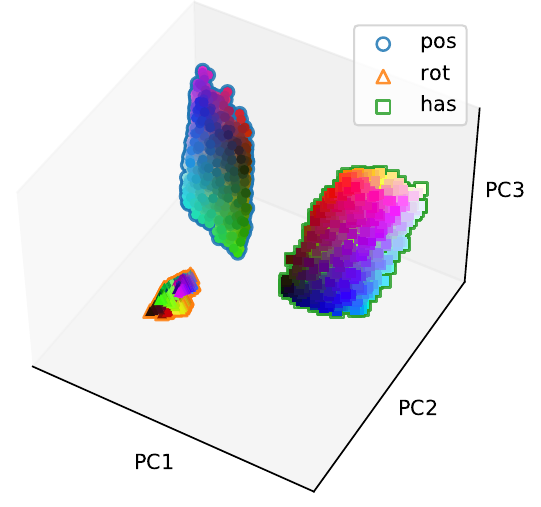} \label{fig:pca_franka:plato_shaped_all}}
    \par\bigskip
    \subfloat[Per-predicate PCAs (Discretize), including explained variance ratios.]{\includegraphics[width=0.99\linewidth]{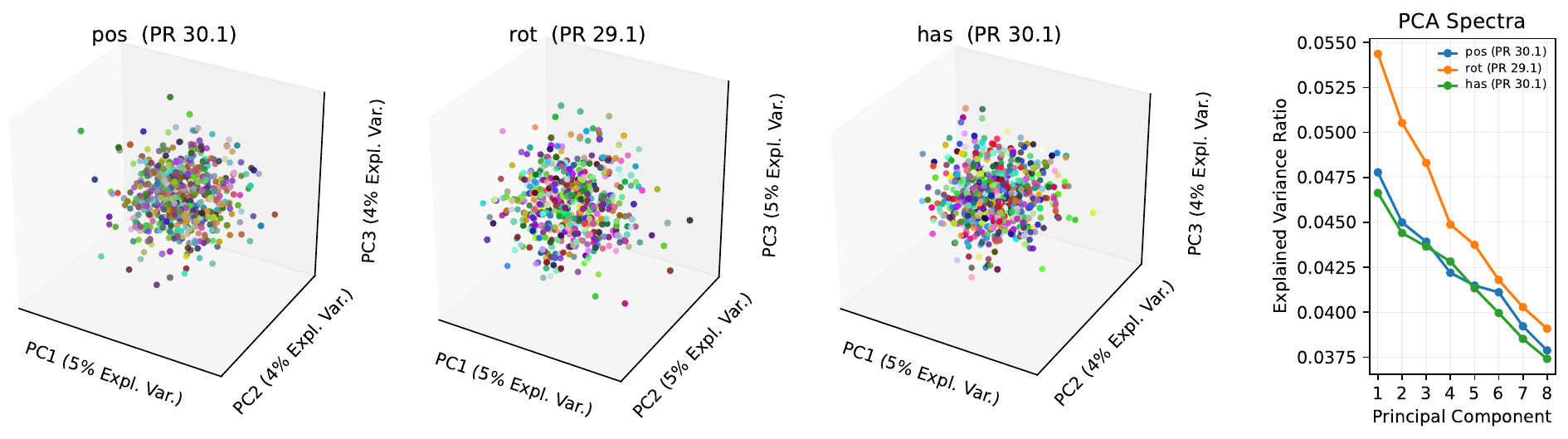} \label{fig:pca_franka:discretize}}
    \par\bigskip
    \subfloat[Per-predicate PCAs (Plato), including explained variance ratios.]{\includegraphics[width=0.99\linewidth]{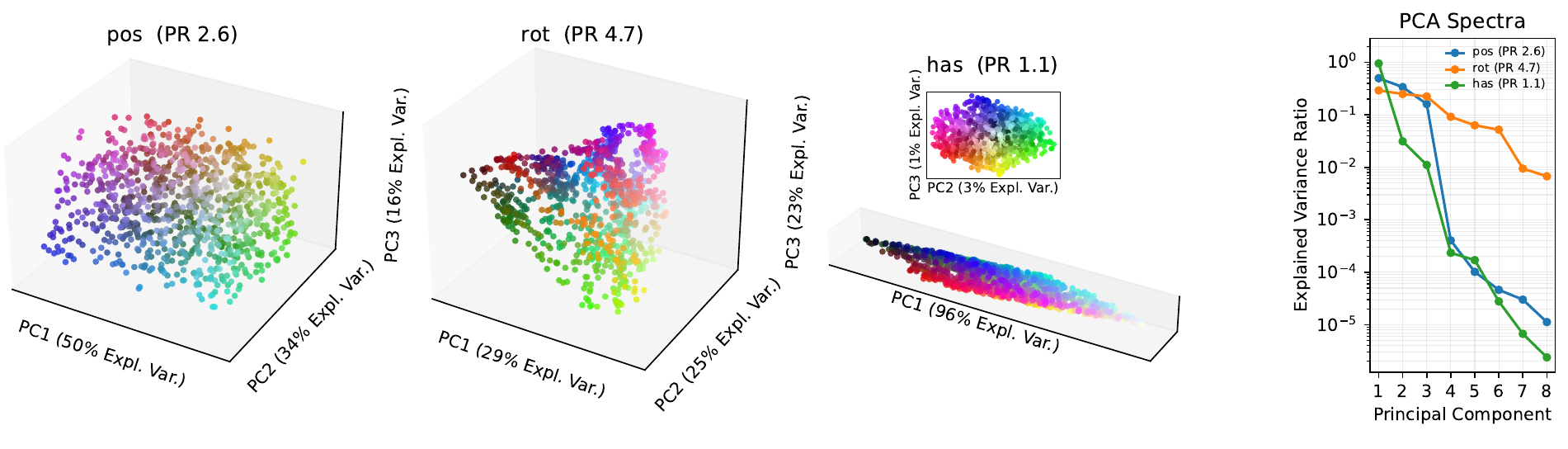} \label{fig:pca_franka:plato}}
    \par\bigskip
    \subfloat[Per-predicate PCAs (Plato + P\textsuperscript{2}BRS), including explained variance ratios.]{\includegraphics[width=0.99\linewidth]{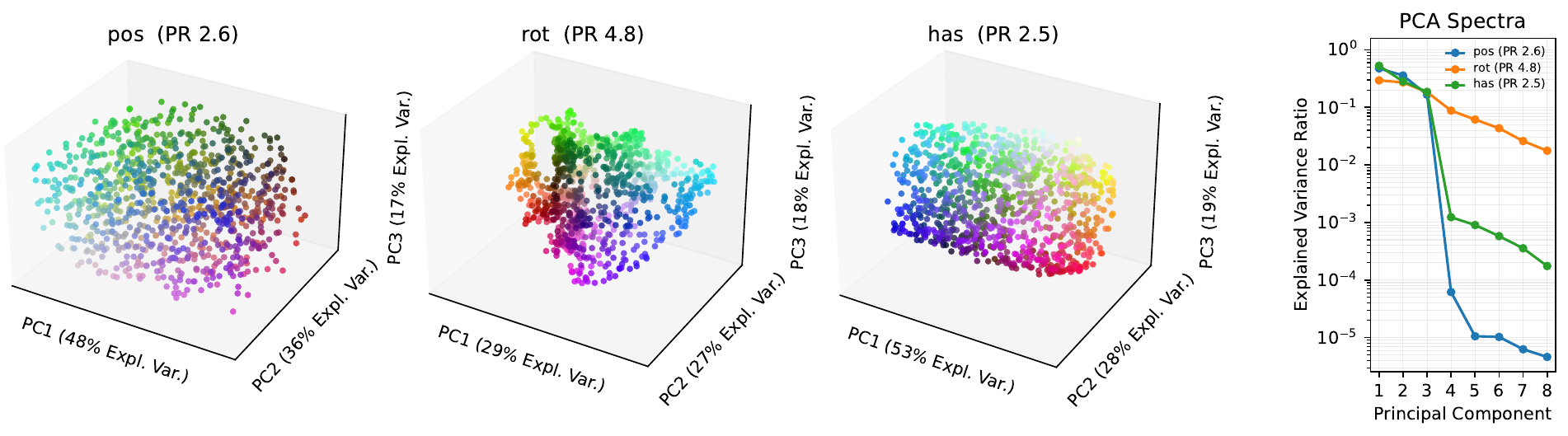} \label{fig:pca_franka:plato_shaped}}
    \caption{Principal component analyses of learned proposition embeddings in \textit{FrankaCubesEnv}. Normalized parameters for a predicate instance $\mathsf{pos}(x, y, z)$, $\mathsf{rot}(\phi,\theta,\zeta)$, or $\mathsf{has}(r, g, b)$ are mapped to the RGB value $(r, g, b) \in [0, 1]^3$.}
    \label{fig:pca_franka}
    \vskip -0.1in
\end{figure*}

Figure~\ref{fig:pca_franka} presents the results for \textit{FrankaCubesEnv}. We see that Plato (both with and without P\textsuperscript{2}BRS) learns smooth, low-dimensional geometries for each predicate, cleanly separated in embedding space. For Plato with P\textsuperscript{2}BRS, the distributions for $\mathsf{pos}$ and $\mathsf{has}$ have a $\mathrm{PR}$ of $2.6$ and $2.5$ respectively with a steep drop in explained variance ratio after the third component, corresponding to roughly 3D shapes (as we would expect given the three roughly independent input parameters for each). The distribution for $\mathsf{rot}$ yields a moderately higher $\mathrm{PR}$ of $4.8$, with its PCA spectrum suggesting a flatter attenuation in explained variance ratio per component. This may be due to a more complex relationship between angles for reachable orientations (since we only sample reachable orientations). Plato without P\textsuperscript{2}BRS shows very similar results except that the distribution for $\mathsf{has}$ is compressed to a $\mathrm{PR}$ of $1.1$, with the dominant component aligned with luminance. This is possibly an artifact of unguided Plato failing to learn to satisfy tasks involving $\mathsf{has}$ predicate instances, which require the manipulator arm to pick up a correctly colored cube (the most difficult aspect of the environment). Since unguided Plato still yields low-dimensional per-predicate geometries despite failing to converge to a strong policy, we note that Plato's inductive bias is not wholly sufficient to solve complex robotics environments such as \textit{FrankaCubesEnv}, with P\textsuperscript{2}BRS needed to close the gap.

Meanwhile, we see significant overlap between the distributions for each predicate under Discretize. Furthermore, the per-predicate $\mathrm{PR}$s are all in the range $29.1$--$30.1$; given the embedding dimension $n_\phi = 32$, these distributions effectively span the entire high-dimensional embedding space. They are also highly non-smooth, with no discernible spatial correlations with respect to parameter values.

\begin{figure*}[t]
    \centering
    \subfloat[Discretize]{\includegraphics[width=0.49\linewidth]{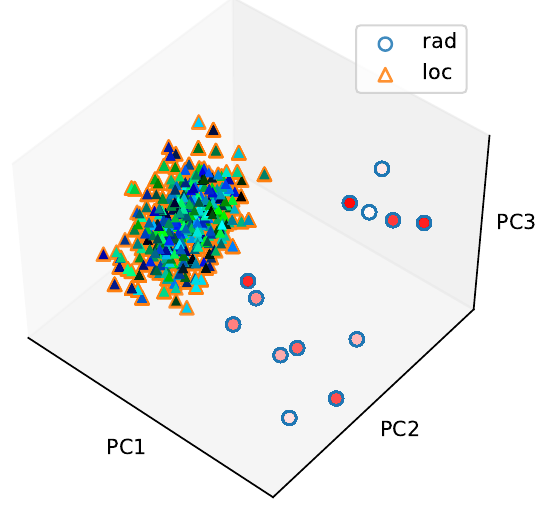} \label{fig:pca_fw:discretize_all}}
    \hfil
    \subfloat[Plato]{\includegraphics[width=0.49\linewidth]{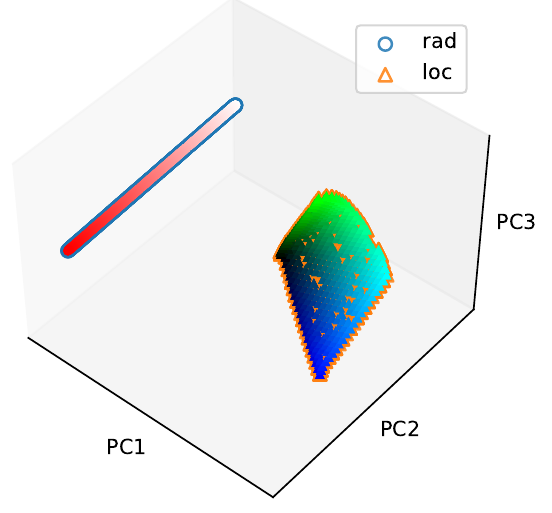} \label{fig:pca_fw:plato_all}}
    \par\bigskip
    \subfloat[Per-predicate PCAs (Discretize), including explained variance ratios.]{\includegraphics[width=0.99\linewidth]{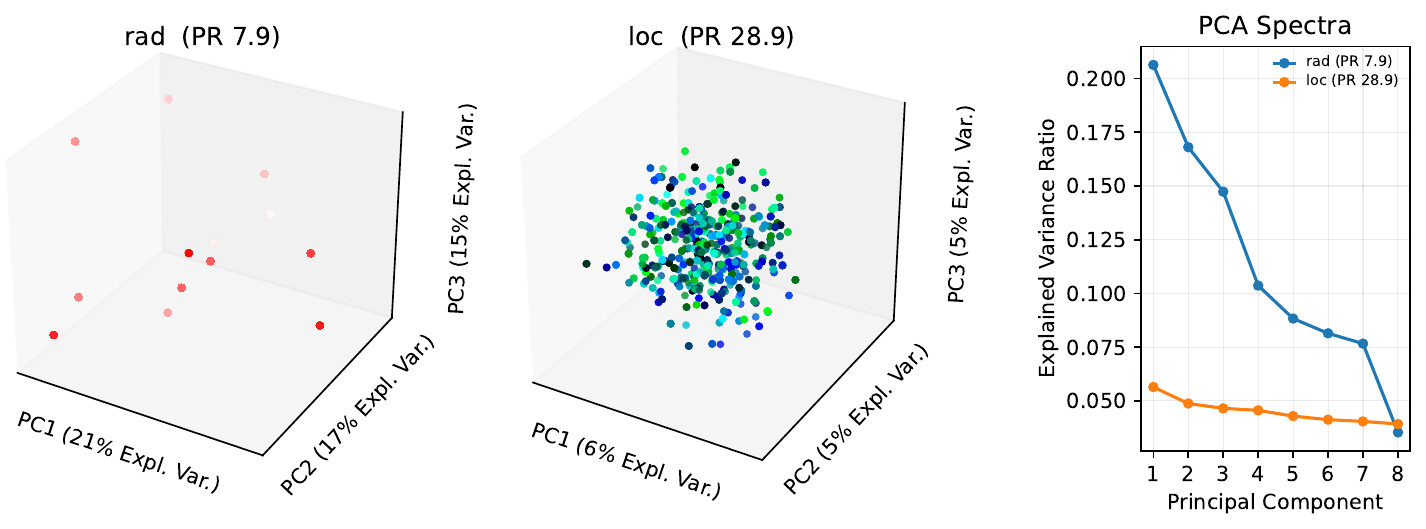} \label{fig:pca_fw:discretize}}
    \par\bigskip
    \subfloat[Per-predicate PCAs (Plato), including explained variance ratios.]{\includegraphics[width=0.99\linewidth]{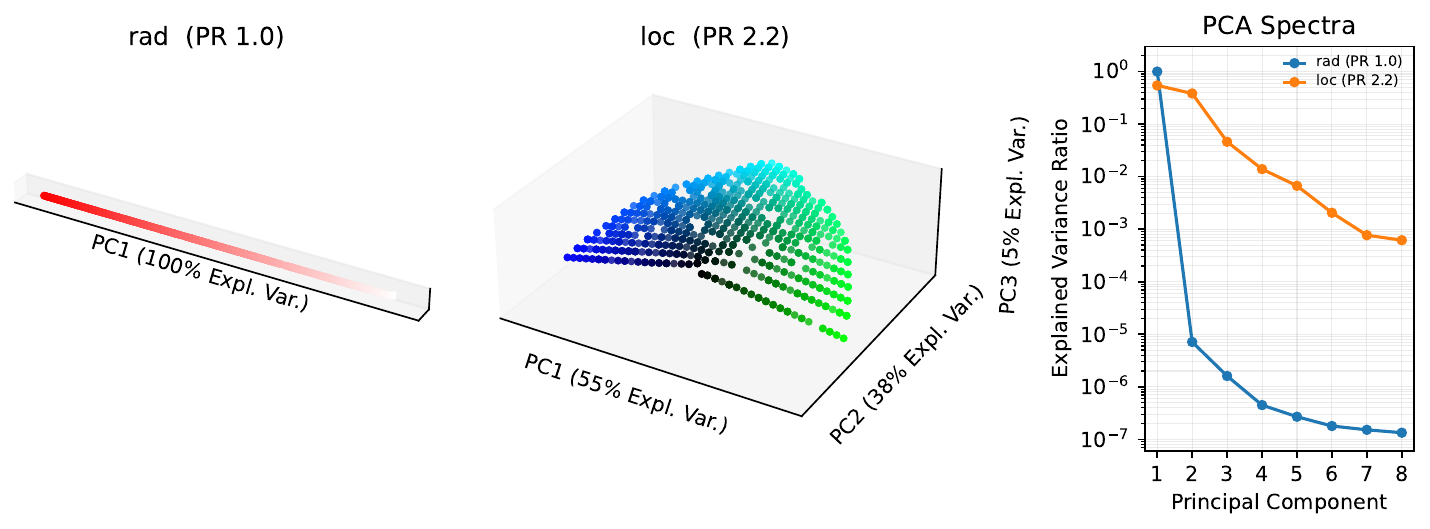} \label{fig:pca_fw:plato}}
    \caption{Principal component analyses of learned proposition embeddings in \textit{FalloutWorld}. The normalized parameter for a predicate instance $\mathsf{rad}(\eta)$ is mapped to the RGB value $(1, 1 - \eta, 1 - \eta) \in [0, 1]^3$ such that color transitions from white to red as radiation intensity threshold increases, and normalized parameters for an instance $\mathsf{loc}(x, y)$ are mapped to the RGB value $(0, x, y) \in [0, 1]^3$. Note that $\mathsf{rad}(\eta)$ only yields $13$ unique data points for Discretize, since it only has $13$ bins under our discretization.}
    \label{fig:pca_fw}
    \vskip -0.1in
\end{figure*}

Figure~\ref{fig:pca_fw} presents the results for \textit{FalloutWorld}, for which we see the same trend, with Plato learning cleanly separated low-dimensional shapes for each predicate while Discretize learns non-smooth distributions spanning the embedding space. Plato yields a $\mathrm{PR}$ of $1.0$ and $2.2$ for $\mathsf{rad}$ and $\mathsf{loc}$ respectively, as is to be expected given their parameter input dimensions. In contrast, Discretize yields a $\mathrm{PR}$ of $28.9$ for $\mathsf{loc}$; the relatively low $\mathrm{PR}$ of $7.9$ for $\mathsf{rad}$ is deceptive and is simply due to there only being $13$ discrete bins for this predicate. Note that Discretize performs relatively well in \textit{FalloutWorld} (at least compared to Concatenate and SIAMS), since the $\mathsf{loc}$ bins correspond exactly to the possible grid locations. Thus, even a ``successful'' policy under Discretize essentially memorizes a dictionary of independent embeddings rather than organizing them into coherent geometries for efficient learning. However, we do at least see some separation between predicates under Discretize for this environment.

\subsection{PlatoLTL with LTL2Action}
\label{app:ablation_studies:ltl2action}

This study demonstrates that both the Plato embedding module and P\textsuperscript{2}BRS can be applied to backbones beyond StructLTL. We use LTL2Action \citep{vaezipoor_ltl2action_2021}, which encodes the syntax tree of the LTL specification using a GNN and updates the specification via LTL progression as the agent progresses through the task. We chose this backbone since it is widely influential within the field and considerably different from sequence-conditioned policies such as DeepLTL \citep{jackermeier_deepltl_2025}, GenZ-LTL \citep{guo_one_2025}, and StructLTL \citep{jackermeier_zero-shot_2026}.

LTL2Action uses a learned encoder $\phi$ to map proposition tokens into embeddings, thus satisfying Assumption \ref{asm:embedding} cleanly and enabling the use of the Plato embedding module. However, to apply P\textsuperscript{2}BRS we require that the LTL2Action policy operates over the product MDP space $\mathcal{S} \times \mathcal{Q}^\varphi$ such that we can obtain $\Phi(s, q^\varphi) = \Phi_\beta\big(s, \text{DFS}_{\mathcal{B}^\varphi}(q^\varphi)\big)$; indeed, we define a policy over product MDP states, $\pi(a \cond s, q^\varphi)$, as part of our problem setting (Section~\ref{sec:problem_setting}). While LTL2Action uses LTL progression rather than constructing an explicit LDBA, the two representations track the same underlying object: each progressed formula corresponds, up to propositional equivalence, to a state of the automaton obtained by the ``after function'' construction of \citet{chaudhuri_limit-deterministic_2016}, from which the LDBA is derived. Conversely, for the co-safe fragment of LTL admitted by LTL2Action, every reachable automaton state is represented by a progressed formula: co-safe automata require no accepting-component bookkeeping, as their accepting states are absorbing and represented by the formula $\true$. We therefore run an LDBA monitor alongside the executing policy and define $\mathit{form}(q^\varphi)$ \textit{operationally} as the formula obtained by progressing $\varphi$ along the trace that reached $q^\varphi$; by the correspondence above, this formula and the monitored LDBA state evolve in lockstep, so $\mathit{form}$ is well-defined along every execution without requiring a canonical representative per abstract state. This allows us to write the policy as $\pi(a \cond s, q^\varphi) = \pi_\mathit{form}\big(a \cond s, \mathit{form}(q^\varphi)\big)$, where $\pi_\mathit{form}$ is LTL2Action's formula-conditioned policy, enabling us to compute $\Phi$ using the monitored automaton state $q^\varphi$.

\begin{figure*}[t]
    \centering
    \subfloat[\textit{Go1RGBZoneEnv}]{\includegraphics[width=0.99\linewidth]{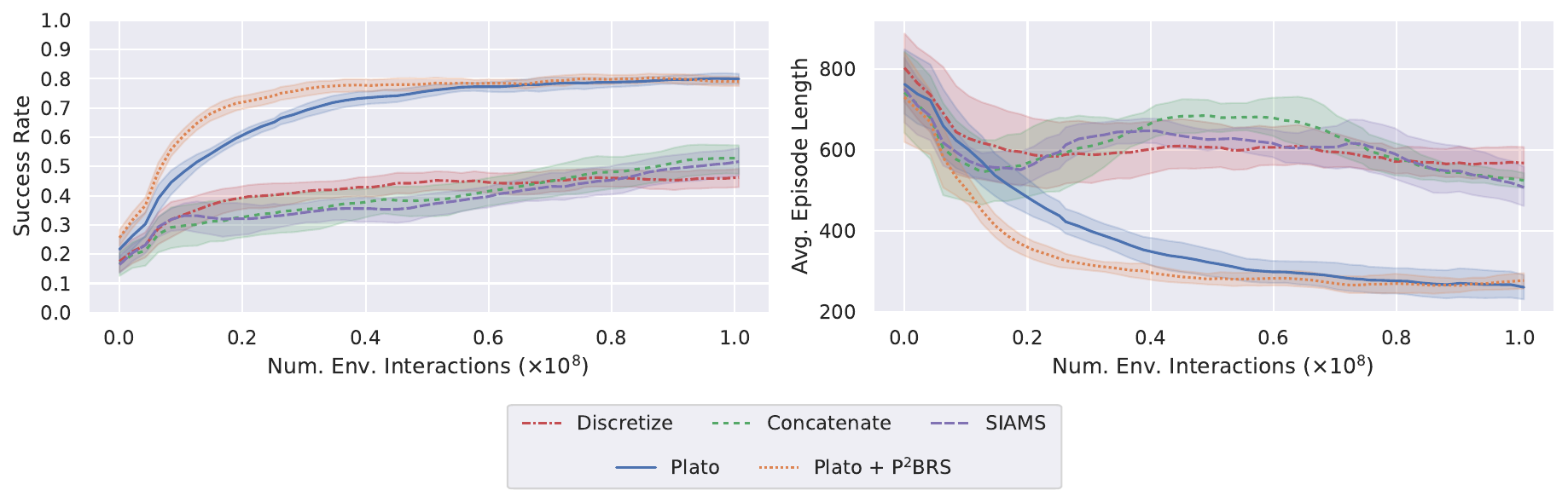} \label{fig:ltl2action:go1}}
    \par\bigskip
    \subfloat[\textit{X2DronesEnv}]{\includegraphics[width=0.99\linewidth]{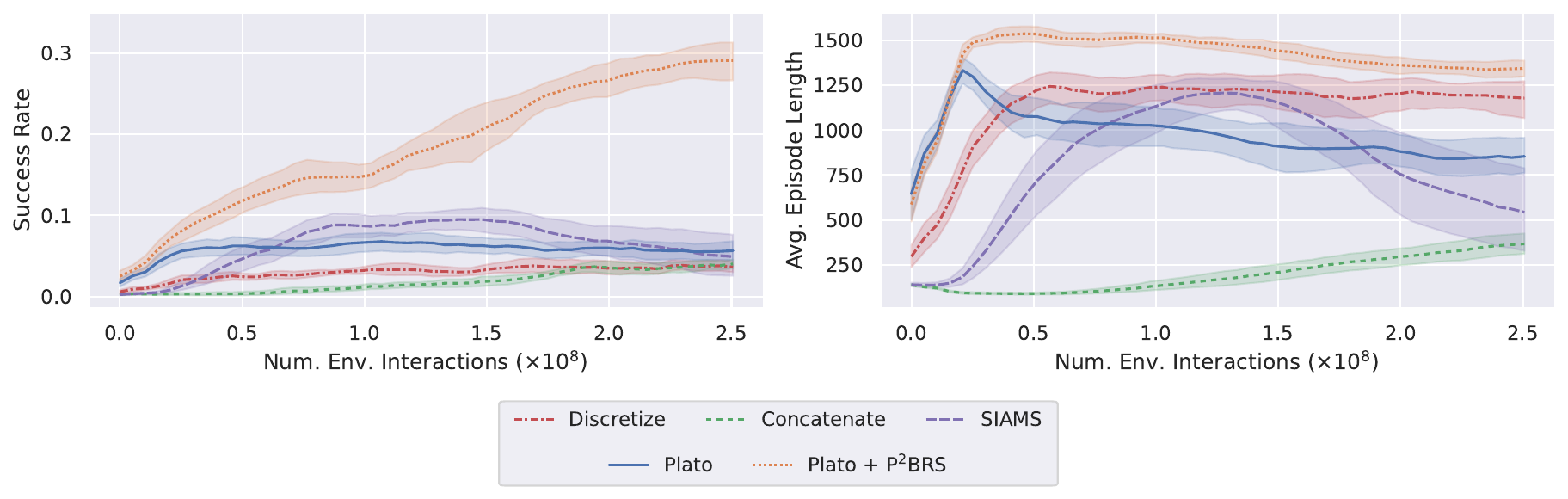} \label{fig:ltl2action:x2}}
    \caption{Evaluation curves for \textit{finite}-horizon specifications using LTL2Action.}
    \label{fig:ltl2action}
    \vskip -0.1in
\end{figure*}

Figure~\ref{fig:ltl2action:go1} presents evaluation curves during training for \textit{finite} tasks in \textit{Go1RGBZoneEnv} using LTL2Action. Much like we saw for StructLTL, Plato demonstrates strong performance, converging to a high success rate and low episode length while the baselines remain far from convergence. We see that converged success rate and time to converge are both notably worse than those when using StructLTL; we attribute this to the backbone itself, since StructLTL has been demonstrated to significantly outperform LTL2Action in modern benchmarks \citep{jackermeier_zero-shot_2026}. Interestingly, applying P\textsuperscript{2}BRS yields markedly faster convergence compared to unguided Plato when using LTL2Action, which was not the case for StructLTL in \textit{Go1RGBZoneEnv}. This suggests that LTL2Action suffers an even greater exploration bottleneck than StructLTL, which P\textsuperscript{2}BRS alleviates.

Figure~\ref{fig:ltl2action:x2} presents evaluation curves for \textit{finite} tasks in \textit{X2DronesEnv}, over an extended training budget of \num{2.5e8} environment interactions. Unguided Plato's success rate flatlines alongside the baselines, a divergence in behavior from that observed with StructLTL, where Plato demonstrated steady learning progress. Again, we attribute this to the capability of the LTL2Action backbone. However, we do see clear training progress upon application of P\textsuperscript{2}BRS, with success rate rising slowly but steadily.

We omit results in \textit{FrankaCubesEnv} since LTL2Action did not yield signs of learning progress even upon application of P\textsuperscript{2}BRS. Thus, it is essential to use a strong backbone in combination with PlatoLTL (e.g., StructLTL) to solve the most challenging environments and tasks; neither is independently sufficient.

\subsection{P\textsuperscript{2}BRS Ablations}
\label{app:ablation_studies:p2brs}

This study ablates key design choices in P\textsuperscript{2}BRS, justifying the need for persistence gating and adaptive robustness scale calibration, as well as investigating the effect of replacing P\textsuperscript{2}BRS's hard $\min$/$\max$ aggregators with smooth combinators. We also consider automaton-based PBRS as a baseline, demonstrating its insufficiency for multi-task RL in high-dimensional environments with complex subgoals. Figure~\ref{fig:shaping_ablations} presents evaluation curves during training for \textit{finite} tasks in \textit{X2DronesEnv} and \textit{FrankaCubesEnv} under each ablation, as well as Plato with and without P\textsuperscript{2}BRS for reference.

We do not include Ground-Compose-Reinforce (GCR; \citealp{li_ground-compose-reinforce_2025}) as an end-to-end baseline, since it trains single-task policies and learns its value functions offline from labeled trajectories, which are unavailable for our environments. However, P\textsuperscript{2}BRS without persistence gating (Appendix~\ref{app:ablation_studies:p2brs:persistence_gating}) composes robustness metrics with the same $\min$/$\max$ operators that GCR applies at the formula level, and thus isolates the effect of this composition scheme alone in our setting.

\subsubsection{Comparison to Automaton-Based PBRS}
\label{app:ablation_studies:p2brs:automaton-based_pbrs}

We employ the approach in Section~4.4 of \cite{icarte_reward_2022} as a representative example for automaton-based PBRS. The idea is to derive the potential function $\Phi$ from the LDBA $\mathcal{B}^\varphi$ alone, by treating $\mathcal{B}^\varphi$ itself as a standalone MDP $\mathcal{M}_{\mathcal{B}^\varphi}$: its states are the automaton states $q^\varphi \in \mathcal{Q}^\varphi$, its actions are the automaton's assignments and $\varepsilon$-transitions $\sigma^\varphi \in \Sigma^\varphi \cup \mathcal{E}^\varphi$, transitions follow the transition function $\delta^\varphi$, the reward is the task reward emitted on each transition (here, $+1$ upon visiting an accepting state $q^\varphi \in \mathcal{F}^\varphi$ and $0$ otherwise), and the discount factor is the $\gamma$ of the environment MDP $\mathcal{M}$. Value iteration on this abstract MDP yields optimal values $V_{\mathcal{B}^\varphi}^*(q^\varphi)$, corresponding to the return achieved by an idealized agent able to trigger any desired automaton transition at every environment step. The potential function is then defined as $\Phi(s, q^\varphi) = -V_{\mathcal{B}^\varphi}^*(q^\varphi)$, and is independent of environment state $s$. The potential function is overridden as $\Phi(s, q^\varphi) = 0$ at terminal states to ensure PBRS invariance; \cite{icarte_reward_2022} note that this leads to a positive reward spike on episode termination regardless of outcome (so long as the task reward is always non-negative).

StructLTL trains directly on reach-avoid sequences, which are equivalent to chains of automaton states. In this setting, value iteration admits the closed form $V_{\mathcal{B}^\varphi}^*(q^\varphi) = \gamma^{d(q^\varphi) - 1}$, with $d(q^\varphi)$ representing the number of transitions separating $q^\varphi$ from acceptance, and thus $\Phi(s, q^\varphi) = -\gamma^{d(q^\varphi) - 1}$. The resulting shaped reward is readily characterized: a step that advances the automaton receives no additional reward (the potential decrease is exactly absorbed by discounting), while a step that self-loops at distance $d(q^\varphi)$ receives the increment $(1 - \gamma)\gamma^{d(q^\varphi) - 1}$. Progress is thus encouraged not through one-off bonuses on advancing transitions, but through a higher per-step annuity at automaton states closer to acceptance. However, $\Phi$ under automaton-based PBRS is by construction constant in the environment state $s$; its gradient within any given automaton state is identically zero. Automaton-based PBRS therefore fundamentally cannot provide guidance \textit{within} a subgoal, only across subgoal boundaries.

Figure~\ref{fig:shaping_ablations:x2} presents results in \textit{X2DronesEnv}. Interestingly, Plato with automaton-based PBRS performs markedly \textit{worse} during early training compared to even unguided Plato, with low success rate paired with a low average episode length; this represents the agent learning to crash the drones into the world bounds and terminate the episode early. We attribute this behavior to the positive reward spikes upon termination conceded by \cite{icarte_reward_2022}; while an \textit{optimal} agent would be unbiased from these spikes, an early agent is not. To an untrained agent, the reward spike upon crashing represents a one-step, zero-credit-assignment positive reward at the world bounds. Only once the critic has internalized $\Phi$ is the advantage from the positive spike canceled and does the perverse incentive evaporate; this incurs a \qty{10}{M}-step training lag before catching up to unguided Plato. While P\textsuperscript{2}BRS can generate similar reward spikes upon termination in principle, the dense guidance that it provides toward subgoal satisfaction clearly overpowers this perverse incentive in practice.

During late training, we see slight gains in performance with automaton-based PBRS compared to unguided Plato, but ultimately no qualitative difference in behavior, unlike P\textsuperscript{2}BRS which demonstrates a substantial learning speedup to convergence.

Figure~\ref{fig:shaping_ablations:franka} presents results in \textit{FrankaCubesEnv}, in which Plato with automaton-based PBRS flatlines alongside unguided Plato. It is in this environment that the fundamental inadequacy of automaton-based PBRS is most clear; its high-level guidance has no effect if the agent is incapable of satisfying individual subgoals in the first place. This emphasizes the necessity for dense guidance towards subgoal satisfaction as is provided by P\textsuperscript{2}BRS, with which Plato converges to a strong success rate.

\subsubsection{Impact of Persistence Gating}
\label{app:ablation_studies:p2brs:persistence_gating}

When tasked with a clause such as $\mathsf{has}_i \land \mathsf{pos}_j$, the agent must learn to satisfy both propositions simultaneously. However, moving to satisfy one proposition often means moving away from satisfying the other, in this case because the cube and the target position are in different locations. Thus, the agent must learn \textit{which} proposition to satisfy first (picking up the cube), and then persistently satisfy that proposition while moving to satisfy the other (carrying it to the target position). While this \textit{implicit} temporal ordering is induced by the environment dynamics, the clause itself provides no \textit{explicit} structural guidance for it, meaning that the potential function produced by P\textsuperscript{2}BRS without persistence gating would guide the agent roughly towards the midpoint of the two locations (due to the $\min$ operator). The agent must then rely on exploration noise to discover how to satisfy the clause, which is virtually impossible for this environment. This can be resolved by the simple labeling of $\mathsf{has}$ as \textit{persistent} and $\mathsf{pos}$ (and $\mathsf{rot}$) as \textit{non-persistent}, which here represents the common-sense domain knowledge that in order to take the cube to a target position, you have to pick it up first. The agent's \textit{goal} remains unchanged, but the \textit{guidance} from P\textsuperscript{2}BRS prioritizes picking up the cube. Note that the ungated composition coincides with the plain min/max operators of \citet{li_ground-compose-reinforce_2025}, with our robustness metrics in place of GCR's learned value functions, so the results below also indicate that such operators alone are insufficient for heterogeneous conjunctions in our setting.

Figure~\ref{fig:shaping_ablations:franka} presents results in \textit{FrankaCubesEnv} without persistence gating. We see that performance quickly flatlines at around $10\%$ success rate; the agent scores highly on $\varphi_{17}$ (a sequence of position and rotation targets) but fails on all the other tasks, which involve variants of picking up cubes and carrying them to a target location or orientation. Thus, persistence gating can be essential to learn heterogeneous conjunctions with implicit temporal ordering induced by the environment dynamics.

We omit performance on \textit{X2DronesEnv} since persistence gating has no effect, because there is only one predicate ($\mathsf{pos}$); both drones can fly towards different target positions simultaneously.

\subsubsection{Impact of Adaptive Robustness Scale Calibration}
\label{app:ablation_studies:p2brs:adaptive_robustness_scale_calibration}

Recall from Section~\ref{sec:p2brs:robustness:bounded_robustness} that the robustness scale $\kappa_\mathsf{f}\tilde{g}_\mathsf{f}$ controls the breadth of guidance from the decision boundary; $\tilde{g}_\mathsf{f}$ is calibrated at the start of training and $\kappa_\mathsf{f}$ is calibrated online to track the agent's typical distance from it. In this ablation, we disable this calibration by freezing $\kappa_\mathsf{f}\tilde{g}_\mathsf{f} = 1$.

\begin{figure*}[t]
    \centering
    \includegraphics[width=0.90\linewidth]{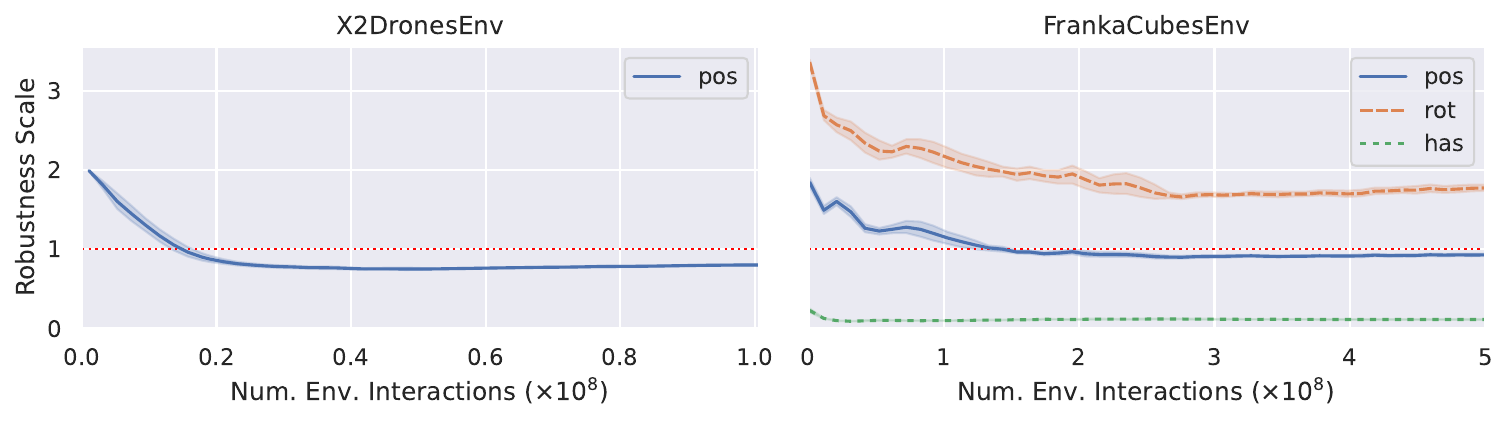}
    \caption{Robustness scales during training for Plato + P\textsuperscript{2}BRS. The dotted red line represents the frozen unit scale under the ablation of no adaptive robustness scale calibration.}
    \label{fig:robustness_scale}
    \vskip -0.1in
\end{figure*}

A frozen unit robustness scale commits every predicate, for all of training, to the sensitivity regime that its raw metric $g_\mathsf{f}$ happens to occupy at its native scale: a metric whose typical magnitudes far exceed $1$ saturates $\rho_\mathsf{f}$ towards $\pm 1$ with near-flat gradients away from a thin shell around the decision boundary, while a metric with typical magnitudes far below $1$ compresses $\rho_\mathsf{f}$ towards zero, providing weak preference between states. Whether the freeze is harmful therefore depends on how far the calibrated robustness scale sits from $1$, and on how much it moves as the trajectory distribution shifts, which Figure~\ref{fig:robustness_scale} reveals directly.

\begin{figure*}[t]
    \centering
    \subfloat[\textit{X2DronesEnv}]{\includegraphics[width=0.99\linewidth]{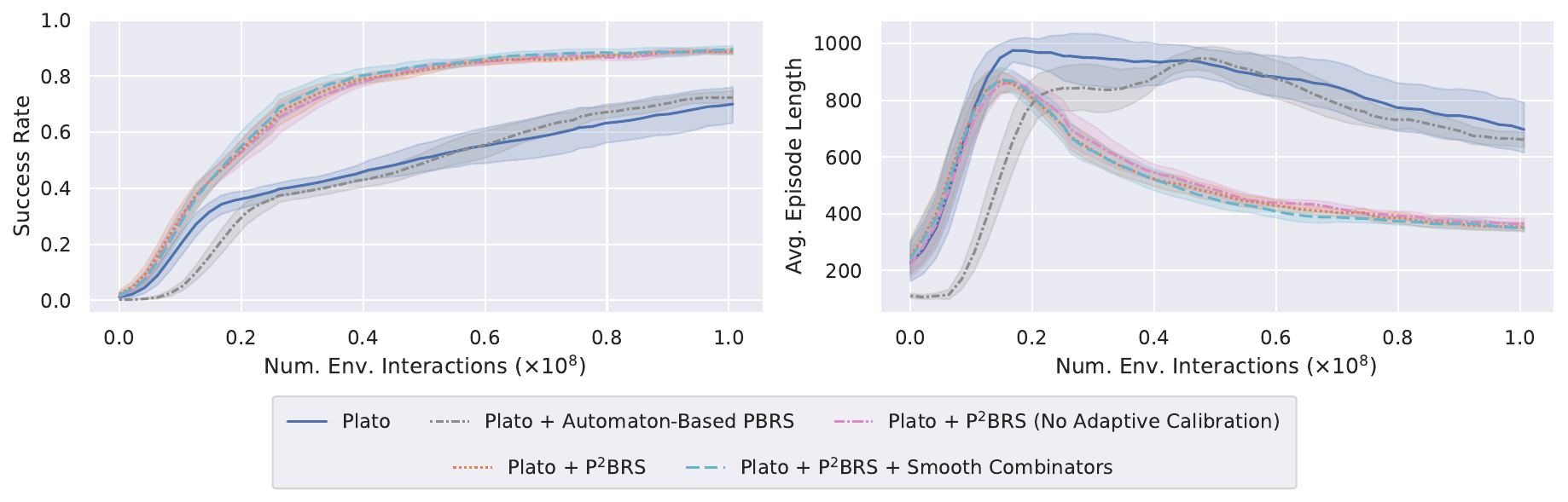} \label{fig:shaping_ablations:x2}}
    \par\bigskip
    \subfloat[\textit{FrankaCubesEnv}]{\includegraphics[width=0.99\linewidth]{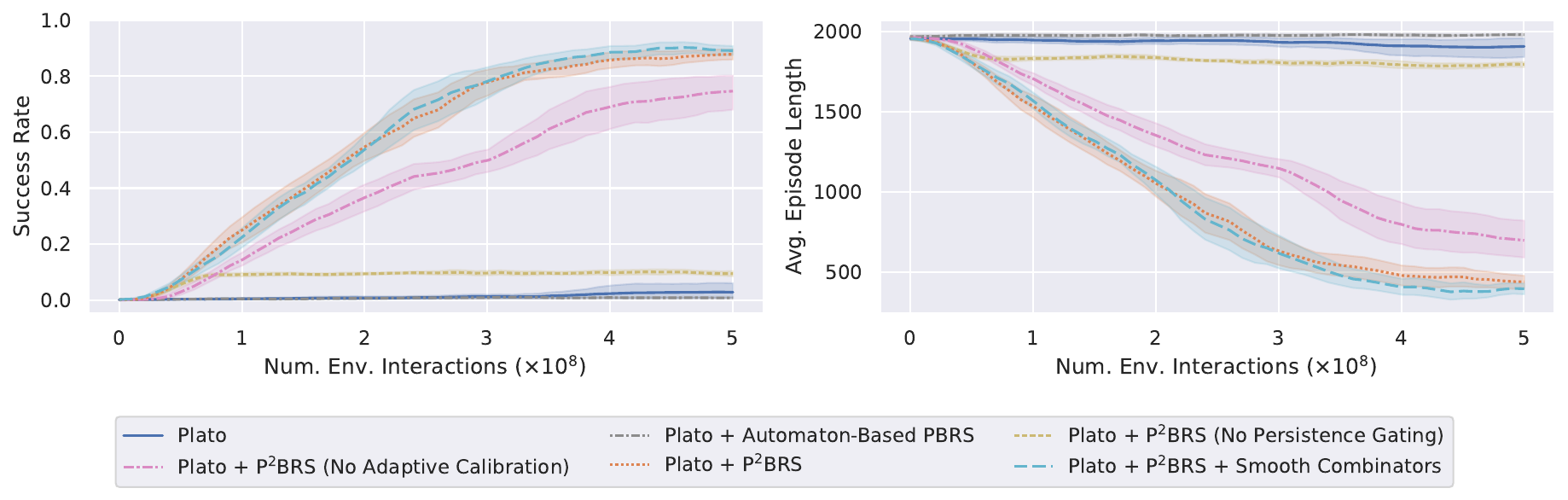} \label{fig:shaping_ablations:franka}}
    \caption{Evaluation curves for \textit{finite}-horizon specifications under each P\textsuperscript{2}BRS ablation.}
    \label{fig:shaping_ablations}
    \vskip -0.1in
\end{figure*}

In \textit{X2DronesEnv}, the calibrated scale for $\mathsf{pos}$ decays steadily over the first fifth of training before settling at around
$0.8$, close to the frozen unit scale (dotted red line in Figure~\ref{fig:robustness_scale}). A unit freeze is thus coincidentally near-calibrated for this environment, and correspondingly, Figure~\ref{fig:shaping_ablations:x2} shows performance with and without calibration to be indistinguishable. We caution against reading this as evidence that calibration is unnecessary: the agreement is an artifact of the calibrated robustness scale happening to coincide with unity.
 
\textit{FrankaCubesEnv} is the informative case. Figure~\ref{fig:robustness_scale} shows the calibrated
scales separating by predicate, with $\mathsf{rot}$ settling at around $1.8$ and $\mathsf{pos}$ at around $0.9$, both after a steady decay over the first half of training, and $\mathsf{has}$ at around $0.1$, roughly an order of magnitude below the frozen unit scale. The decay of the $\mathsf{pos}$ and $\mathsf{rot}$ scales agrees with our claim that online calibration behaves as the automatic curriculum described in Section~\ref{sec:p2brs:robustness:bounded_robustness}, providing broad guidance while early trajectories remain far from the decision boundaries and tightening it as they concentrate; a frozen scale forgoes this tracking outright. The unit freeze also mis-scales the predicate that matters most: with typical magnitudes of $g_\mathsf{has}$ roughly an order of magnitude below unity, the uncalibrated $\rho_\mathsf{has}$ is compressed towards zero, weakening guidance in the gated first phase that persistence gating prioritizes, for the subgoal (grasping the correct cube) that is hardest to discover through exploration. It additionally distorts the relative weighting of $\mathsf{rot}$ against $\mathsf{pos}$ within the non-persistent phase, where the two are aggregated by a hard $\min$. Consistent with this, Figure~\ref{fig:shaping_ablations:franka} shows the uncalibrated variant converging markedly more slowly and plateauing at a lower success rate than full P\textsuperscript{2}BRS, with correspondingly longer episode lengths.
 
Thus, Figure~\ref{fig:robustness_scale} makes visible the two roles of adaptive calibration: normalizing heterogeneous predicates onto a common scale, and tracking the shifting trajectory distribution as an automatic curriculum. The results from \textit{X2DronesEnv} demonstrate that the freeze is tolerable when the calibrated scale happens to coincide with unity, but when raw metric values are distributed far from unity, as for $\mathsf{has}$ and $\mathsf{rot}$ in \textit{FrankaCubesEnv}, calibration becomes essential. Meanwhile, its cost is negligible, requiring only quantile tracking over a sliding window of already-computed metric values.

\subsubsection{Impact of Smooth Combinators}
\label{app:ablation_studies:p2brs:smooth_combinators}

The hard $\min$/$\max$ aggregations described in Section \ref{sec:p2brs:robustness:boolean_formulae} propagate gradient to only a single operand --- the arg-min (or arg-max) --- leaving every other unsatisfied conjunct (or disjunct) with zero gradient. For a conjunctive subgoal such as $\mathsf{pos}_i\wedge\mathsf{pos}_j$ in \textit{X2DronesEnv}, this starves all but the worst-satisfied conjunct of guidance at any given time step, even when the agent can make simultaneous progress on both. For disjunctions, it greedily starves all but the best-satisfied disjunct.

Inspired by smooth $\min$/$\max$ reformulations for trajectory optimization from the STL literature \citep{han_exact_2026}, we design \textit{smooth product t-norm combinators}. Assuming bounded robustness metrics $\rho_{\beta_j}(s) \in [-1, 1]$, we soften each aggregation analytically as follows:
\begin{align*}
\beta = & \bigwedge_{j=1}^{m}\beta_j \iff \rho_\beta(s) =
\begin{cases}
\displaystyle\prod_{j=1}^{m}\bigl(1-\max\big\{-\rho_{\beta_j}(s),0\big\}\bigr) - 1 & \text{if }
\min\limits_{j\in 1:m}\rho_{\beta_j}(s) < 0, \\
\min\limits_{j\in 1:m}\rho_{\beta_j}(s) & \text{else,}
\end{cases} \\
\beta = & \bigvee_{j=1}^{m}\beta_j \iff \rho_\beta(s) =
\begin{cases}
-\displaystyle\prod_{j=1}^{m}\bigl(-\rho_{\beta_j}(s)\bigr) & \text{if }
\max\limits_{j\in 1:m}\rho_{\beta_j}(s) < 0, \\
\max\limits_{j\in 1:m}\rho_{\beta_j}(s) & \text{else.}
\end{cases}
\end{align*}

The conjunction yields a gradient term $\partial\rho_\beta/\partial\rho_{\beta_k} = \mathds{1}(\rho_{\beta_k}(s)<0)\prod_{j\neq k}(1-\max\{-\rho_{\beta_j}(s),0\})$ that pulls each unsatisfied conjunct $\rho_{\beta_k}$ towards satisfaction, not just the furthest; the gradient magnitude grows the closer the other conjuncts are to satisfaction, which is a smooth approximation of the hard $\min$ aggregation gradient. Meanwhile, the disjunction yields a gradient term $\partial\rho_\beta/\partial\rho_{\beta_k} = \prod_{j\neq k}(-\rho_{\beta_j}(s))$ within the all-negative orthant that pulls each disjunct $\rho_{\beta_k}$ towards satisfaction, not just the closest; the gradient magnitude grows the further the other disjuncts are from satisfaction, which is a smooth approximation of the hard $\max$ aggregation gradient.

Note that smoothing is applied only while $\beta$ is unsatisfied, and the satisfied regions retain the hard $\min$/$\max$ aggregation. The smoothed combinators are sign-exact to their hard counterparts, so the labeling function $L_\beta(s)=\mathcal{H}\bigl(\rho_\beta(s)\bigr)$ is preserved, and their outputs are bounded to $[-1,1]$. Importantly, they also reduce to identity in the case of a single conjunct or disjunct.

Figure \ref{fig:shaping_ablations} presents results in \textit{X2DronesEnv} and \textit{FrankaCubesEnv} with P\textsuperscript{2}BRS augmented with smooth combinators; in both environments, performance gains relative to standard P\textsuperscript{2}BRS are inconsequential. This is somewhat to be expected for \textit{FrankaCubesEnv}, wherein persistence gating overrides the smooth $\min$ combinator when $\mathsf{has}$ is present in a clause (which is true for the majority of clauses). However, the trend persists in \textit{X2DronesEnv}, which represents the ideal use case with two drones simultaneously moving to different target positions.

The reason boils down to multi-step credit assignment. Since potential-based shaping leaves the optimal policy unchanged for any potential function, the smoothness of the combinator can influence learning dynamics only, and the natural concern is that the hard $\min$ combinator provides no gradient for any conjunct other than the current minimizer. This flatness is an \textit{instantaneous} property, however, and the \textit{learning} signal is not instantaneous. Over a trajectory segment of many steps the minimizer is likely to alternate between conjuncts, especially during early learning when policies are highly stochastic, and improvement itself drives the minimum up to meet the next conjunct. Thus, the gradient contribution typically alternates between conjuncts over a segment. The learning signal aggregates these contributions: each GAE advantage estimate sums TD residuals, and with them the shaping increments, over an effective horizon of many steps, and the critic's bootstrapped values propagate them further. The update an action receives is therefore a smoothed temporal average of per-conjunct gradients, weighted by the time each conjunct spends as the minimizer, much as the smooth combinator weights conjunct gradients by their robustness gaps at a single step.

For segments with a leading conjunct whose robustness stays well clear of the minimum (and therefore never alternates as the minimizer), the smooth combinator's sensitivity to the leading conjunct is low, decaying toward $0$ as the robustness gap increases; thus, the benefit of analytical smoothing would be inconsequential whenever temporal smoothing from alternation does not already apply. This carries over to disjunctions; near-tied disjuncts are temporally smoothed by alternation of the maximizer, as above, and when one disjunct leads, the analytical smooth maximum's weights on the trailing disjuncts decay toward $0$.

Multi-step advantage estimation thus already performs the gradient smoothing that smooth combinators provide analytically, making the latter redundant. Accordingly, we omit smooth combinators from the proposed P\textsuperscript{2}BRS design in the main paper, since we cannot justify their usefulness theoretically or empirically.

%%%%%%%%%%%%%%%%%%%%%%%%%%%%%%%%
% LIMITATIONS
%%%%%%%%%%%%%%%%%%%%%%%%%%%%%%%%

\section{Limitations}
\label{app:limitations}

\textbf{Specification and vocabulary.} While PlatoLTL generalizes parametrically across instances of known predicates, it cannot generalize to entirely new ones, and thus requires all predicates in the evaluation set to be represented within the training curriculum. Note that prior LTL-guided RL approaches share an analogous limitation, requiring all propositions to be known a priori; PlatoLTL's predicate-based formulation is strictly more general. Moreover, a parameterization always exists, whether trivially (each proposition as its own zero-parameter predicate) or via a predicate parameterized by satisfying sets $\mathcal{G} \subseteq \mathcal{S}$; the gains of the Plato embedding module instead require parameters that are both explicitly specifiable and structurally related to the labeling function (cf.\ Appendix~\ref{app:ablation_studies:pca_prop_embeddings}), where the former construction lacks the structure and the latter the specifiability. The choice of specification granularity behaves similarly: under the degenerate specification $\F \mathsf{done}$, the automaton exposes no subgoals, and PlatoLTL reduces to sparse-reward RL for the monolithic task, whereas a granular specification over simpler predicates exposes subgoals as individually shaped milestones, at the cost of prescribing one particular means of achieving the task (cf.\ Appendix~\ref{app:plato_theory_of_forms}). The specification and vocabulary are thus modeling choices: we hypothesize that a small number of well-chosen predicates suffices to capture a meaningful range of tasks, as small libraries of parameterized action primitives do for manipulation \citep{hausknecht_deep_2016, dalal_accelerating_2021}, and we discuss automating these choices via predicate discovery as future work in Appendix~\ref{app:future_work}.

\textbf{Robustness metrics.} The simple pose-based priors used throughout our environments assume access to the poses of task-relevant entities; these are directly available in simulation, and mature methods exist for estimating them from raw sensing \citep{xiang_posecnn_2018, labbe_megapose_2022, wen_foundationpose_2024}. Since shaping with any bounded potential leaves the optimal policy unchanged, estimation error degrades only the quality of guidance; this is in contrast to the labeling function, whose accuracy directly determines task satisfaction semantics. A separate challenge is the construction of informative metrics. A valid metric always exists, as any metric on $\mathcal{S}$ induces one via the signed distance to a predicate's satisfying set; however, for semantically complex predicates (e.g., that an assembly is complete, or that a deformable object is folded), the satisfying set is specified only implicitly, and a hand-specified surrogate must instead compose multiple geometric quantities in nontrivial and potentially misleading ways. The challenge is thus one of construction cost rather than of existence. Under weak or trivial metrics, P\textsuperscript{2}BRS degrades gracefully to unguided exploration for the affected predicate; however, as our results demonstrate, unguided exploration can be the bottleneck for learning to satisfy the hardest predicates. We further note that while P\textsuperscript{2}BRS guarantees invariance of the optimal policy, it does not guarantee that shaping helps: an actively misleading surrogate could in principle slow learning relative to no shaping at all, and the practical burden is thus to supply metrics that are coarsely correct (or else use the trivial one).

\textbf{Persistence annotations.} Persistence annotations (Section~\ref{sec:p2brs:robustness:boolean_formulae}) are likewise coarse, manually supplied domain knowledge, and affect only the guidance and not the optimal policy; however, given that persistence gating can be essential for heterogeneous conjunctions (Appendix~\ref{app:ablation_studies:p2brs:persistence_gating}), a poorly annotated vocabulary could forfeit much of the benefit of P\textsuperscript{2}BRS. A predicate's persistence may also in principle be context-dependent rather than fixed; inferring persistence automatically, e.g., from demonstration trajectories, is an interesting direction for removing this annotation burden.

%%%%%%%%%%%%%%%%%%%%%%%%%%%%%%%%
% FUTURE WORK
%%%%%%%%%%%%%%%%%%%%%%%%%%%%%%%%

\section{Directions for Future Work}
\label{app:future_work}

\textbf{Uncertain Labeling Functions.} Like most work in LTL-guided RL, we assume access to a ground-truth labeling function; in real-world applications, labels would instead be estimated by perception and are therefore uncertain. A natural direction for future work is making PlatoLTL robust in this regime. Prior work handles uncertain labels via joint action and query policies \citep{hatanaka_reinforcement_2023} or beliefs over automaton states \citep{li_reward_2024}; such approaches operate at the labeling and automaton layer, and are thus, in principle, compatible with PlatoLTL, whose contributions concern the goal embedding and reward shaping. An open question is how the potential function of P\textsuperscript{2}BRS should be evaluated when the automaton state itself becomes a belief, e.g., in expectation.

\textbf{Learned Labeling Functions and Robustness Metrics.} A complementary direction is to learn the labeling functions from data, e.g., as per-predicate success classifiers $\hat{L}_{\mathsf{f}}(s, x_{\mathsf{f}})$ trained on labeled outcomes \citep{fu_variational_2018, eysenbach_replacing_2021, du_vision-language_2023}. Notably, this would unify the labeling function and the robustness metric into a single learned artifact: thresholding the classifier score recovers the label, while the score itself serves as a raw metric satisfying sign-consistency by construction, offering a route to metrics for the semantically complex predicates of Appendix~\ref{app:limitations}. Denser guidance could instead be derived from goal-conditioned value functions or temporal distances, pre-trained on passive video or demonstrations \citep{ma_liv_2023, ma_vip_2023} or learned per proposition via offline RL from labeled trajectories \citep{li_ground-compose-reinforce_2025}; such metrics need not agree in sign with the labeling function, since the construction of Section~\ref{sec:p2brs:robustness} repairs any raw metric $\hat{g}_{\mathsf{f}}$ into a valid $g_{\mathsf{f}}$ given $L_{\mathsf{f}}$, and the adaptive calibration of Section~\ref{sec:p2brs:robustness:bounded_robustness} consumes metrics of arbitrary scale without manual tuning. Under our predicate formulation, such models would instead be conditioned on the parameters, so that their number scales with $|\mathit{Pred}|$ rather than with the number of propositions. Learned metrics do, however, require training data with adequate state coverage, which is unavailable for the environments studied here; our simple priors avoid this requirement.

\textbf{Automatic Predicate Discovery.} A promising direction for relaxing the assumption that predicates are known a priori (Appendix~\ref{app:limitations}) is automatic predicate discovery: recent work explores the discovery of grounded predicates from raw observations or natural language \citep{asai_unsupervised_2019, pang_natural_2023, gupta_learning_2025}. Future work could leverage these methods to identify useful predicates and parameterizations in a pre-training phase, which are then consumed downstream by PlatoLTL.

\textbf{LTL from Natural Language.} Another natural direction for future work is grounding PlatoLTL in natural language (NL) interfaces. Like all LTL-guided RL approaches, PlatoLTL and its backbones require tasks to be specified in LTL syntax, which trades accessibility for precision and verifiability. This is in contrast to approaches such as Vision-Language-Action models (VLAs; \citealp{zitkovich_rt-2_2023, kim_openvla_2024, black_pi_0_2025}), which condition agents directly on NL.

Formal specification languages such as LTL come with the benefit of being unambiguous and verifiable, allowing users to precisely define the problem and rigorously monitor task progress. LTL is well-established in formal verification, where it is routinely used to specify properties of safety-critical systems such as nuclear power plants and aircraft control systems \citep{baier_principles_2008}; indeed, safety-critical applications often require LTL specifications for this reason. Our \textit{FalloutWorld} environment is one such example, representing inspection of nuclear sites. Furthermore, unlike VLAs, LTL-guided approaches like PlatoLTL require comparatively small, inexpensive networks while still generalizing across tasks with high specification complexity.

Recent work demonstrates grounded automatic translation from NL to LTL \citep{fuggitti_nl2ltl_2023}, as well as zero-shot autoformalization of reward machines from NL reward descriptions \citep{li_ground-compose-reinforce_2025}. This suggests that PlatoLTL could, in principle, inherit the accessibility of NL interfaces while retaining the interpretability and verifiability benefits of LTL specifications.

\clearpage

\clearpage

\begin{table*}[t]
    % Main Caption
    \caption{Template LTL specifications used for evaluation. The notation $\mathsf{f}_i$ represents the $i$-th instance of the predicate $\mathsf{f}$.}
    \label{tab:formulae}
    
    \centering
    \begin{tiny}

    % --- Finite Horizon ---
    \begin{subtable}[t]{\textwidth}
    \caption{Finite-horizon specifications $\varphi$}
    \label{tab:formulae:finite}
    \centering
    \setlength{\tabcolsep}{1.3pt}

    \begin{tabular}{llll}
        \toprule

        & $\varphi$ & Specification & Summary \\

        \midrule
        
        % ================= Go1RGBZoneEnv =================
        \multirow{10}{*}{\rotatebox[origin=c]{90}{Go1RGBZoneEnv}} 
        
        & $\varphi_{1}$  & $\F (\mathsf{at}_0 \land (\lnot \mathsf{at}_1 \U \mathsf{at}_2)) \land \F \mathsf{at}_3$ & Sequential reach with temporary obstacle avoidance \\
        
        & $\varphi_{2}$  & $\F \mathsf{at}_0 \land (\lnot \mathsf{at}_0 \U (\mathsf{at}_1 \land \F \mathsf{at}_2))$ & Guarded sequential reach \\
        
        & $\varphi_{3}$  & $\F (\mathsf{at}_0 \lor \mathsf{at}_1) \land \F \mathsf{at}_2 \land \F \mathsf{at}_3$ & Unordered reach with choice \\
        
        & $\varphi_{4}$  & $\lnot (\mathsf{at}_0 \lor \mathsf{at}_1) \U (\mathsf{at}_2 \land \F \mathsf{at}_3)$ & Sequential reach with global obstacle avoidance \\
        
        & $\varphi_{5}$  & $\lnot \mathsf{at}_0 \U ((\mathsf{at}_1 \lor \mathsf{at}_2) \land (\lnot \mathsf{at}_0 \U \mathsf{at}_3))$ & Sequential reach with choice and obstacle avoidance \\
        
        & $\varphi_{6}$  & $((\mathsf{at}_0 \lor \mathsf{at}_1) \Rightarrow (\lnot \mathsf{at}_2 \U \mathsf{at}_3)) \U \mathsf{at}_2$ & Reach with post-hazard recovery \\
        
        & $\varphi_{7}$  & $\F (\mathsf{at}_0 \land \X \F (\mathsf{at}_1 \land \X \F (\mathsf{at}_2 \land \X \F (\mathsf{at}_3 \land \X \F (\mathsf{at}_2 \land \X \F (\mathsf{at}_0$ & $12$-step reach \\
        &                & $\land \X \F (\mathsf{at}_3 \land \X \F (\mathsf{at}_1 \land \X \F (\mathsf{at}_3 \land \X \F (\mathsf{at}_1 \land \X \F (\mathsf{at}_2 \land \X \F \mathsf{at}_0)))))))))))$ & \\
        
        & $\varphi_{8}$  & $\lnot \mathsf{at}_0 \U (\mathsf{at}_1 \land ( \lnot \mathsf{at}_2 \U (\mathsf{at}_0 \land ( \lnot \mathsf{at}_3 \U (\mathsf{at}_2 \land ( \lnot \mathsf{at}_0 \U (\mathsf{at}_3$ & $8$-step reach-avoid \\
        &                & $\land ( \lnot \mathsf{at}_2 \U (\mathsf{at}_1 \land ( \lnot \mathsf{at}_0 \U (\mathsf{at}_2 \land ( \lnot \mathsf{at}_3 \U (\mathsf{at}_0 \land ( \lnot \mathsf{at}_1 \U \mathsf{at}_3))))))))))))))$ & \\
        
        \midrule
        
        % ================= X2DronesEnv =================
        \multirow{8}{*}{\rotatebox[origin=c]{90}{X2DronesEnv}} 
        
        & $\varphi_{9}$  & $\lnot \mathsf{pos}_0 \U (\mathsf{pos}_1 \lor (\mathsf{pos}_2 \land \mathsf{pos}_3))$ & Reach with choice and global obstacle avoidance \\

        & $\varphi_{10}$ & $\F \mathsf{pos}_0 \land (\lnot (\mathsf{pos}_1 \lor \mathsf{pos}_2) \U \mathsf{pos}_3)$ &  Reach with temporary obstacle avoidance \\
        
        & $\varphi_{11}$ & $((\mathsf{pos}_0 \lor \mathsf{pos}_1 \lor \mathsf{pos}_2) \Rightarrow \F (\mathsf{pos}_3 \land \mathsf{pos}_4)) \U (\mathsf{pos}_2 \land \mathsf{pos}_3)$ & Simultaneous reach with post-hazard recovery \\
        
        & $\varphi_{12}$ & $\F (\mathsf{pos}_0 \land \F ((\mathsf{pos}_1 \land \mathsf{pos}_2) \land \F ((\mathsf{pos}_2 \land \mathsf{pos}_3) \land (\lnot (\mathsf{pos}_0 \lor \mathsf{pos}_1) \U (\mathsf{pos}_3 \land \mathsf{pos}_4)))))$ & Mixed sequence including simultaneous reach \\
        
        & $\varphi_{13}$ & $\lnot (\mathsf{pos}_0 \lor \mathsf{pos}_1 \lor (\mathsf{pos}_2 \land \mathsf{pos}_3)) \U ((\mathsf{pos}_3 \land \mathsf{pos}_4) \land \F (\mathsf{pos}_4 \land \mathsf{pos}_5))$ & Simultaneous reach with global obstacle avoidance \\
        
        & $\varphi_{14}$ & $\F (\mathsf{pos}_0 \land \F (\mathsf{pos}_1 \land \F (\mathsf{pos}_2 \land \F (\mathsf{pos}_3 \land \F (\mathsf{pos}_4 \land \F (\mathsf{pos}_0 \land \F (\mathsf{pos}_2 \land \F \mathsf{pos}_4)))))))$ & $8$-step reach \\

        & $\varphi_{15}$ & $(\lnot \mathsf{pos}_0 \U (\mathsf{pos}_1 \land \F \mathsf{pos}_2)) \lor (\lnot \mathsf{pos}_0 \U (\mathsf{pos}_3 \land \F \mathsf{pos}_4))$ & Choice of two reach-avoid paths \\

        & $\varphi_{16}$ & $(\lnot (\mathsf{pos}_0 \lor \mathsf{pos}_1) \U \mathsf{pos}_2) \land \F (\mathsf{pos}_0 \land \mathsf{pos}_1)$ & Guarded simultaneous reach \\

        \midrule

        % ================= FrankaCubesEnv =================
        \multirow{11}{*}{\rotatebox[origin=c]{90}{FrankaCubesEnv}} 

        & $\varphi_{17}$ & $\F (\mathsf{pos}_0 \land \F (\mathsf{rot}_0 \land \F (\mathsf{pos}_1 \land \F (\mathsf{pos}_2 \land \F (\mathsf{rot}_1$ & $9$-step sequence of position and orientation targets \\
        &                & $\land \F (\mathsf{pos}_3 \land \F (\mathsf{pos}_0 \land \F (\mathsf{rot}_0 \land \F \mathsf{pos}_2))))))))$ & \\

        & $\varphi_{18}$ & $\lnot \mathsf{pos}_0 \U (\mathsf{pos}_1 \land \mathsf{has}_0)$ & Pick-and-carry with global obstacle avoidance \\

        & $\varphi_{19}$ & $\F (\mathsf{has}_0 \land \F ((\mathsf{has}_0 \land \mathsf{pos}_0) \land \F (\mathsf{has}_0 \land \mathsf{pos}_1 \land \mathsf{rot}_0)))$ & Pick-and-carry to a target pose \\

        & $\varphi_{20}$ & $\F ((\mathsf{has}_0 \land \mathsf{rot}_0) \land \F ((\mathsf{has}_0 \land \mathsf{rot}_1) \land \F (\mathsf{has}_0 \land (\mathsf{pos}_0 \lor \mathsf{pos}_1))))$ & Oriented carry with choice of position target \\

        & $\varphi_{21}$ & $\lnot \mathsf{pos}_0 \U (\mathsf{has}_0 \land (\lnot \mathsf{pos}_1 \U (\mathsf{has}_0 \land \mathsf{pos}_2)))$ & Pick-and-carry with phased obstacle avoidance \\

        & $\varphi_{22}$ & $\F ((\mathsf{has}_0 \land \mathsf{pos}_0) \land (\mathsf{pos}_0 \U (\lnot \mathsf{has}_0 \land \mathsf{pos}_0)))$ & Unordered $2$-object pick-and-place \\
        &                & $\land \F ((\mathsf{has}_1 \land \mathsf{pos}_1) \land (\mathsf{pos}_1 \U (\lnot \mathsf{has}_1 \land \mathsf{pos}_1)))$ & \\

        & $\varphi_{23}$ & $\F ((\mathsf{has}_0 \land \mathsf{pos}_0) \land (\mathsf{pos}_0 \U ((\lnot \mathsf{has}_0 \land \mathsf{pos}_0)$ & Pick-and-place with relocation \\
        &                & $\land \F ((\mathsf{has}_0 \land \mathsf{pos}_1) \land (\mathsf{pos}_1 \U (\lnot \mathsf{has}_0 \land \mathsf{pos}_1))))))$ & \\

        & $\varphi_{24}$ & $\F ((\mathsf{has}_0 \land (\mathsf{pos}_0 \lor \mathsf{pos}_1)) \land ((\mathsf{pos}_0 \lor \mathsf{pos}_1) \U (\lnot \mathsf{has}_0 \land (\mathsf{pos}_0 \lor \mathsf{pos}_1))))$ & Pick-and-place with choice of target position \\
        
        \midrule
        
        % ================= FalloutWorld =================
        \multirow{8}{*}{\rotatebox[origin=c]{90}{FalloutWorld}} 
        
        & $\varphi_{25}$ & $\lnot (\mathsf{rad}_0 \lor \mathsf{loc}_0) \U ((\mathsf{loc}_1 \land \lnot \mathsf{rad}_0) \land \F \mathsf{loc}_2)$ & Reach with temporary radiation avoidance \\
        
        & $\varphi_{26}$ & $\lnot \mathsf{rad}_0 \U ((\mathsf{loc}_0 \lor \mathsf{loc}_1) \land (\lnot \mathsf{rad}_0 \U (\mathsf{loc}_2 \land \lnot \mathsf{rad}_0)))$ & Reach with global radiation avoidance \\
        
        & $\varphi_{27}$ & $(\mathsf{rad}_0 \Rightarrow (\lnot \mathsf{loc}_0 \U (\mathsf{loc}_1 \lor \mathsf{loc}_2))) \U (\mathsf{loc}_0 \land \lnot \mathsf{rad}_0)$ & Reach with post-hazard recovery (decontamination) \\
        
        & $\varphi_{28}$ & $(\lnot \mathsf{rad}_0 \U (\mathsf{loc}_0 \land \lnot \mathsf{rad}_0)) \land \F (\mathsf{rad}_0 \land \F \mathsf{loc}_1)$ & Radiation-free reach followed by radiation probing \\
        
        & $\varphi_{29}$ & $\F \mathsf{loc}_0 \land \F \mathsf{loc}_1 \land (\lnot \mathsf{rad}_0 \U (\mathsf{loc}_2 \land \lnot \mathsf{rad}_0))$ & Unordered rad-sensitive and rad-agnostic reach \\
        
        & $\varphi_{30}$ & $(\lnot (\mathsf{loc}_0 \lor \mathsf{loc}_1) \U \mathsf{rad}_0) \land \F (\mathsf{loc}_0 \land \F \mathsf{loc}_1)$ & Sequential reach guarded by radiation probing \\

        & $\varphi_{31}$ & $\F (\mathsf{loc}_0 \land \F (\mathsf{loc}_1 \land \F (\mathsf{loc}_2 \land \F (\mathsf{rad}_0 \land \F (\mathsf{loc}_3 \land \F (\mathsf{loc}_4 \land \F \mathsf{loc}_5))))))$ & $7$-step reach including radiation probe \\

        & $\varphi_{32}$ & $((\lnot \mathsf{rad}_0 \U (\mathsf{loc}_0 \land \lnot \mathsf{rad}_0)) \land \F \mathsf{loc}_1) \lor ((\lnot \mathsf{rad}_0 \U (\mathsf{loc}_2 \land \lnot \mathsf{rad}_0)) \land \F \mathsf{loc}_3)$ & Choice of two paths with temporary rad-avoidance \\
        
        \bottomrule
    \end{tabular}
        
    \end{subtable}
    
    \par\bigskip
    
    % --- Infinite Horizon ---
    \begin{subtable}[t]{\textwidth}
    \caption{Infinite-horizon specifications $\psi$}
    \label{tab:formulae:infinite}
    \centering
    \setlength{\tabcolsep}{1.3pt}
        
    \begin{tabular}{llll}
        \toprule

        & $\psi$ & Specification & Summary \\

        \midrule
        
        % ================= Go1RGBZoneEnv =================
        \multirow{6}{*}{\rotatebox[origin=c]{90}{Go1RGBZoneEnv}} 
        
        & $\psi_{1}$  & $\F \G \mathsf{at}_0$ & Persistent reach \\
        
        & $\psi_{2}$  & $\F \G \mathsf{at}_0 \land \F (\mathsf{at}_1 \land \F \mathsf{at}_2)$ & Persistent reach and sequential reach \\
        
        & $\psi_{3}$  & $\F \G \mathsf{at}_0 \land \G \lnot \mathsf{at}_1$ & Persistent reach with global obstacle avoidance \\
        
        & $\psi_{4}$  & $\G ((\mathsf{at}_0 \lor \mathsf{at}_1) \Rightarrow \F \mathsf{at}_2) \land \F \G (\mathsf{at}_0 \lor \mathsf{at}_3)$ & Persistent reach with post-hazard recovery \\
        
        & $\psi_{5}$  & $\G \F \mathsf{at}_0 \land \G \F \mathsf{at}_1$ & $2$-step cycle \\
        
        & $\psi_{6}$  & $\G \F \mathsf{at}_0 \land \G \F \mathsf{at}_1 \land \G \F \mathsf{at}_2 \land \G \lnot \mathsf{at}_3$ & Unordered $3$-step cycle with global obstacle avoidance \\[.1cm]
        
        \midrule

        % ================= X2DronesEnv =================
        \multirow{6}{*}{\rotatebox[origin=c]{90}{X2DronesEnv}} 

        & $\psi_{7}$  & $\G \F \mathsf{pos}_0 \land \G \F \mathsf{pos}_1 \land \G \F \mathsf{pos}_2 \land \G \F \mathsf{pos}_3 \land \G \F \mathsf{pos}_4 \land \G \F \mathsf{pos}_5$ & Unordered $6$-step cycle \\
        
        & $\psi_{8}$  & $\G \F (\mathsf{pos}_0 \land \mathsf{pos}_1) \land \G \F \mathsf{pos}_2 \land \G \lnot (\mathsf{pos}_3 \lor \mathsf{pos}_4)$ & $2$-step mixed cycle with global obstacle avoidance \\
        
        & $\psi_{9}$  & $\G \F (\mathsf{pos}_0 \land \X \F (\mathsf{pos}_1 \land \X \F (\mathsf{pos}_2 \land \X \F (\mathsf{pos}_1 \land \X \F \mathsf{pos}_0)))) \land \G \lnot \mathsf{pos}_3$ & Ordered $4$-step cycle with global obstacle avoidance \\
        
        & $\psi_{10}$ & $\G \F (\mathsf{pos}_0 \land \mathsf{pos}_1) \land \G \F (\mathsf{pos}_2 \land \mathsf{pos}_3) \land \F (\mathsf{pos}_4 \land \F (\mathsf{pos}_0 \land \F \mathsf{pos}_2)) \land \G \lnot \mathsf{pos}_5$ & $2$-step cycle and sequential reach with global avoidance \\
        
        & $\psi_{11}$ & $\G \F \mathsf{pos}_0 \land \G \F \mathsf{pos}_1 \land \G \F \mathsf{pos}_2 \land \G ((\mathsf{pos}_3 \lor \mathsf{pos}_4) \Rightarrow \lnot \mathsf{pos}_0 \U \mathsf{pos}_1)$ & Unordered $3$-step cycle with post-hazard recovery \\
        
        & $\psi_{12}$ & $\G \F (\mathsf{pos}_0 \land \mathsf{pos}_1) \land \G \F (\mathsf{pos}_2 \land \mathsf{pos}_3) \land \G \lnot (\mathsf{pos}_0 \land \mathsf{pos}_2)$ & $2$-step simultaneous reach cycle with a forbidden overlap \\

        \midrule

        % ================= FrankaCubesEnv =================
        \multirow{7}{*}{\rotatebox[origin=c]{90}{FrankaCubesEnv}} 

        & $\psi_{13}$ & $\G \F \mathsf{pos}_0 \land \G \F \mathsf{pos}_1 \land \G \F \mathsf{pos}_2 \land \G \F \mathsf{pos}_3$ & Unordered $4$-step cycle of position targets \\
        
        & $\psi_{14}$ & $\G \F \mathsf{pos}_0 \land \G \F \mathsf{pos}_1 \land \G \F \mathsf{pos}_2 \land \G \lnot \mathsf{pos}_3 \land (\lnot \mathsf{pos}_1 \U \mathsf{pos}_0) \land (\lnot \mathsf{pos}_2 \U \mathsf{pos}_1)$ & $3$-step cycle with ordered start and global avoidance \\
        
        & $\psi_{15}$ & $\G \F (\mathsf{pos}_0 \land \F (\mathsf{rot}_0 \land \F (\mathsf{pos}_1 \land \F \mathsf{rot}_1))) \land \G \lnot \mathsf{pos}_2$ & Ordered $4$-step mixed cycle with global obstacle avoidance \\
        
        & $\psi_{16}$ & $(\lnot \mathsf{has}_0 \U (\mathsf{pos}_0 \lor \mathsf{pos}_1)) \land \G \lnot(\mathsf{has}_0 \land (\mathsf{pos}_0 \lor \mathsf{pos}_1))$ & Pick-and-cycle with guarded pick and avoid while carrying \\
        &             & $\land \G \F (\mathsf{has}_0 \land \mathsf{pos}_2) \land \G \F (\mathsf{has}_0 \land \mathsf{pos}_3)$ \\
        
        & $\psi_{17}$ & $\G \F (\mathsf{has}_0 \land\mathsf{pos}_0) \land \G \F (\mathsf{has}_1 \land \mathsf{pos}_1)$ & Pick-and-carry cycle alternating between two cubes \\
        
        & $\psi_{18}$ & $\G \F \mathsf{has}_0 \land \G \F \lnot \mathsf{has}_0 \land \G (\mathsf{has}_0 \Rightarrow \F (\mathsf{pos}_0 \land \F \mathsf{pos}_1))$ & Recurrent pick-carry-release \\

        \midrule
        
        % ================= FalloutWorld =================
        \multirow{6}{*}{\rotatebox[origin=c]{90}{FalloutWorld}} 
        
        & $\psi_{19}$ & $\G (\mathsf{rad}_0 \Rightarrow \F (\mathsf{loc}_0 \land \F \mathsf{loc}_1)) \land \G \F \mathsf{loc}_2 \land \G \F \mathsf{loc}_3$ & $2$-step cycle with post-hazard recovery (decontamination) \\
        
        & $\psi_{20}$ & $\G \F \mathsf{loc}_0 \land \G \F \mathsf{loc}_1 \land \G \F \mathsf{loc}_2 \land \G \lnot \mathsf{rad}_0$ & Unordered $3$-step cycle with global radiation avoidance \\
        
        & $\psi_{21}$ & $\G \F (\mathsf{loc}_0 \land \F (\mathsf{loc}_1 \land \F (\mathsf{loc}_2 \land \F \mathsf{loc}_3))) \land \G \lnot \mathsf{rad}_0$ & Ordered $4$-step cycle with global radiation avoidance \\
        
        & $\psi_{22}$ & $\G \F \mathsf{loc}_0 \land \G \F \mathsf{loc}_1 \land \G \F \mathsf{loc}_2 \land \G \F \mathsf{loc}_3 \land \G \F \mathsf{loc}_4 \land \G \F \mathsf{loc}_5$ & Unordered $6$-step cycle \\
        
        & $\psi_{23}$ & $\G \F (\mathsf{rad}_0 \land \F (\mathsf{loc}_0 \land \F (\lnot \mathsf{rad}_0 \land \F \mathsf{loc}_1)))$ & Ordered $4$-step cycle including radiation cycling \\
        
        & $\psi_{24}$ & $\G \F \mathsf{rad}_0 \land \G \F \mathsf{loc}_0 \land \G \F \mathsf{loc}_1 \land \G (\mathsf{rad}_0 \Rightarrow (\lnot (\mathsf{loc}_0 \lor \mathsf{loc}_1) \U \mathsf{loc}_2))$ & Unordered $3$-step cycle including cyclical decontamination \\
        
        \bottomrule
    \end{tabular}
    \end{subtable}
    \end{tiny}
    \vskip -0.1in
\end{table*}

\clearpage

\begin{table*}[t]
    \caption{Details regarding the curriculum for sampling reach-avoid sequences during training.}

    \label{tab:curriculum}
    \begin{center}

    \begin{subtable}[t]{\textwidth}
    \caption{Curriculum stages. For each stage, we list the sequence type (reach-avoid/reach-stay), the probability $\mathrm{P}_\mathrm{seq}$ of sampling a sequence of that type, the number of steps $n$ (or, for reach-stay tasks, the number of consecutive time steps for satisfying the ``stay'' sub-task), the sets of Boolean formulae from which each $\beta_i^+$ and $\beta_i^-$ are sampled, and the probability $\mathrm{P}_\mathrm{avd}$ of sampling an active avoid-formula $\beta_i^- \not= \false$ at each subgoal. Avoid-formulae that would overlap with the current or preceding reach-formula are discarded and resampled.}
    
    \label{tab:curriculum:stages}
    \begin{center}
    
    \begin{tabular}{ll ccccc}
        \toprule
        
        Stage & Type & $\mathrm{P}_\mathrm{seq}$ & $n$ & $\beta_i^+$ & $\beta_i^-$ & $\mathrm{P}_\mathrm{avd}$ \\
        
        \midrule
        
        1                  & Reach-Avoid & 1.0 & 1      & $\mathit{AP}$   & $\false$       & 0.0 \\
        
        \midrule
        
        2                  & Reach-Avoid & 1.0 & 1      & $\mathcal{V}_r$ & $\false$       & 0.0 \\
        
        \midrule
        
        3                  & Reach-Avoid & 1.0 & 1      & $\mathcal{V}_r$ & $\mathcal{V}_n$ & 0.5 \\
        
        \midrule
        
        4                  & Reach-Avoid & 1.0 & 1      & $\mathcal{V}_r$ & $\mathcal{V}_w$ & 0.7 \\
        
        \midrule
        
        5                  & Reach-Avoid & 1.0 & [1, 2] & $\mathcal{V}_r$ & $\mathcal{V}_w$ & 0.7 \\
        
        \midrule
        
        \multirow{2}{*}{6} & Reach-Avoid & 0.4 & [1, 2] & $\mathcal{V}_r$ & $\mathcal{V}_w$ & 0.7 \\
        
                           & Reach-Stay  & 0.6 & 30     & $\mathcal{V}_s$ & $\mathcal{V}_n$ & 0.2 \\
        
        \midrule
        
        \multirow{2}{*}{7} & Reach-Avoid & 0.8 & [1, 2] & $\mathcal{V}_r$ & $\mathcal{V}_w$ & 0.7 \\
        
                           & Reach-Stay  & 0.2 & 60     & $\mathcal{V}_s$ & $\mathcal{V}_w$ & 0.5 \\
        
        \bottomrule
    \end{tabular}
    %}
    \end{center}
    \vskip -0.1in
    \end{subtable}

    \begin{subtable}[t]{\textwidth}
    \caption{Sets of Boolean formulae from which $\beta_i^+$ and $\beta_i^-$ are sampled during training. We define $\neg\mathit{AP} = \{\neg \mathsf{p} : \mathsf{p} \in \mathit{AP}\}$, $\textstyle\bigwedge_k(S) = \{\beta_1 \wedge \cdots \wedge \beta_k : \beta_i \in S\}$, and $\bigvee_k(S)$ analogously. Note that the Boolean formulae must correspond to valid sets of assignments $\sigma \in \Sigma$ for each environment; if a formula contains an impossible clause (e.g., $\mathsf{at}_i \land \mathsf{at}_j$ conjunctions in \textit{Go1RGBZoneEnv}), that formula is excluded.}
    
    \label{tab:curriculum:sets}
    \begin{center}

    \begin{tabular}{lll}
        \toprule
        
        Set & Contents & Summary \\
        
        \midrule

        $\Lambda_k$ & $\bigwedge_k(\mathit{AP})$ & Clauses without negations \\

        $\Lambda^\pm_k$ & $\bigwedge_k(\mathit{AP}\cup\neg\mathit{AP})\ \setminus\ \big(\bigwedge_k(\mathit{AP}) \cup \bigwedge_k(\neg\mathit{AP})\big)$ & Clauses with negations \\

        $\mathrm{D}_p$ & $\bigcup_{k=2}^{4} \bigvee_k(\mathit{AP})$ & Disjunctions of propositions \\

        $\mathrm{D}_m$ & $\bigvee_2(\mathit{AP}\cup \Lambda_2 \cup \Lambda_3 \cup \Lambda_2^{\pm})\ \setminus\ \bigvee_2(\mathit{AP})$ & Disjunctions of clauses \\

        \midrule
        
        $\mathcal{V}_r$ & $\mathit{AP} \cup \Lambda_2 \cup \Lambda_3 \cup \Lambda^\pm_2 \cup \Lambda^\pm_3$ & ``Reach'' set \\

        $\mathcal{V}_s$ & $\mathit{AP} \cup \neg \mathit{AP} \cup \Lambda_2 \cup \Lambda_3 \cup \Lambda^\pm_2$ & ``Stay'' set \\

        $\mathcal{V}_n$ & $\mathit{AP} \cup \Lambda_2$ & ``Narrow avoid'' set \\

        $\mathcal{V}_w$ & $\mathit{AP} \cup \Lambda_2 \cup \mathrm{D}_p \cup \mathrm{D}_m$ & ``Wide avoid'' set \\
        
        \bottomrule
    \end{tabular}
    %}
    \end{center}
    \vskip -0.1in
    \end{subtable}

    \begin{subtable}[t]{\textwidth}
    \caption{Template sets of propositions $\mathit{AP}$. For each predicate $\mathsf{f}$ we list the maximum number of predicate instances, as well as the probabilities $\mathrm{P}_\mathrm{rch}$ and $\mathrm{P}_\mathrm{avd}$ that an instance of that predicate is sampled when a proposition is sampled to build $\beta_i^+$ and $\beta_i^-$, respectively. While trained Plato and Discretize models admit an arbitrary number of predicate instances, Concatenate and SIAMS do not because the number of instances determines their model input sizes, and thus they require a maximum number of instances per predicate defined a priori.}
    
    \label{tab:curriculum:propositions}
    \begin{center}

    \begin{tabular}{ll ccc}
        \toprule
        
        Environment & $\mathsf{f}$ & Count & $\mathrm{P}_\mathrm{rch}$ & $\mathrm{P}_\mathrm{avd}$ \\
        
        \midrule

        Go1RGBZoneEnv & $\mathsf{at}$ & 4 & $1$ & $1$ \\

        \midrule

        X2DronesEnv & $\mathsf{pos}$ & 6 & $1$ & $1$ \\

        \midrule

        \multirow{3}{*}{FrankaCubesEnv} & $\mathsf{pos}$ & 4 & $\sfrac{1}{3}$ & $\sfrac{1}{3}$ \\

        & $\mathsf{rot}$ & 4 & $\sfrac{1}{3}$ & $\sfrac{1}{3}$ \\

        & $\mathsf{has}$ & 2 & $\sfrac{1}{3}$ & $\sfrac{1}{3}$ \\

        \midrule

        \multirow{2}{*}{FalloutWorld} & $\mathsf{rad}$ & 1 & $\sfrac{1}{6}$ & $\sfrac{3}{4}$ \\

        & $\mathsf{loc}$ & 6 & $\sfrac{5}{6}$ & $\sfrac{1}{4}$ \\
        
        \bottomrule
    \end{tabular}
    %}
    \end{center}
    \vskip -0.1in
    \end{subtable}
    \end{center}
\end{table*}

\clearpage

\clearpage

\clearpage

\clearpage

\end{document}